\documentclass[10pt,twocolumn,letterpaper]{article}

\usepackage{cvpr} 
\usepackage{times}
\usepackage{epsfig}
\usepackage{graphicx}
\usepackage{amsmath}
\usepackage{amssymb}

\usepackage{multirow}
\usepackage{caption}
\usepackage{subcaption}
\usepackage[table]{xcolor}
\usepackage{enumitem}
\usepackage[pagebackref=false,breaklinks=true,colorlinks,bookmarks=false]{hyperref}

\newcommand{\M}[1]{\mathtt{#1}}
\newcommand{\V}[1]{\M{#1}}

\newcommand{\arr}[2]{\begin{array}{#1} #2\end{array}}
\newcommand{\mat}[2]{\left[\!\!\arr{#1}{#2}\!\!\right]}

\newcommand{\rotname}[1]
{\rotatebox{90}{#1}}

\definecolor{green}{rgb}{0.4, 1.0, 0.0}
\definecolor{red}{rgb}{1.0, 0.03, 0.0}
\definecolor{orange}{rgb}{0.93, 0.57, 0.13}

\usepackage{epstopdf}
\DeclareGraphicsExtensions{.jpg,.pdf}

\def\cvprPaperID{11129} 
\def\confName{CVPR}
\def\confYear{2022}

\title{3D Reconstruction from public webcams}
\begin{document}

\author{Tianyu Wu\\
ETH Zurich\\
{\tt\small wuti@student.ethz.ch}
\and
Konrad Schindler\\
ETH Zurich\\
{\tt\small schindler@ethz.ch}
\and
Cenek Albl\\
ETH Zurich\\
{\tt\small cenek.albl@geod.baug.ethz.ch}
}

\maketitle

\begin{abstract}
  We investigate the possibility of 3D scene reconstruction from two or more overlapping webcam streams. A large, and growing, number of webcams observe places of interest and are publicly accessible. The question naturally arises: can we make use of this free data source  for 3D computer vision? It turns out that the task to reconstruct scene structure from webcam streams is very different from standard structure-from-motion (SfM), and conventional SfM pipelines fail. In the webcam setting there are very few views of the same scene, in most cases only the minimum of two. These viewpoints often have large baselines and/or scale differences, their overlap is rather limited, and besides unknown internal and external calibration also their temporal synchronisation is unknown. On the other hand, they record rather large fields of view continuously over long time spans, so that they regularly observe dynamic objects moving through the scene. We show how to leverage recent advances in several areas of computer vision to adapt SfM reconstruction to this particular scenario and reconstruct the unknown camera poses, the 3D scene structure, and the 3D trajectories of dynamic objects.
\end{abstract}

\section{Introduction}

Reconstructing a 3D scene from multiple images is fundamental problem of computer vision, and solutions for increasingly general and uncontrolled settings have matured over the past decades. Packages derived from former research codes, like Bundler~\cite{Snavely2007ModelingTW} or COLMAP~\cite{schoenberger2016sfm}, as well as commercial software like Pix4D~\cite{Pix4d2021} or Metashape~\cite{Agisoft2021} are able to reconstruct camera poses and 3D scene representations from large, unordered image sets. Structure from motion (SfM) has also been adapted to run on a mobile devices~\cite{SchopsSHP17}, and to process street-view images~\cite{Torii2009} and even photographs taken under water~\cite{KangWY12}.

Yet, none of the existing methods work when fed images from public webcams. The reason is that these deviate from the common scenario in several important aspects and challenge the assumptions underlying the classical SfM pipeline.
Real webcams in some sense present a worst-case scenario for conventional SfM: they combine {\bf unknown internal and external calibration}, typically including strong lens distortion, with a {\bf very small number of viewpoints} (predominantly only two), in many cases {\bf very wide baselines}, and a chronical {\bf lack of good feature points}.
On the contrary, webcams have the potential advantage that they continuously record from the same viewpoint for a long time. With recent advances in dynamic SfM, this brings about the possibility to use {\bf moving objects} for calibration purposes. But to that end one must reconstruct those objects, rather than discard them as non-static outliers, as standard SfM does. And one must recover the normally also {\bf unknown synchronisation of the cameras}.

\begin{figure}[!t]
    \centering
    \includegraphics[width=.48\columnwidth]{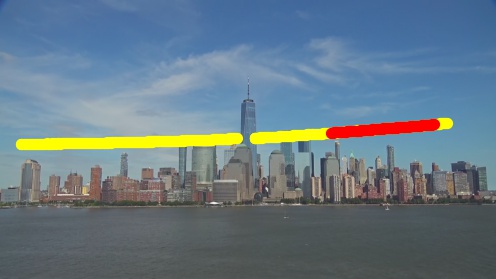}
    \includegraphics[width=.48\columnwidth]{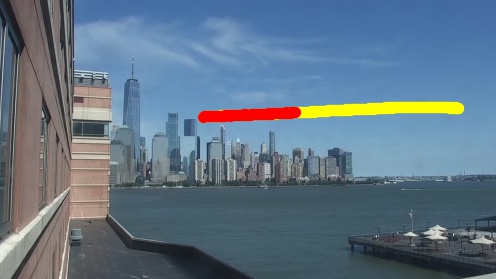}
    \includegraphics[width=.96\columnwidth]{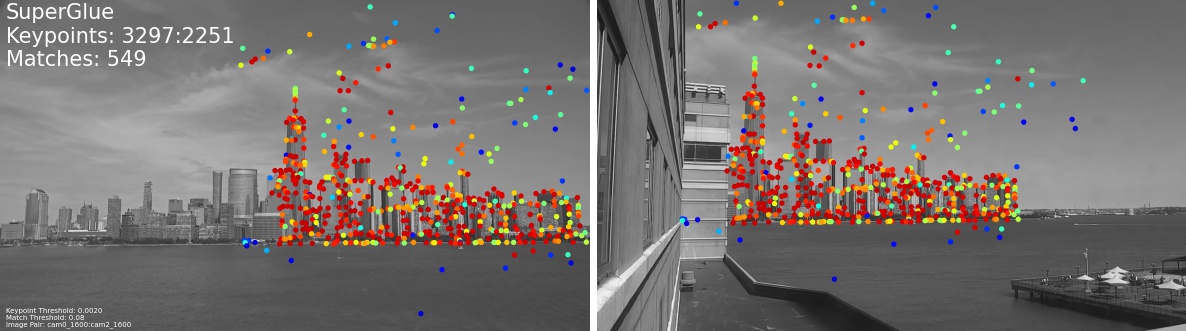}
    \includegraphics[width=.48\columnwidth, trim={10cm 5cm 10cm 4cm}, clip]{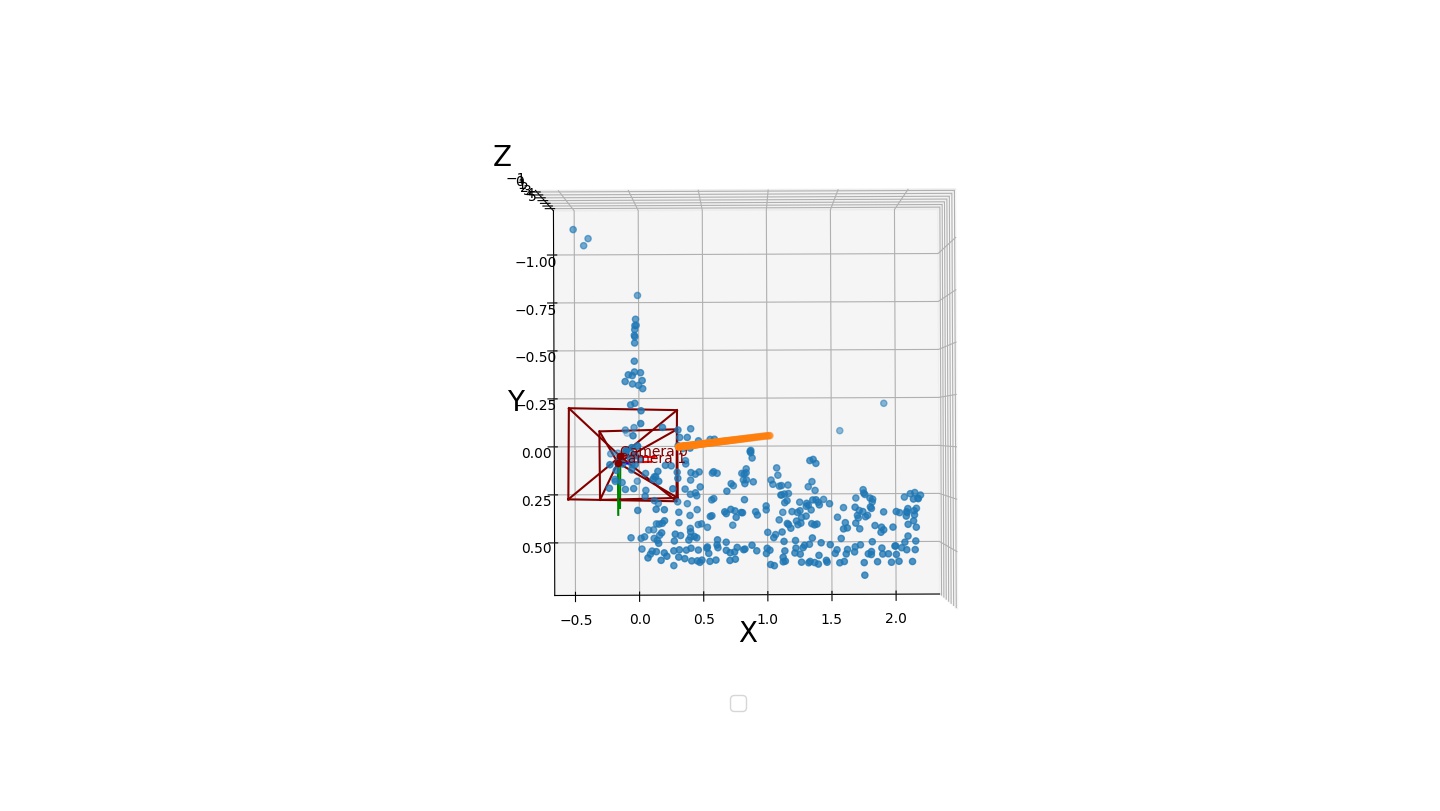}
    \includegraphics[width=.48\columnwidth, trim={10cm 5cm 10cm 4cm}, clip]{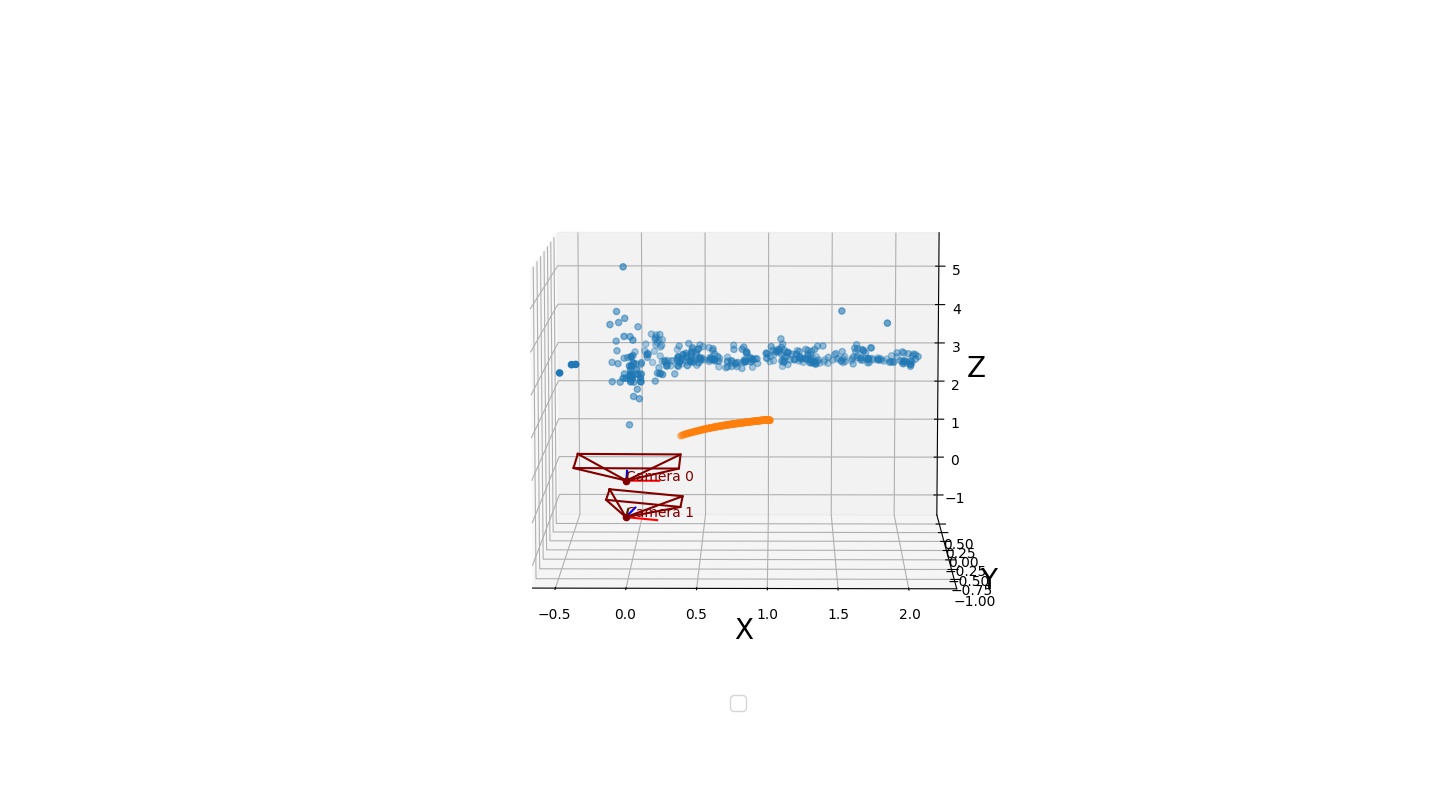}
    \caption{We demonstrate 3D reconstruction from public webcams, using the static scene content as well as the moving objects.}
    \label{fig:teaser}
\end{figure}

The starting point for our work is that recent research has actually addressed many of these challenges in isolation, including single-view camera calibration~\cite{Pritts-PAMI20,lochman2021minimal}, two-view camera calibration~\cite{KukelovaEfficient2015, KukelovaRadial2019}, image matching under extreme viewpoint differences~\cite{detone18superpoint,sarlin20superglue} and joint camera synchronisation and reconstruction of 3D moving objects~\cite{li2020reconstruction}.
We make use of these new capabilities to overcome the challenges listed above, and augment the established SfM pipeline in several ways so as to enable 3D reconstruction from webcams in the wild.
Webcam videos may comprise a large volume of pixels, but the amount of actual 3D information in them is actually rather limited. Hence, we aim to exploit as much as possible of that information, including the static scene that forms the basis of standard SfM, but also scene parts that are usually neglected or actively discarded, such as the sky and dynamic objects.

Webcams are mounted all over the world, and a large portion of them is freely accessible to the public. Typically they are hosted online 24/7 as live streams (e.g., \url{EarthCam.com}, \url{webcamtaxi.com}). For obvious reasons, most webcams are set up such that they view an open space and avoid clutter and occlusions: they show large squares, skylines from afar, harbours, landscapes from scenic viewpoints, etc.
It may seem that such unencumbered views would benefit computer vision, but there is a subtle caveat, namely \emph{what} we see from such viewpoints. Often large portions of the images are taken up by surfaces that are generally hard to match, such as the sky or water bodies; or that are hard to match across wide baselines, such as road surfaces, grassland, etc.

Perhaps the fact that most webcam streams are designed to simply offer a beautiful or interesting view to the spectator, from the comfort of their living room, is a reason why they have not yet been used for 3D vision: they are just not made for it.
We argue that one may nevertheless want to try. An important driver of modern SfM has been to handle images "not made" for 3D reconstruction. And webcams with their wide and continuous coverage are a rich source of information that could potentially serve more, and more advanced, applications if linked to the 3D world.
With this paper we set out to answer the following questions:
\begin{itemize}[leftmargin=4mm,topsep=4pt,itemsep=2pt,partopsep=6pt, parsep=0pt]
    \item Is it feasible to reconstruct scenes in 3D from in-the-wild, public webcam streams?
    \item What extensions of the SfM pipeline are needed to reconstruct under the uncommon recording conditions of webcams, and which recent computer vision building blocks can cover them?
    \item What practical considerations arise when working with such data, and how to tackle them?
\end{itemize}   
To the best of our knowledge, the present work is the first to explore 3D computer vision and multi-view reconstruction from video streams of public webcams.

\subsection*{Related work}

In 2006, the work of~\cite{Snavely2006PhotoTE,Snavely2007ModelingTW} took the field of computer vision by storm, by putting together an SfM pipeline that could reconstruct 3D scenes, in particular much-photographed landmarks, from community photo collections. Many follow-up works have since contributed to making the general approach more efficient and more robust, allowing for reconstruction at city scale~\cite{bulbul2015,agarwal2010reconstructing,agarwal2011building} and eventually world scale~\cite{HeinlyReconstructingWorld,schoenberger2016sfm,schoenberger2016mvs}, in the latter case ingesting up to 100 million crowd-sourced images. Moreover, up-to-date information from the most recent images can be used to detect scene changes and to efficiently update reconstructions through time \cite{bulbul2015}. Overall, this line of work---a sort of "second wave" of SfM research after the pioneering works of the 1990ies that established the theoretical foundations~\cite{hartley_zisserman_2004}---proved that, in locations that many people take pictures of, 3D reconstruction from opportunistically sourced, uncontrolled images was feasible and potentially useful. Here, we take that same ambition one step further and ask: is it also possible to reconstruct the 3D geometry of locations observed by two (or more) webcams?


However, while reconstruction from community photo collections and from webcams share common theoretical foundations, there are significant differences. These arguably make SfM from webcam more challenging, and they definitely mean that conventional SfM pipelines designed for photo collections are not applicable.
Unlike the hundreds or thousands of viewpoints per scene that afford redundancy, and thus robustness, in the classical setting~\cite{Agarwal2010}, webcams are placed such that the scene of interest is covered by few viewpoints, in most cases only two to three views of the same area.
As the layout is optimised for coverage, the viewpoint differences are typically large and often extreme (\cf~Figure~\ref{fig:rott_match}), and the actual overlap that is visible in two or more cameras is small.
Hence, feature matching is predictably challenging.
Also, while SfM can in principle include self-calibration, it is in practice greatly stabilised if at least for some images approximate intrinsics are available. In community photo collections, rough focal lengths for some images are often provided in the image files' EXIF headers. Whereas for webcams no such information is available, moreover they typically feature very wide fields-of-view with associated lens distortions. And while the static camera poses, in principle, admit the use of correspondences from moving objects, this requires the frame rates and temporal offsets (the synchronisation) between cameras to be known.

Fortunately, the past few years have brought significant progress in several research fields that are relevant in our context. Camera calibration has been revisited in the light of modern algebraic solvers, and it is now possible to calibrate intrinsic parameters, including radial distortion, from as few as two~\cite{KukelovaEfficient2015, KukelovaRadial2019} or even a single view\cite{Pritts-PAMI20,lochman2021minimal}. 
Image matching under challenging conditions, and across large baselines, has seen further improvements \cite{mishkin2015mods}, especially with the rise of feature representations extracted with deep learning~\cite{sarlin20superglue}, and can increasingly cope even with extreme conditions \cite{Jin_2020}.
Robust estimation continues to evolve and achieve efficient calibration and pose estimation despite unavoidable, high outlier ratios \cite{raguram2012usac, barath2019magsac,barath2020magsacpp}.
Finally, recent work on dynamic SfM~\cite{albl2017two,li2020reconstruction} has shown how to jointly estimate the time synchronisation and relative pose of two cameras, using only the trajectories of moving objects (respectively, feature points) as input.

Taken together, several new algorithmic components are at our disposal that may play a role for 3D reconstruction from webcams. So far, they have been developed and tested in isolation, on custom data geared towards the specific problem (e.g., two-view calibration \emph{given} suitable correspondences; pose estimation under hard matching conditions with \emph{known} calibration; etc.).
In this work we assemble, for the first time, an SfM pipeline and all these recent advances into a complete system, and show that this makes 3D reconstruction possible with only a small number (typically two) uncontrolled webcam streams as input.
Besides the scientific curiosity to "see what is feasible" and the potential applications of 3D webcam vision, we also argue that the webcam scenario constitutes an ideal playground to investigate the benefits and limitations of recently developed calibration and matching algorithms, since it confronts them with a realistic situation where they are needed and can, potentially, add value over their established predecessors.



\section{Problem formulation} \label{sec:problem_form}
Suppose we have $N \geq 2$ webcams observing the same area. Camera $i$ captures $M_i$ frames $I^i_j$, $j \in 0, \ldots, M_i$ at times $t^i_j$ and will be described by a set of parameters $\M{P}_i, \V{d}_i$ where $\M{P}_i$ is a $3 \times 4$ camera matrix and $\V{d}_i$ are the distortion parameters. $\M{P}_i$ can be further decomposed into $\M{P}_i = \M{K}_i[\M{R}_i | \V{t}_i]$ where $\M{R}_i$ is a $3 \times 3$ matrix describing the orientation of the camera, $\V{t}$ is the camera center in the camera coordinate frame and $\M{K}_i$ is the $3 \times 3$ calibration matrix. A 3D point $\V{X}_k, k \in 1,\ldots,L$ projects into camera $i$ as
\begin{equation}
    \V{u}^i_k = \rho(\V{d}_i,\mu(\M{K}_i(\M{R}_i \V{X}_k + \V{t}_i))),
    \label{eq:projection}
\end{equation}
where $\mu$ is the perspective division and $\rho$ is the distortion function. We choose $\V{X}_k$ to describe the static part of the scene, i.e. the landscape, buildings, etc. Apart from the static part, the cameras also observe moving objects $O_m$, $m \in 0, \ldots, Q$ whose immediate 3D coordinates $\V{O}^i_{mj}$ are projected at times $t^i_j$ into frame $j$ of camera $i$ as $\hat{\V{o}}^i_{mj}$. We call these objects and their projections the dynamic part of the scene. Our aim is to retrieve $\V{X}_k$, $\V{O}^i_{mj}$, $\M{P}_i$ and $d_i$ from the observed $\hat{\V{u}}^i_k$ and $\hat{\V{o}}^i_{mj}$.
\section{Method} \label{sec:method}
\subsection{Camera calibration}
Since the web cameras we are dealing with can be located almost anywhere around the globe, it would be impractical to introduce fiducials, such as a calibration board, for camera calibration~\cite{Zhang2000AFN}. Instead, we have to rely only on the scene content, commonly referred to as auto-calibration. Most sites are observed with only two webcams, thus there are two possible approaches to auto-calibration. One is to use a single image, making some assumptions about the scene content~\cite{Pritts-PAMI20,lochman2021minimal}, \eg, presence of repetitive patterns or a Manhattan frame. The other considers two views of the scene and computes the fundamental matrix along with the distortion parameters~\cite{KukelovaEfficient2015}. Both methods model lens distortion with the one-parameter division model for its algebraic simplicity and effectiveness for distortion removal. Under this model, the distortion function $\rho$ in equation~\ref{eq:projection} can be written as
\begin{equation}
    \rho(\mat{cc}{x & y}^\top) = \frac{\mat{cc}{x & y}^\top}{1 + d_0 ((x-x_0)^2 + (y-y_0)^2)}
\end{equation}
where $x_0, y_0$ are the coordinates of the principal point.

For the single-view approach, we adopt the hybrid solver with six circular arcs (Solver 6CA) proposed by \cite{lochman2021minimal}, which takes an input image and recovers the parameters by identifying Manhattan Frames (MF) in the image. The algorithm deals with two configurations in particular: (i) the vanishing points are coplanar; (ii) the vanishing points are mutually orthogonal. In both configurations, vanishing points are represented as the intersection of the imaged parallel lines in the scene. With the single-view approach, sample images from every camera are processed individually to acquire their corresponding calibrations.

Auto-calibration of radially distorted cameras was introduced in~\cite{Fitzgibbon2001SimultaneousLE}, where the fundamental matrix $\mathtt{F}$ was estimated along with a single radial distortion parameter, assuming the two images were taken with the same camera. This approach was later extended, \eg,~\cite{KukelovaEfficient2015} for the case of two different cameras, each with its own radial distortion coefficient, which is more suitable for our task. The input to the algorithm is a set of matches between a pair of cameras, based on which the fundamental matrix $\M{F}$ and radial distortion coefficients are estimated using a 10-point solver and RANSAC. The focal lengths of the camera pair are then recovered from $\M{F}$~\cite{hartley_zisserman_2004}, assuming the camera principal point is in the center of the image and there is no skew.

\subsection{Feature matching}
Both dynamic features (\ie, moving objects) and static features (\ie, the structure) in the scene are extracted in our method, using the feature extraction and matching module and the object tracking module, respectively.

\subsubsection{Static features}
Considering the challenging image matching conditions, we employ the state-of-the-art feature extraction and matching techniques SuperPoint~\cite{detone18superpoint} and SuperGlue~\cite{sarlin20superglue}  to extract and match the static features of the scene. SuperPoint is a fully convolutional model trained for interest point detection and description, which feeds a full-sized image as an input into a shared VGG-style encoder to encode the information, then employs two separate decoders to learn task-specific weights for feature extraction and description. This SuperPoint model is connected to the SuperGlue architecture, which acts as a deep middle-end matcher that takes two sets of local features and their descriptors and outputs the correspondences as well as rejects non-matchable points. To produce matches between each pair of camera views, images sampled from the video sequences are processed by the end-to-end SuperPoint+SuperGlue model provided by \cite{sarlin20superglue}, which was trained on Megadepth for the outdoor environment. The inferred matches are stored and used directly as input for the subsequent 3D reconstruction process. We compare it to the classic approach of exhaustively matching SIFT~\cite{lowe2004distinctive} features. The tentative matches from both methods are subsequently verified using RANSAC and its state-of-the-art variant MAGSAC~\cite{barath2019magsac,barath2020magsacpp}.

\subsubsection{Dynamic features}
To extract the dynamic components of the scene, we exploit object tracking algorithms to follow both large slow-moving objects (\eg, ships, pedestrians in the near field), and small fast-moving objects (\eg, distant helicopters and planes). The large, slow-moving objects are tracked using the off-the-shelf CSRT~\cite{Lukei2017DiscriminativeCF} tracker as implemented in the OpenCV library, together with a pretrained YOLOv5 (based on~\cite{Bochkovskiy2020YOLOv4OS}) + DeepSORT~\cite{Wojke2018deep}, while the small fast-moving objects are tracked by a Chained Tracker~\cite{Peng2020ChainedTrackerCP} that we train on drone videos. The extracted tracklets are converted into 2D trajectories of the bounding box center points with associated frame ids.

The video sequences of each camera are processed separately to extract their 2D trajectories. In the case of synchronized cameras, the trajectories of the same object can be considered as dynamic correspondences across different cameras. However, due to the lack of synchronization of the webcams, one further step is needed. The quasi-minimal solver proposed in \cite{albl2017two}, \cite{li2020reconstruction} is applied to jointly estimate the synchronization parameters and the fundamental matrix of the two camera views from a set of 2D trajectories. The time mapping between cameras is parametrised with a scale factor $\alpha$ that encodes the relationship between the frame rates (\ie, FPS), and a time offset $\beta$ that represents the initial time offset. As the raw dynamic features may contain noise, splines are fitted to the trajectories to achieve better accuracy. With the modeled time mapping, correspondences are generated by sampling the spline at a series of timestamps, leading to a linear system of equations that requires a minimum of eight points and is solved robustly with RANSAC. The process is carried out exhaustively for every pair of cameras to synchronise them and acquire an initial estimate of the time mappings between pairs.

\subsection{3D reconstruction}
At this point both the dynamic features and the static features of the scene are ready for reconstruction. The standard incremental SfM procedure is extended such that it can use only the static or only the dynamic features for the reconstruction, or use both groups together. In our experiments we will separately analyse the static-only, dynamic-only and static-dynamic settings.

To initialise the two-view geometry, the pair of cameras that has the largest temporal overlap is selected in the cases of dynamic-only, and static-dynamic settings, whereas the two cameras that generate the most matches are used for the static-only setting. For structure computation, the standard triangulation is applied for the static features; while the dynamic objects are represented by splines. Moreover, the bundle adjustment (BA) in the incremental SfM process is also adapted according to the 3D reconstruction settings. Here, we introduce three variants of BA that we experiment with in the pipeline.

\subsubsection{Standard BA}
In the static-only setting, the BA procedure is the same as for conventional SfM~\cite{Lourakis2009SBAAS,agarwal2010bundle}, where the 3D coordinates together with the camera intrinsics and extrinsics are optimized jointly based on the reprojection error.

\subsubsection{Spatio-temporal BA}
To parametrise the trajectories of moving objects we use the 3D spline representation, which allows to simultaneously estimate the time mapping parameters $\alpha_i, \beta_i$ between the cameras, as shown in~\cite{li2020reconstruction}. The projection of a dynamic object $O_m$ into camera $i$ at the time of capture of the $j$-th frame can be described as
\begin{equation}
    \V{o}^i_{mj} = \rho(\V{d}_i, \mu(\M{K}_i (\M{R}_i T_r(t_i^j) + \V{t}_i)))
\end{equation}
The cost function for the spatio-temporal BA is then
\begin{equation}
    c_t =  \sum_{(i,j,m) \in S} \|\V{o}^i_{mj} - \hat{\V{o}}^i_{mj}\|^2,
\end{equation}
where $C_u$ are the spline coefficients of the fitted 3D trajectory $T_r(t_i^j)$ and $S$ is a set of indices where each $(i,j,m)$ corresponds to object $m$ being visible in frame $j$ of camera $i$. Here, only the spline coefficients are optimised, while the knot vectors are left unchanged to limit the complexity of the optimisation.

\subsubsection{Spatio-temporal BA extended with static features}
The optimisation of both the static and dynamic features together can be achieved by combining the two variants of BA. In this setup, both the reprojection error of the static features and the dynamic features contribute to the formulation of the optimisation problem as
\begin{equation}
    \underset{\mathbf{P}_i, \V{d}_i, \alpha_i, \beta_i, C_u}{\text{arg\,min}} c_t  + \sum_{(i,k) \in S} \|\V{u}^i_k - \hat{\V{u}}^i_k\|^2,
\end{equation}
where $S$ is the set of indices where a 3D point $k$ is visible in camera $i$.

\section{Pipeline}
\begin{figure*}[t]
    \centering
    \includegraphics[width=0.9\linewidth]{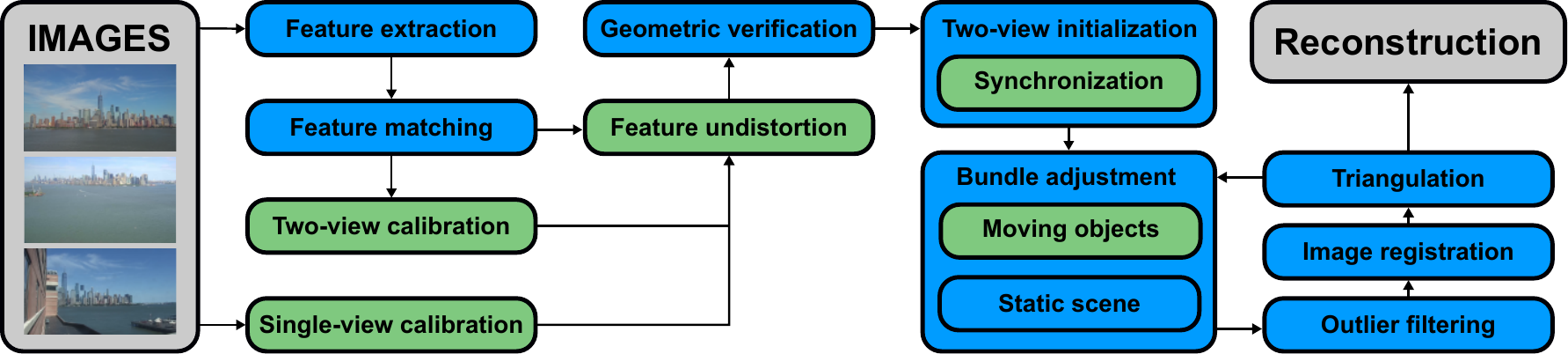}
    \caption{The proposed adjustment (green) to the standard SfM pipeline (blue).}
    \label{fig:pipeline}
\end{figure*}

Using the aforementioned methods, we propose an upgrade of the classic SfM pipeline (\eg, COLMAP~\cite{schoenberger2016sfm}) to accomodate 3D reconstruction from webcams, see Figure \ref{fig:pipeline}. The pipeline takes as input a set of video sequences from multiple public web cameras with overlapping fields of view, and preprocesses them with the camera auto-calibration module to obtain coarse initial values for the intrinsics and the distortion model. The static and dynamic features in the scene are extracted using the feature extraction and matching module, respectively the object tracking module. All 2D image information is then fed into the reconstruction module to recover the 3D model.

\section{Experiments} \label{sec:experiment}
\subsection{Datasets}
To test the performance of the proposed pipeline and its variants we have gathered nine real-world datasets by recording publicly available webcam streams from different locations. As shown in Table ~\ref{tab:datasets}, various types of moving objects appear in the data (most commonly pedestrians and ships), with varying frequencies. In addition, because the video streams may come from different sources, the frame rates are often different.
\begin{table}[th]
    \centering
    \renewcommand{\tabcolsep}{4pt}
    \renewcommand{\arraystretch}{1}
    \begin{tabular}{c|c|c|c|c}
         Name                 & Location               & FPS & Type & $\#$ Obj \\ \hline
         Croatia              & Croatia   & 15, 10 & Pedestrian  & 160 \\
         NewYork             & USA        & 30, 25 & Ships, Planes & 15 \\
         TimeSq         & USA          & 29, 29 & Pedestrian, Cars & 500 \\
         AalsHav       & Netherland   & 20, 18 & Ships & 300 \\
         BranHot     & Netherland  & 15, 15 & Ships, Bikes & 300 \\
         BranPor       & Netherland  & 20, 15 & Ships & 50 \\
         RottPor       & Netherland  & 25, 15 & Ships & 80 \\
         TexAir        & Netherland      & 10, 10 & Jets & 80 \\
         LauwHav & Netherland & 20, 20 & Ships & 20
    \end{tabular}
    \caption{Real-world datasets and their properties. Note: the number of objects are the rough estimates over a period of one hour.}
    \label{tab:datasets}
\end{table}

\subsection{Feature matching}
\begin{figure}[t]
    \centering
    \includegraphics[width=\columnwidth]{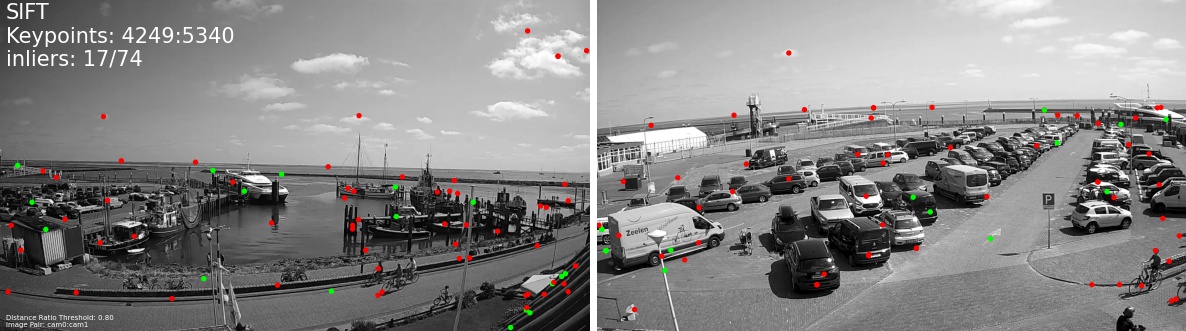}
    \includegraphics[width=\columnwidth]{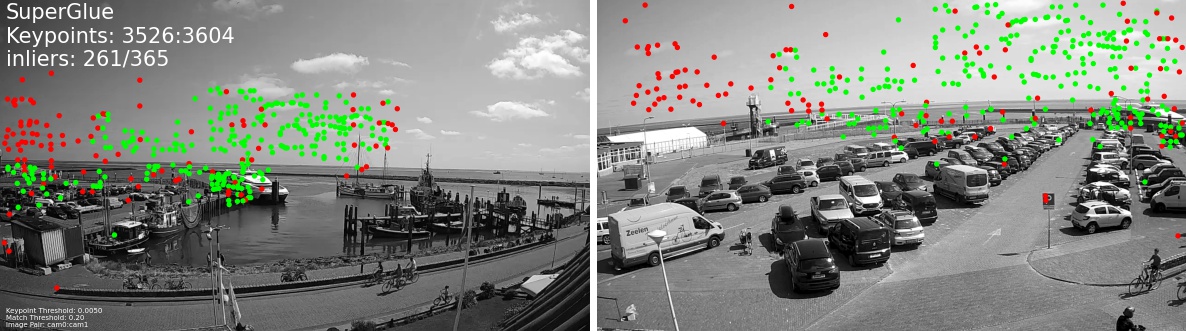}
    \includegraphics[width=\columnwidth]{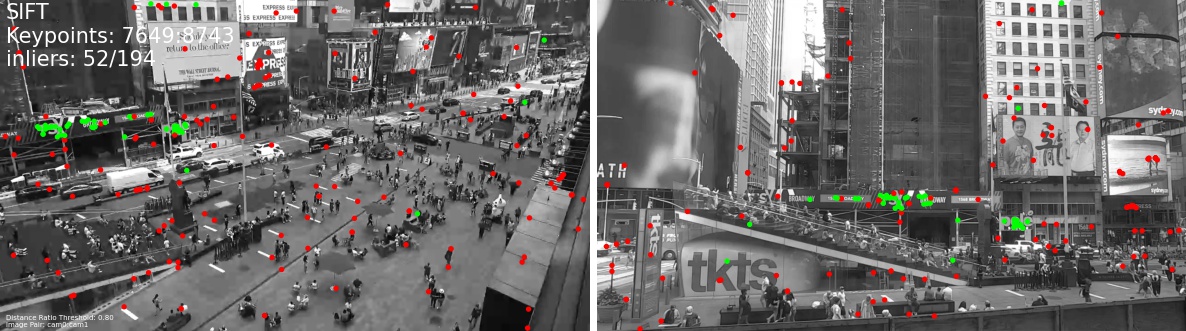}
    \includegraphics[width=\columnwidth]{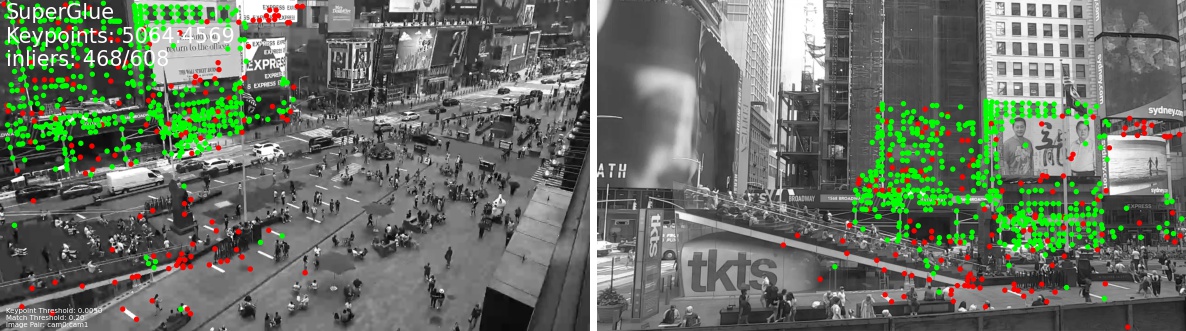}
    \caption{Sample images and matches obtained by SIFT and SuperGlue on BranPor (top 2) and TimeSq (bottom 2).}
    \label{fig:static_matches}
\end{figure}

First, we present the results of feature matching using state-of-the-art  SuperGlue, and compare it with the still ubiquitous standard approach of exhaustively matching SIFT~\cite{lowe2004distinctive} features. For every dataset, the image pairs are matched and inliers complying with the epipolar geometry (Fundamental matrix) are found using the recent MAGSAC robust estimator~\cite{barath2019magsac}. Note that, because there are lens distortions present in the dataset, the original image coordinates are undistorted with the estimated calibration parameters from the camera calibration process. Table \ref{tab:matching_results} summarizes the total number of tentative matches and the verified matches with the two matchers. For all datasets, SuperGlue consistently finds a larger number of reliable matches, especially when there are repetitive structures and significant viewpoint changes (Figure \ref{fig:static_matches}). A possible explanation for this may be that the classic matching approach with Lowe's ratio test does not account for the large-scale context of the features as effectively as the deep encoder, and gets confused more easily by repetitive structures. In contrast, SuperGlue benefits from the self- and cross-attention in the GNN. One interesting finding is that SuperGlue is often able to correctly recognise correspondences on clouds (Figure \ref{fig:rott_match}), which is helpful when the images contain mostly sky and little structure. In our experiments, MAGSAC provided significantly more inliers than classic RANSAC, and that gain was even more significant in combination with SuperGlue matches. For full results and comparisons please see the Appendix.

\begin{table}[th]
    \centering
    \begin{tabular}{c|c|c|c|c}
         Dataset              & SIFT & Verified & SuperGlue & Verified \\ \hline
         Croatia              & 434  & 17       & 505       & 379      \\
         NewYork             & 279  & 169      & 549       & 514      \\
         TimeSq          & 272  & 52       & 608       & 468      \\
         AalsHav       & 325  & 61       & 161       & 118       \\
         BranHot      & 660  & 580      & 607       & 580      \\
         BranPor       & 119  & 17       & 365       & 261      \\
         RottPor       & 454  & 177       & 453       & 255      \\
         TexAir        & 154  & 83       & 184       & 176      \\
         LauwHav  & 13   & 8        & 323       & 275      \\
    \end{tabular}
    \caption{Results of matching}
    \label{tab:matching_results}
\end{table}

\subsection{Single-view versus two-view calibration}
\begin{figure}[t]
    \centering
    \includegraphics[width=.3\columnwidth]{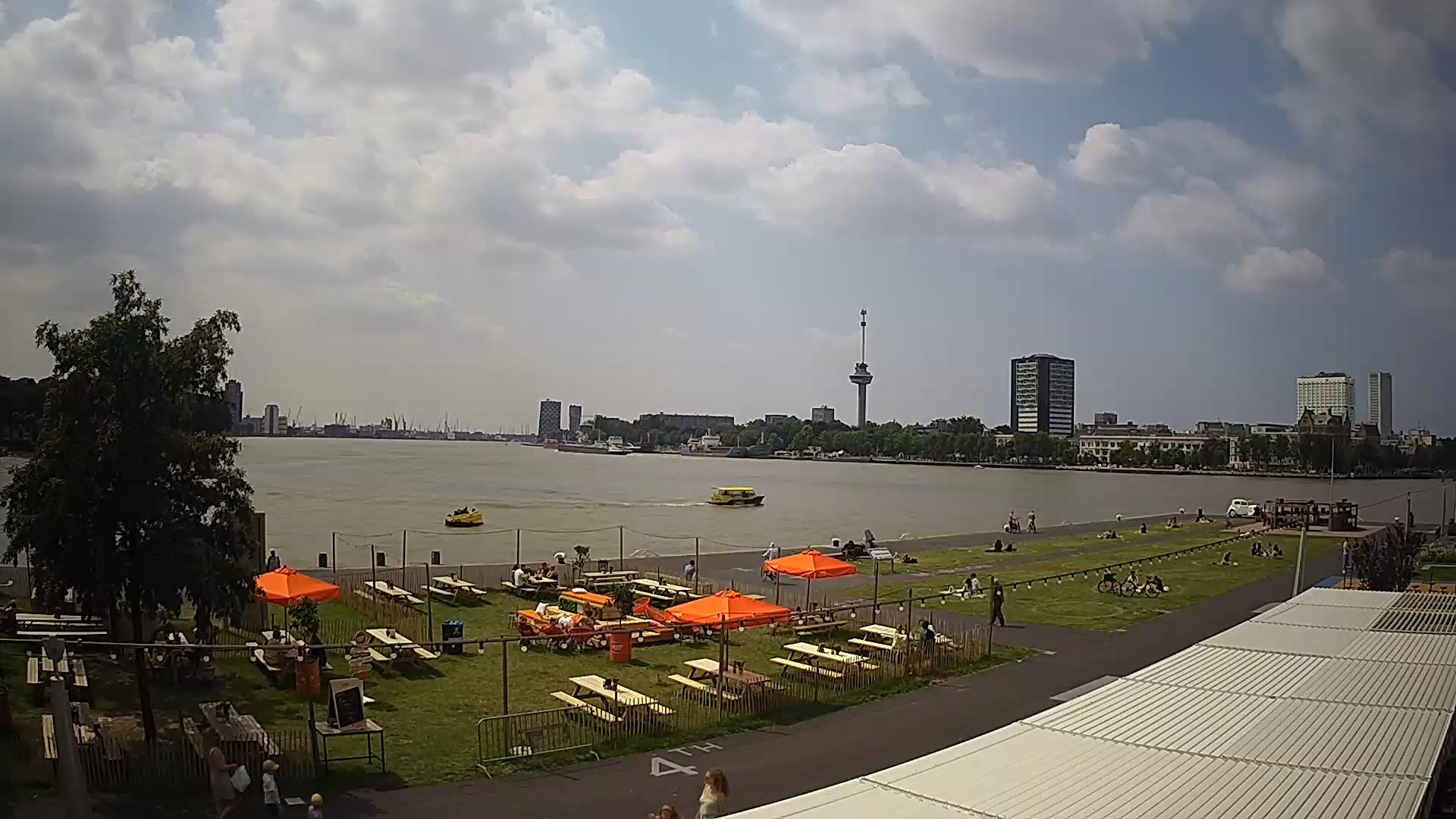}
    \includegraphics[width=.3\columnwidth]{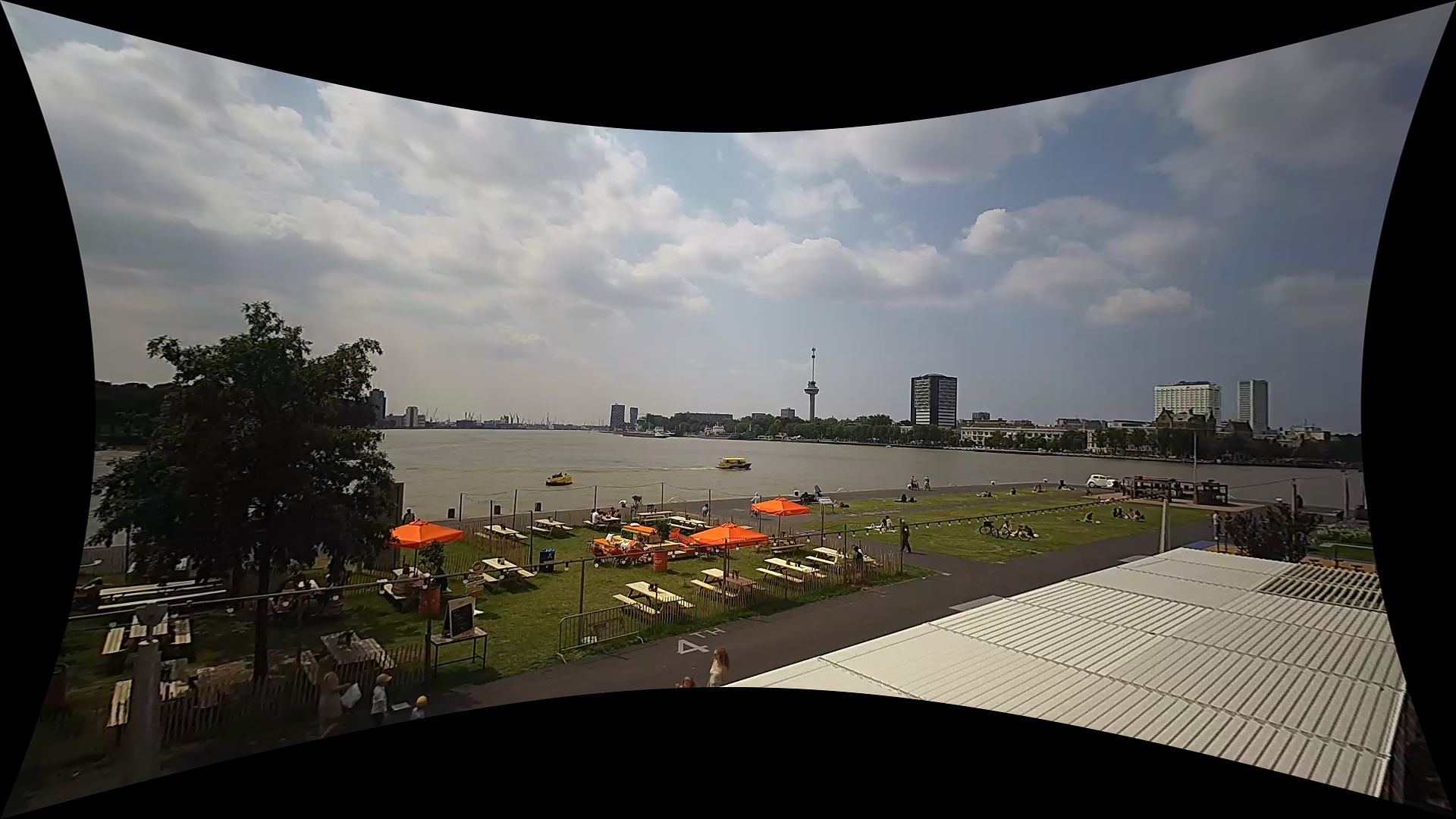}
    \includegraphics[width=.3\columnwidth]{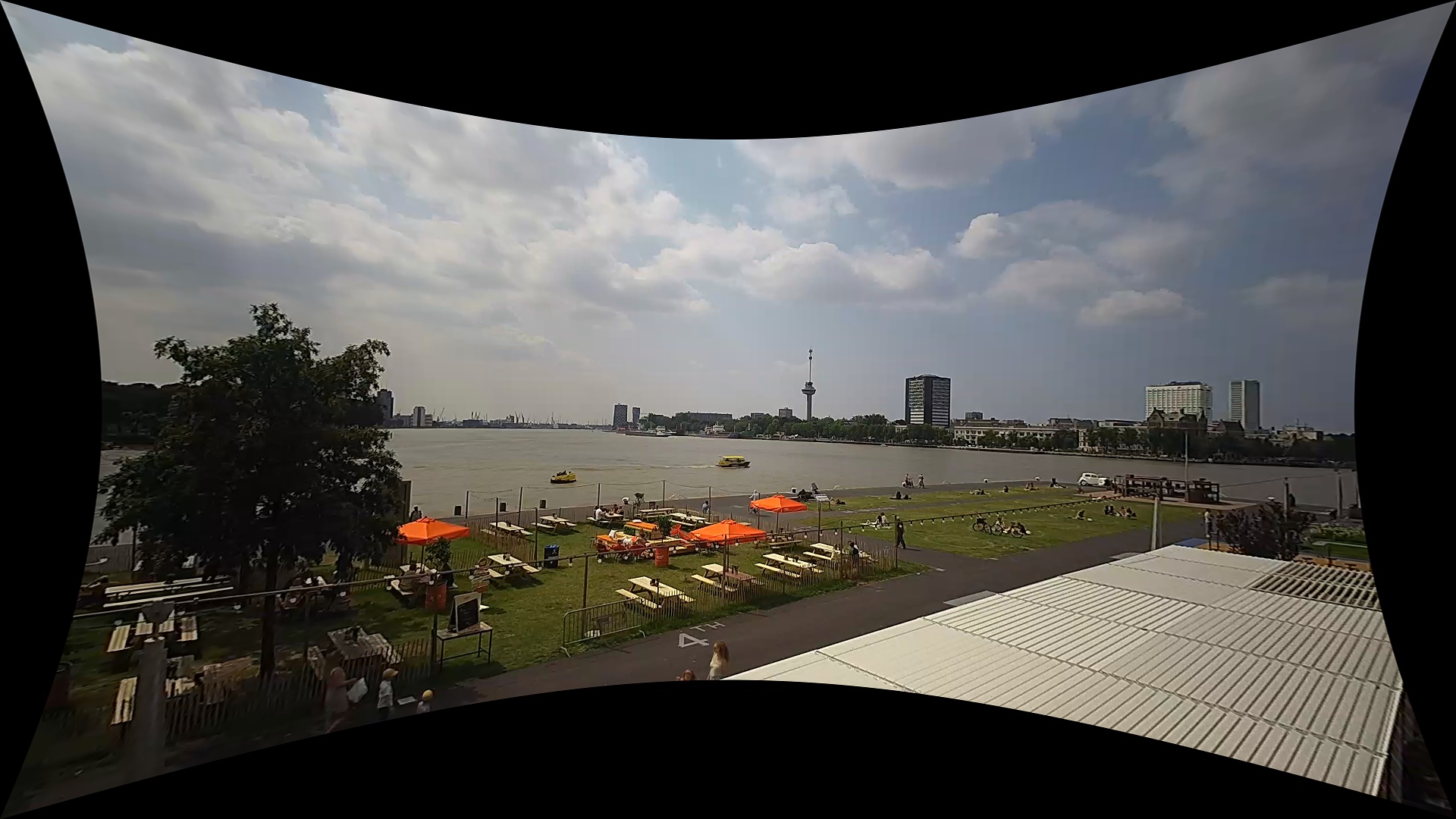}
    \includegraphics[width=.3\columnwidth]{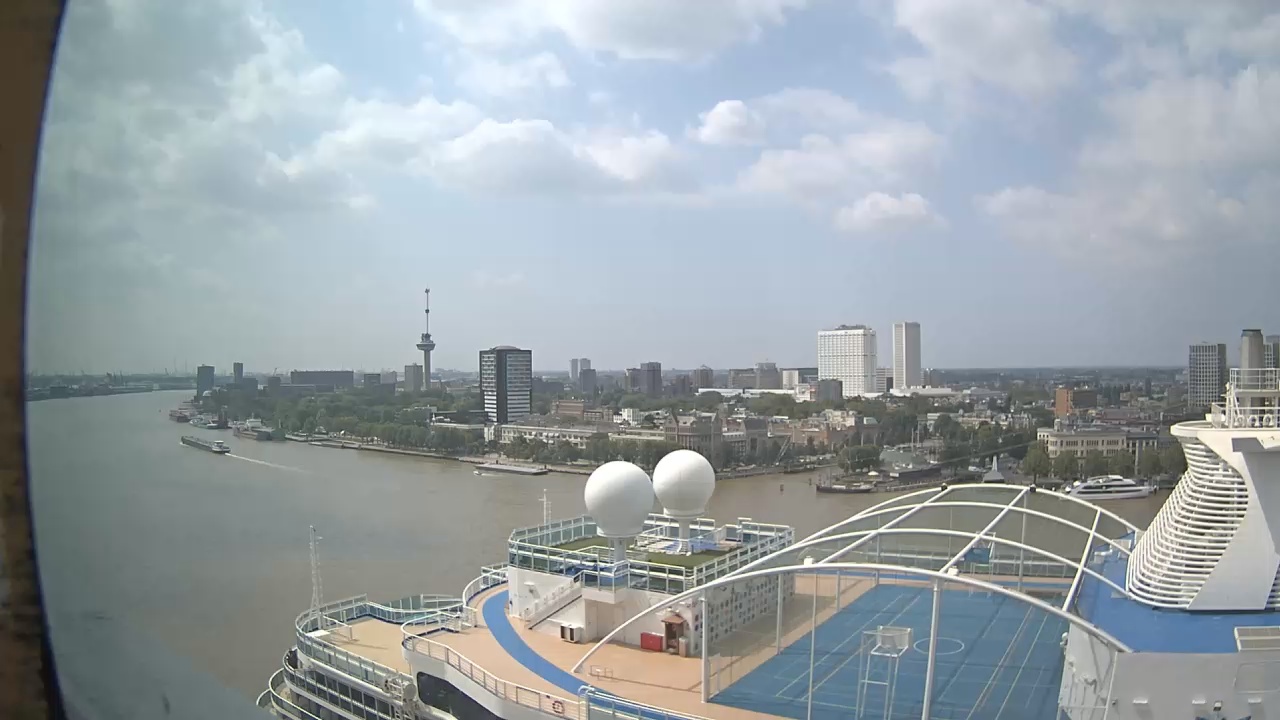}
    \includegraphics[width=.3\columnwidth]{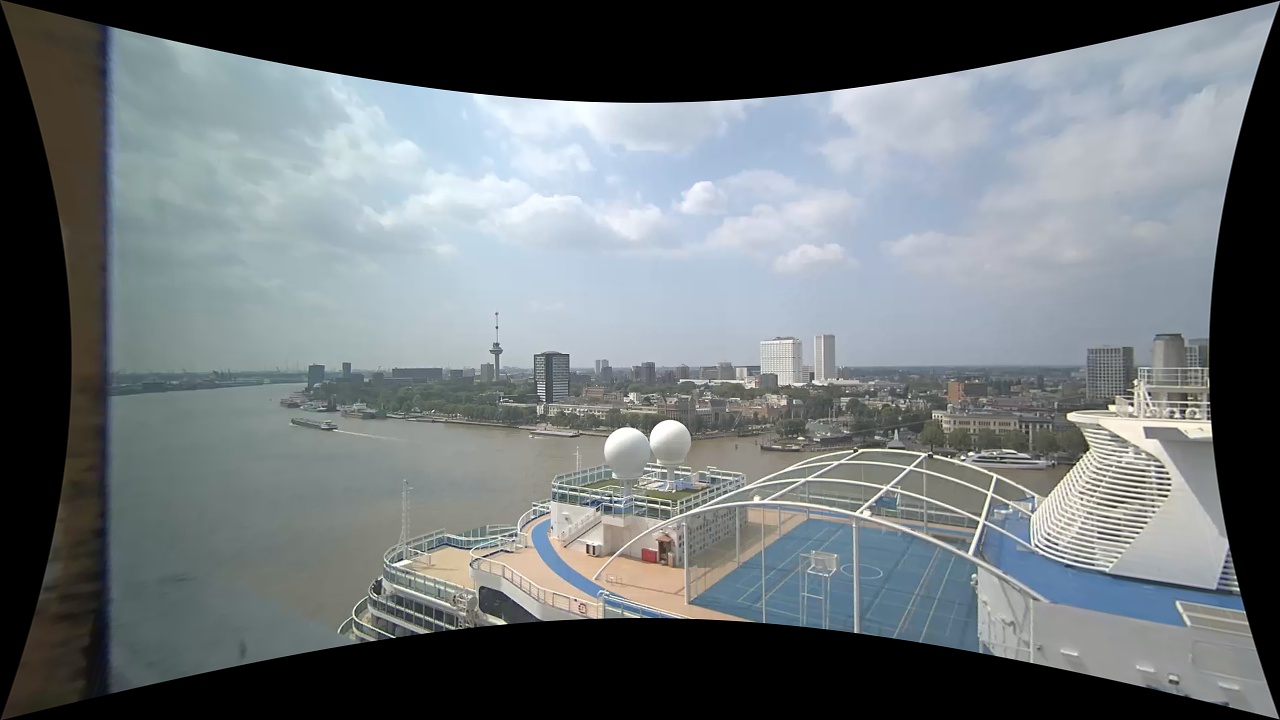}
    \includegraphics[width=.3\columnwidth]{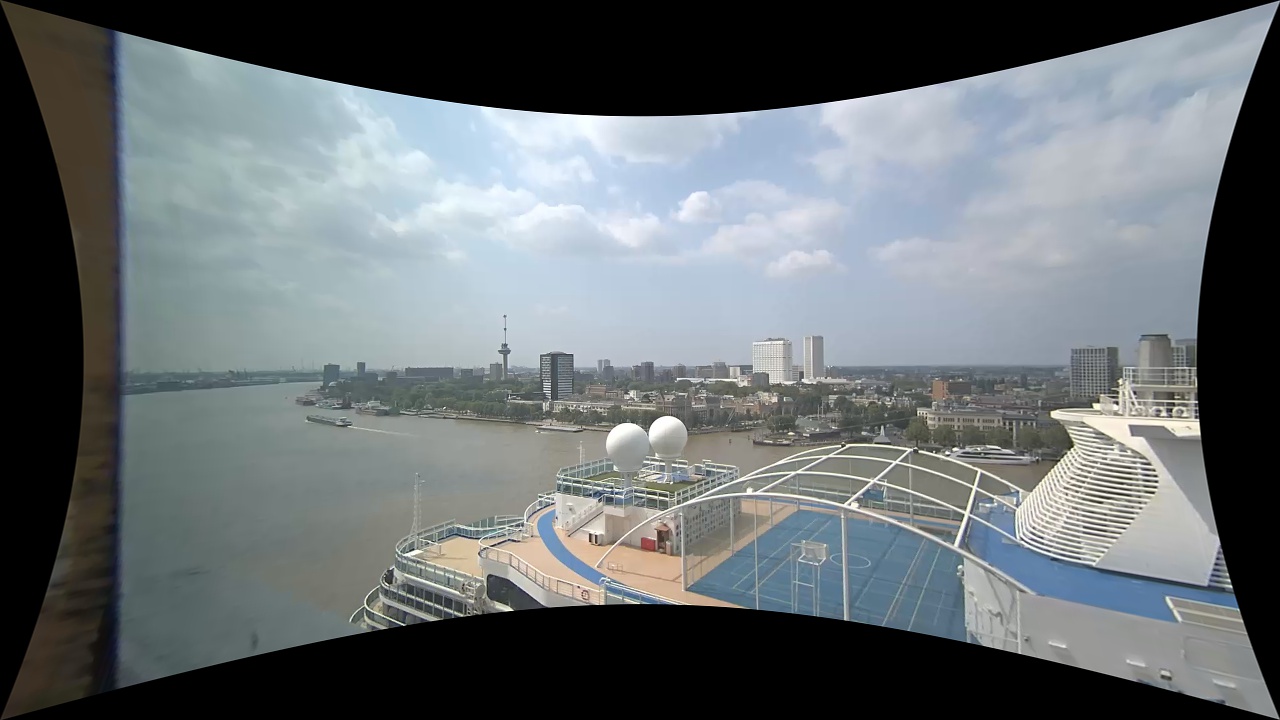}
    \includegraphics[width=.3\columnwidth]{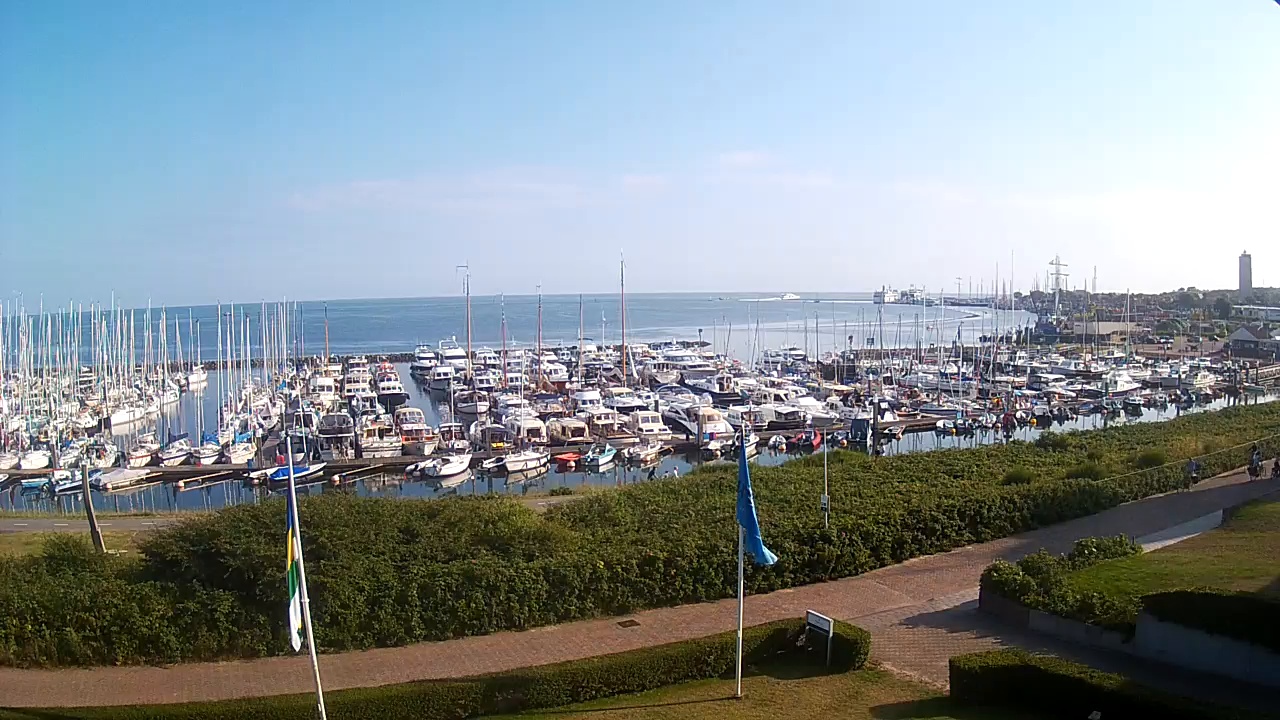}
    \includegraphics[width=.3\columnwidth]{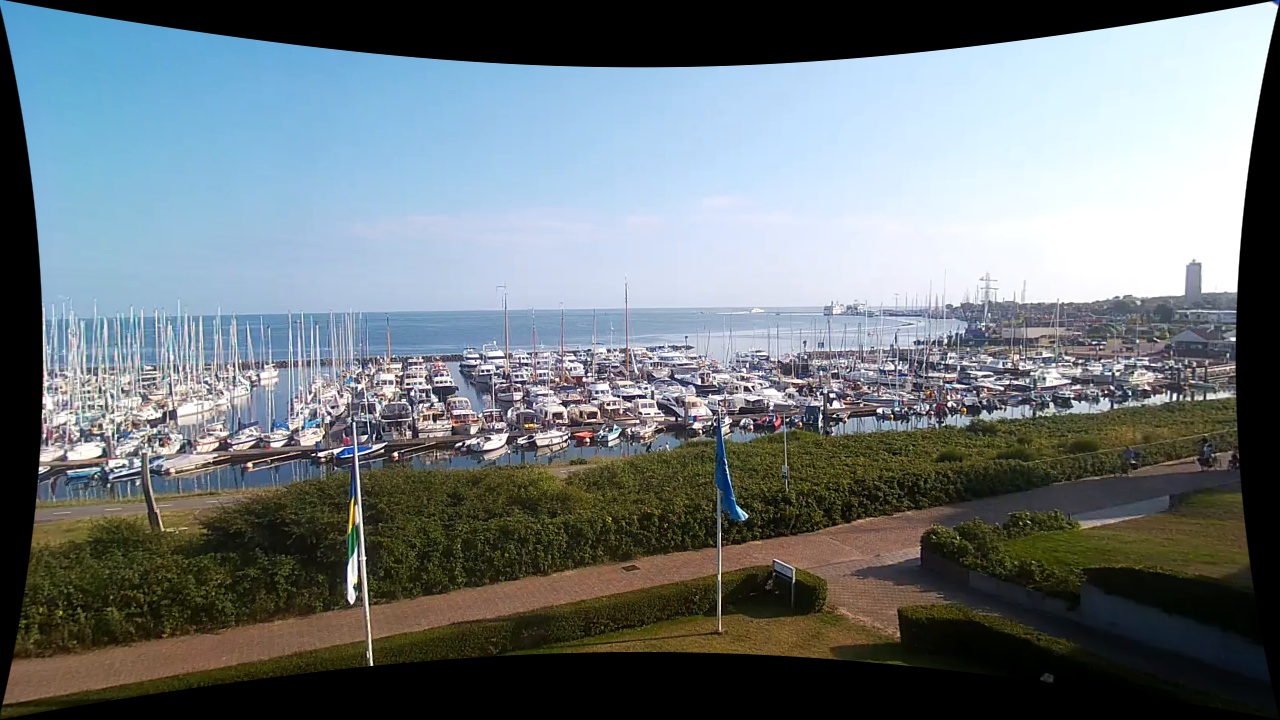}
    \includegraphics[width=.3\columnwidth]{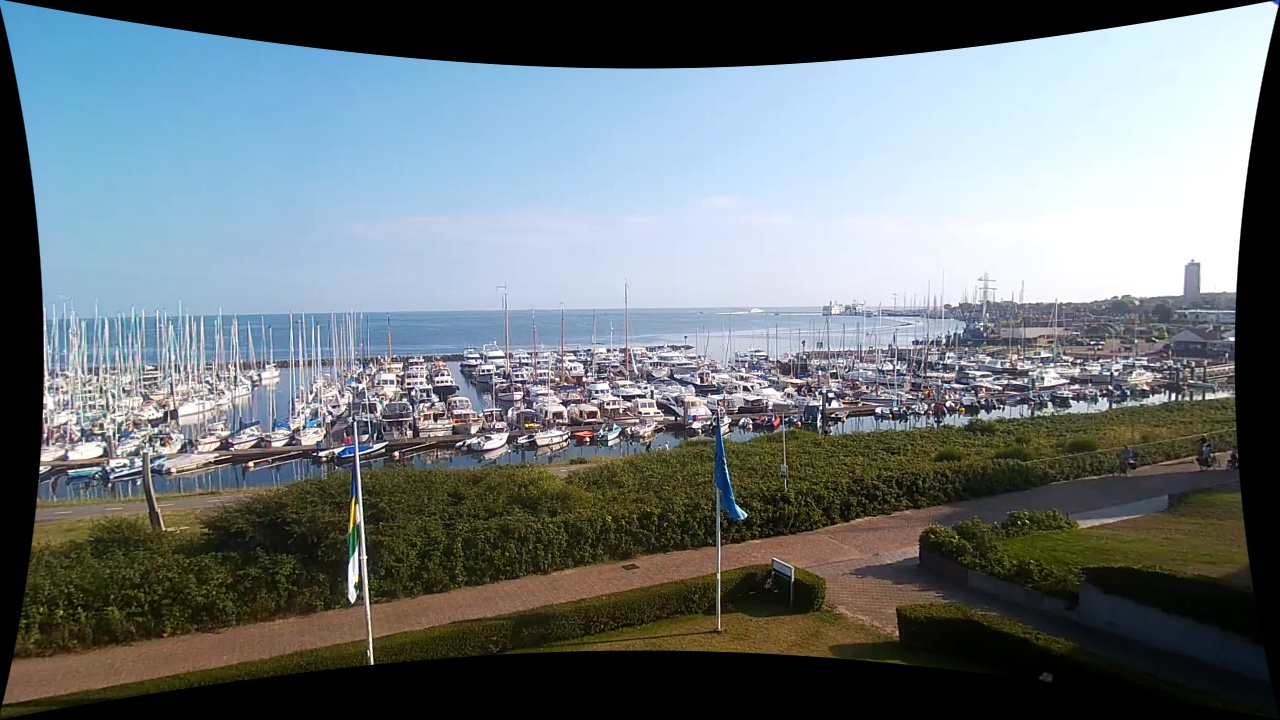}
    \includegraphics[width=.3\columnwidth]{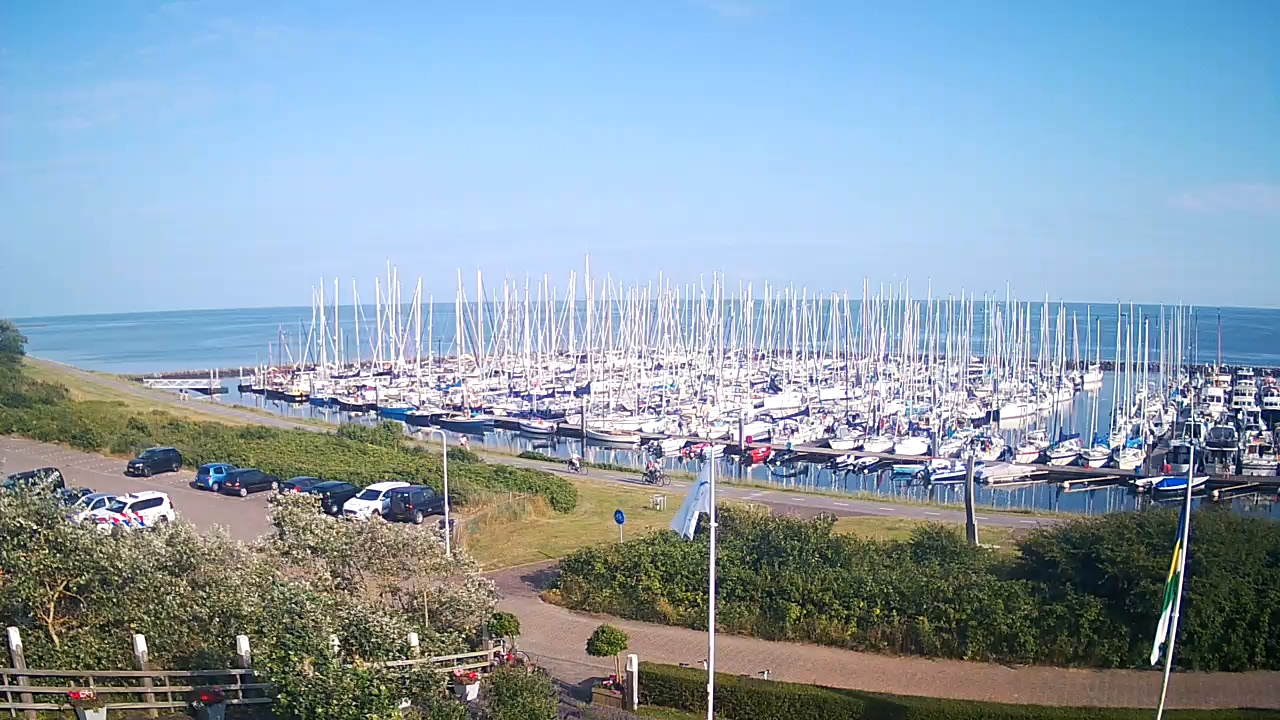}
    \includegraphics[width=.3\columnwidth]{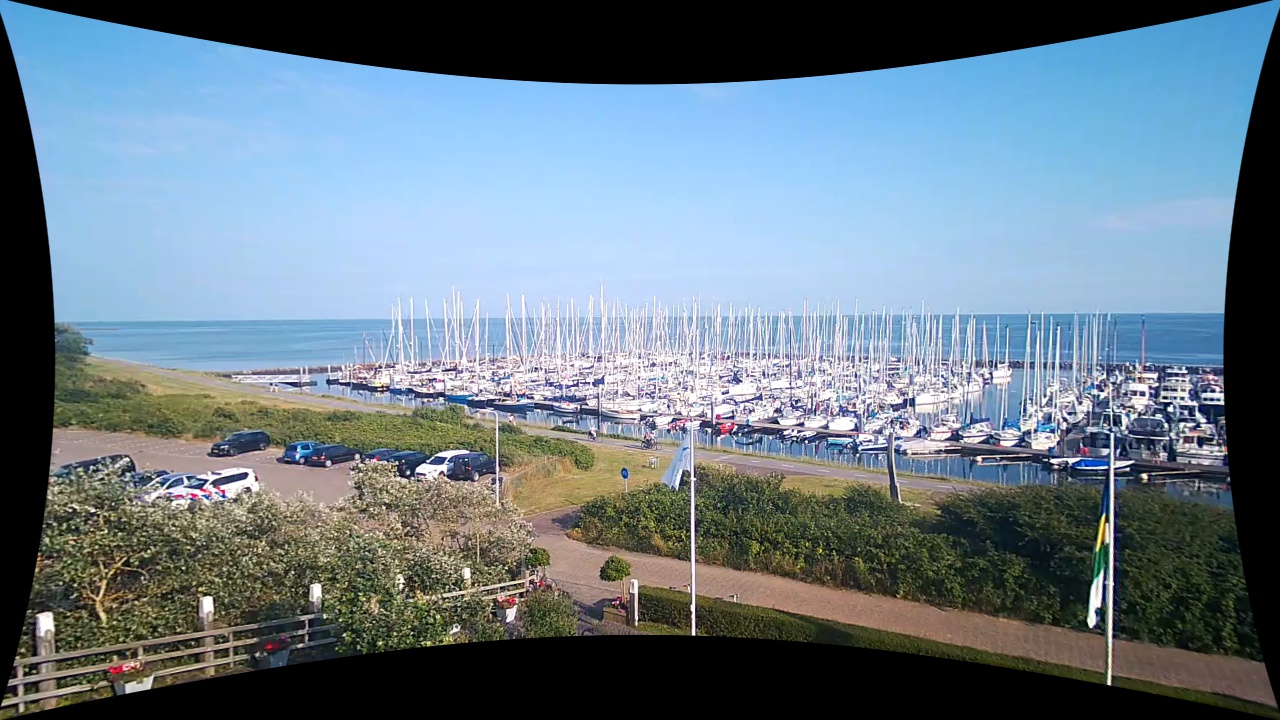}
    \includegraphics[width=.3\columnwidth]{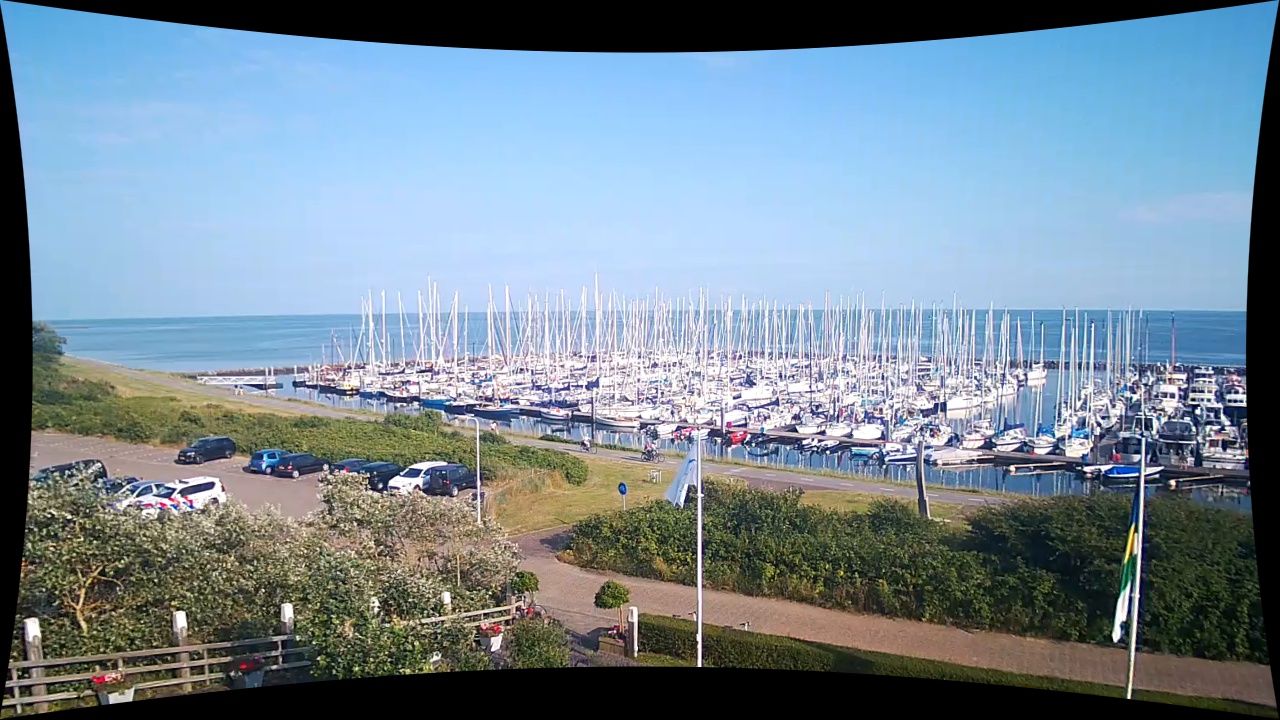}
    \caption{Sample images before and after undistortion using single and two view calibration methods on RottPor (top 2) and BranHot (bottom 2).}
    \label{fig:calib_samples}
\end{figure}

\begin{figure}[t]
    \centering
    \includegraphics[width=\columnwidth]{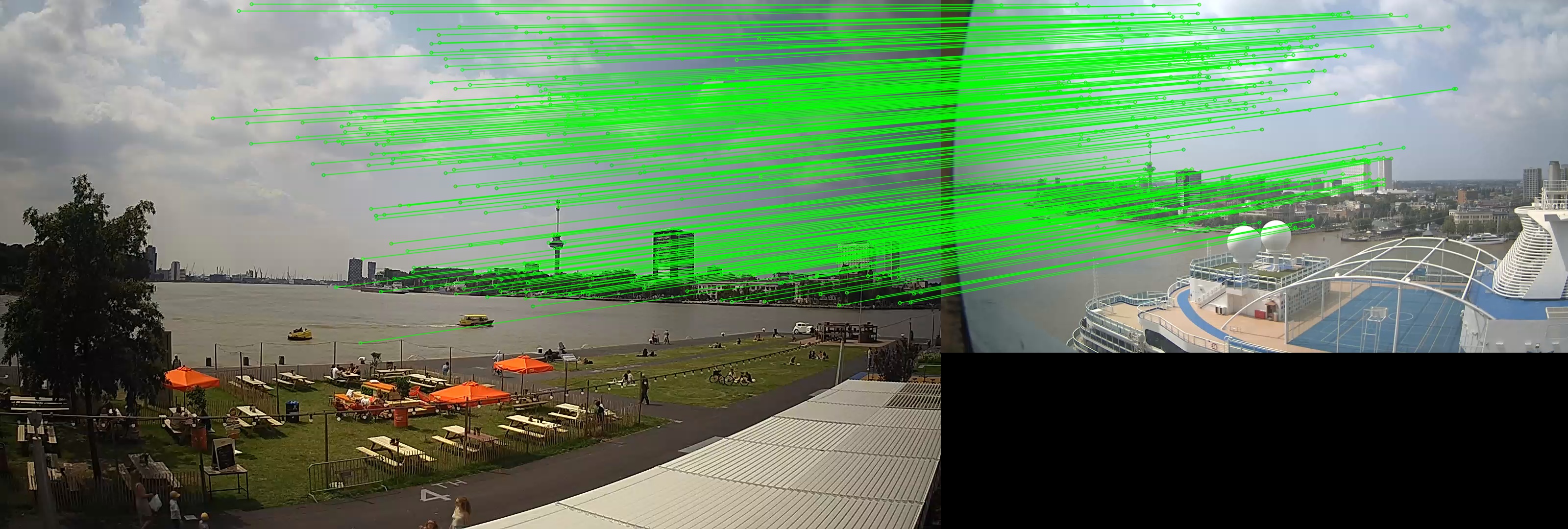}
    \caption{Matching inliers used to compute the two-view calibration of the RottPor cameras.}
    \label{fig:rott_match}
\end{figure}

\begin{figure}[t]
    \centering
    \includegraphics[width=.3\columnwidth]{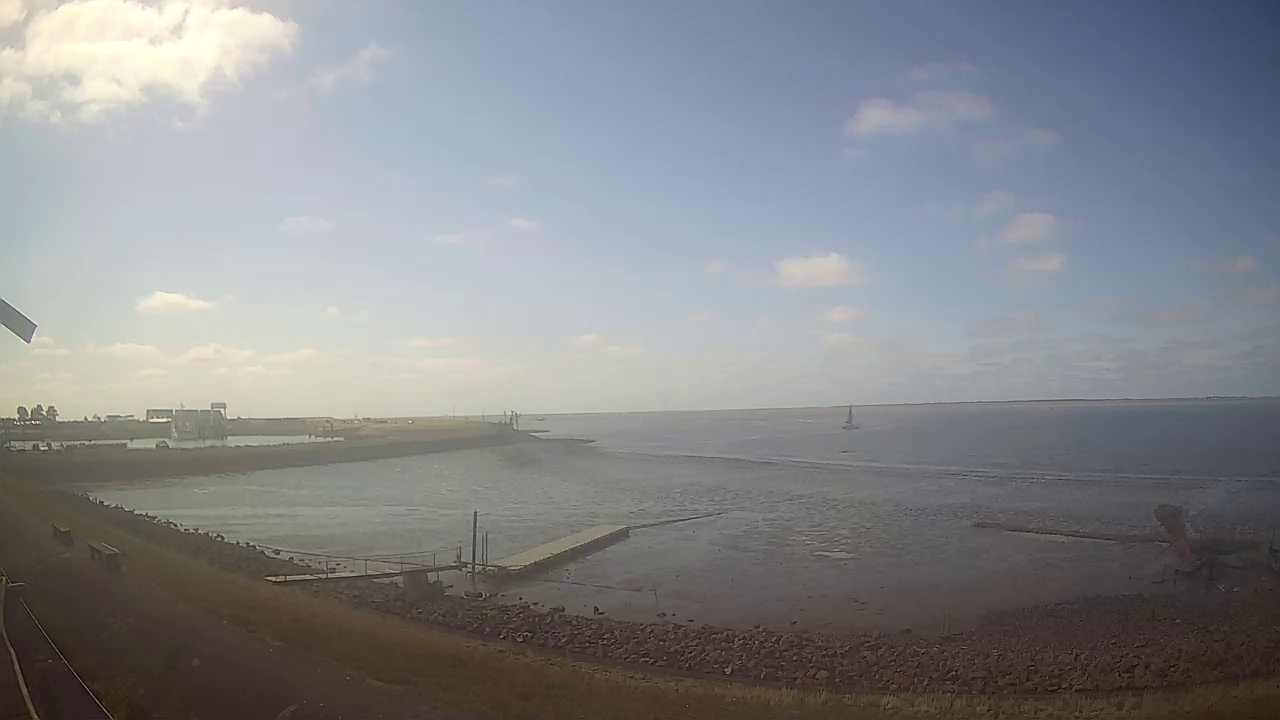}
    \includegraphics[width=.3\columnwidth]{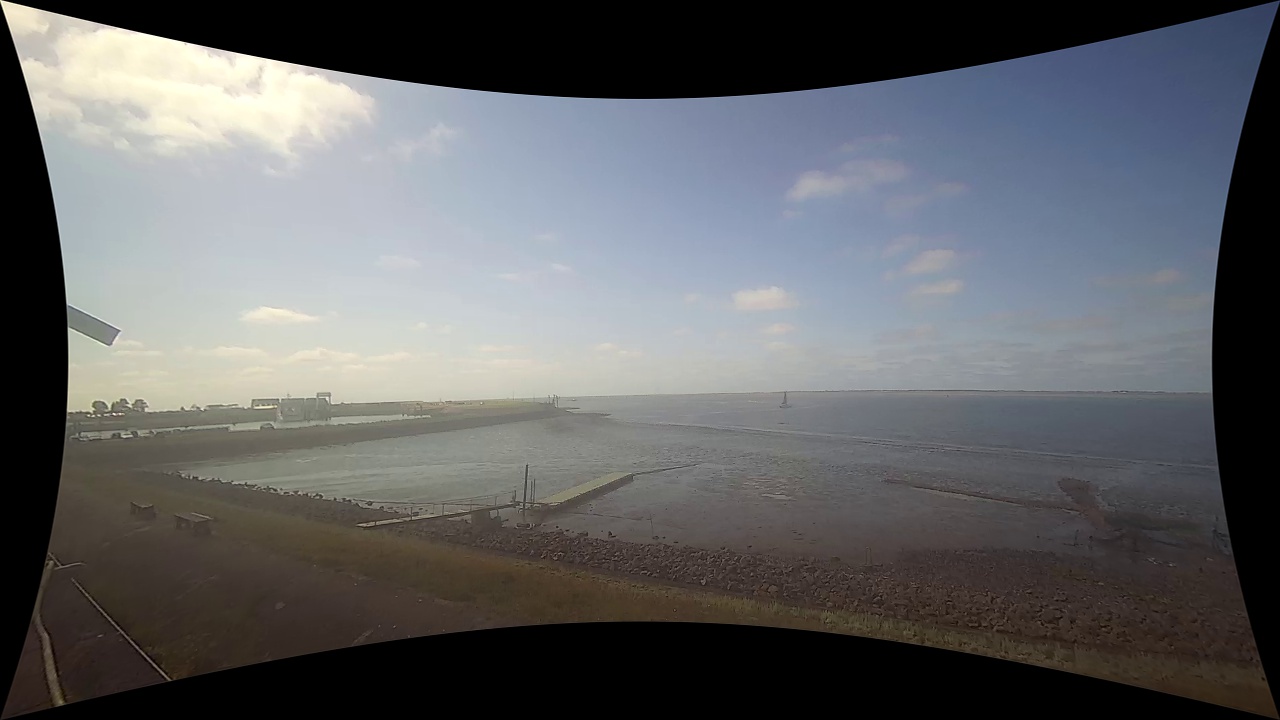}\\
    \includegraphics[width=.3\columnwidth]{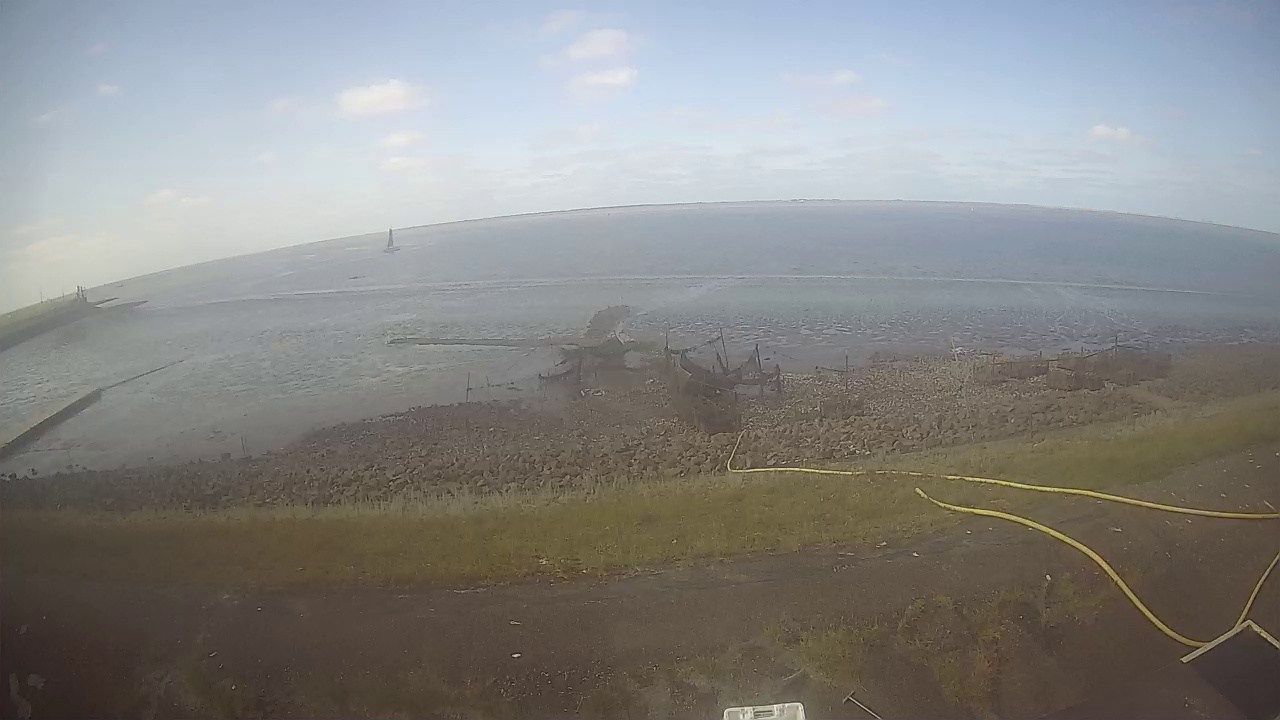}
    \includegraphics[width=.3\columnwidth]{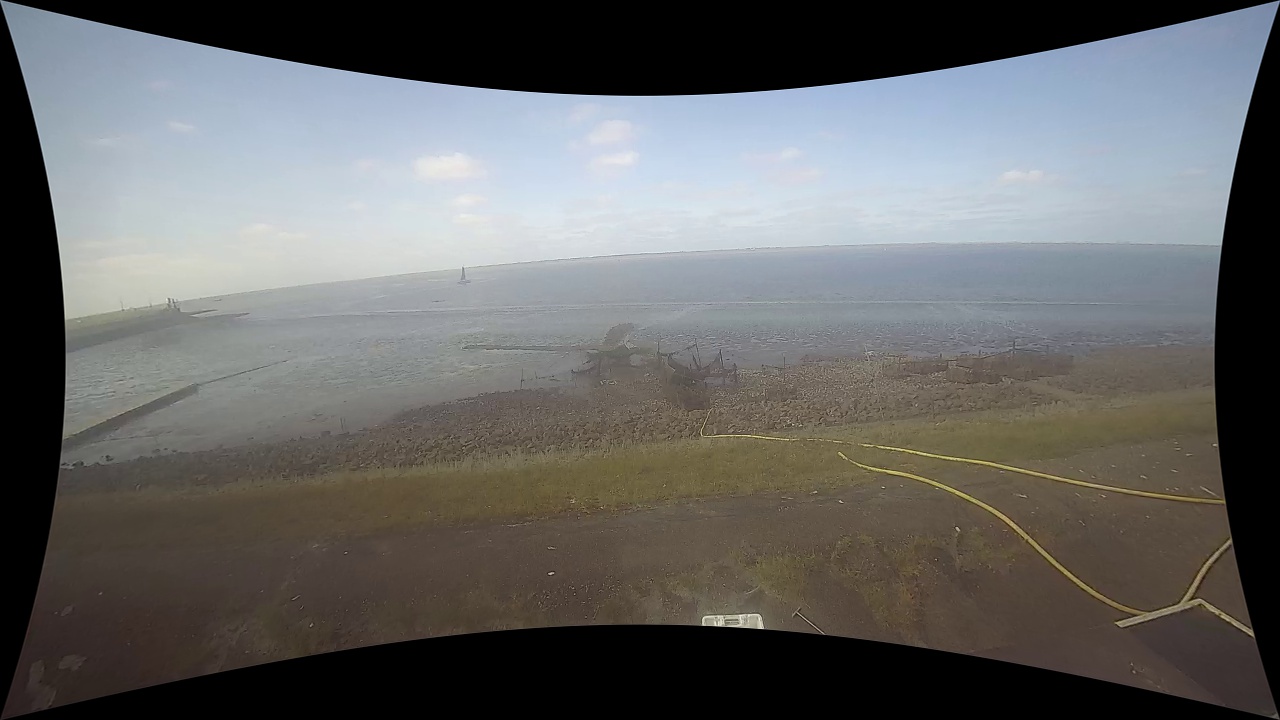}
    \caption{Example from the LauwHav dataset, where the single-view method (left column) fails but the two-view method (right column) successfully calibrates the cameras.}
    \label{fig:calib_lauwersoog}
\end{figure}

\begin{figure}[t]
    \centering
    \includegraphics[width=.3\columnwidth]{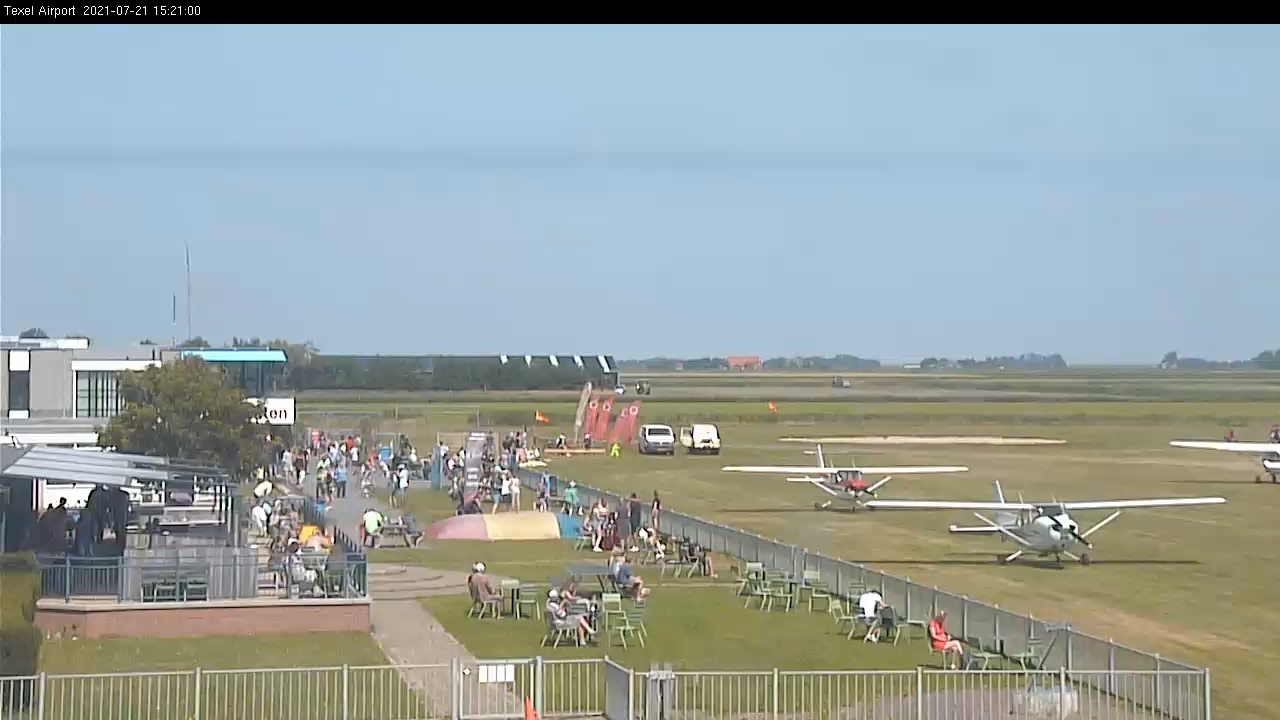}
    \includegraphics[width=.3\columnwidth]{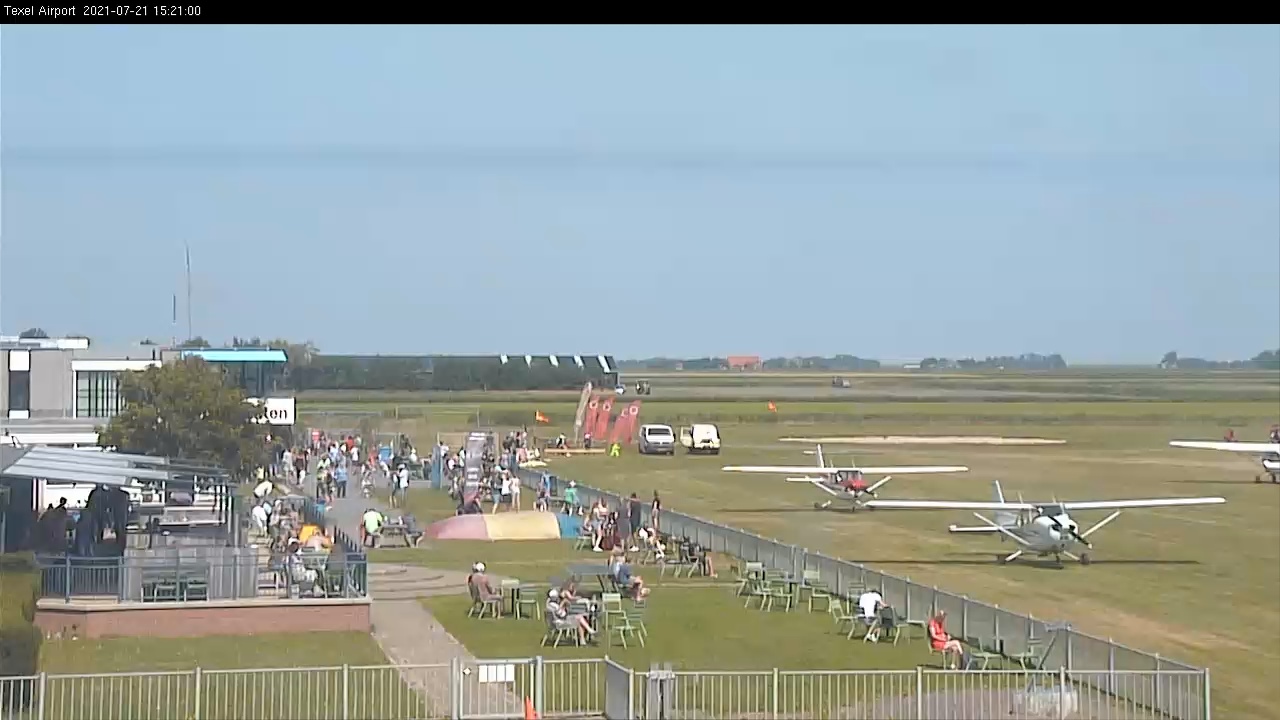}
    \includegraphics[width=.3\columnwidth]{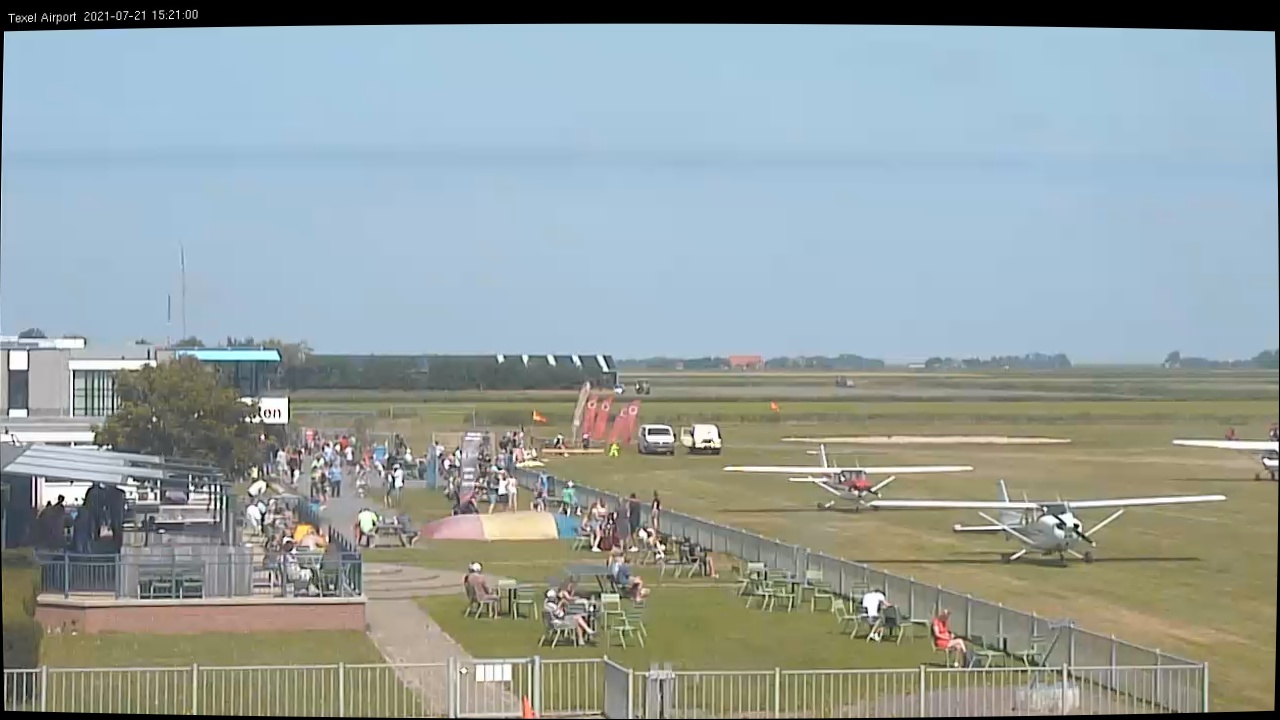}
    \includegraphics[width=.3\columnwidth]{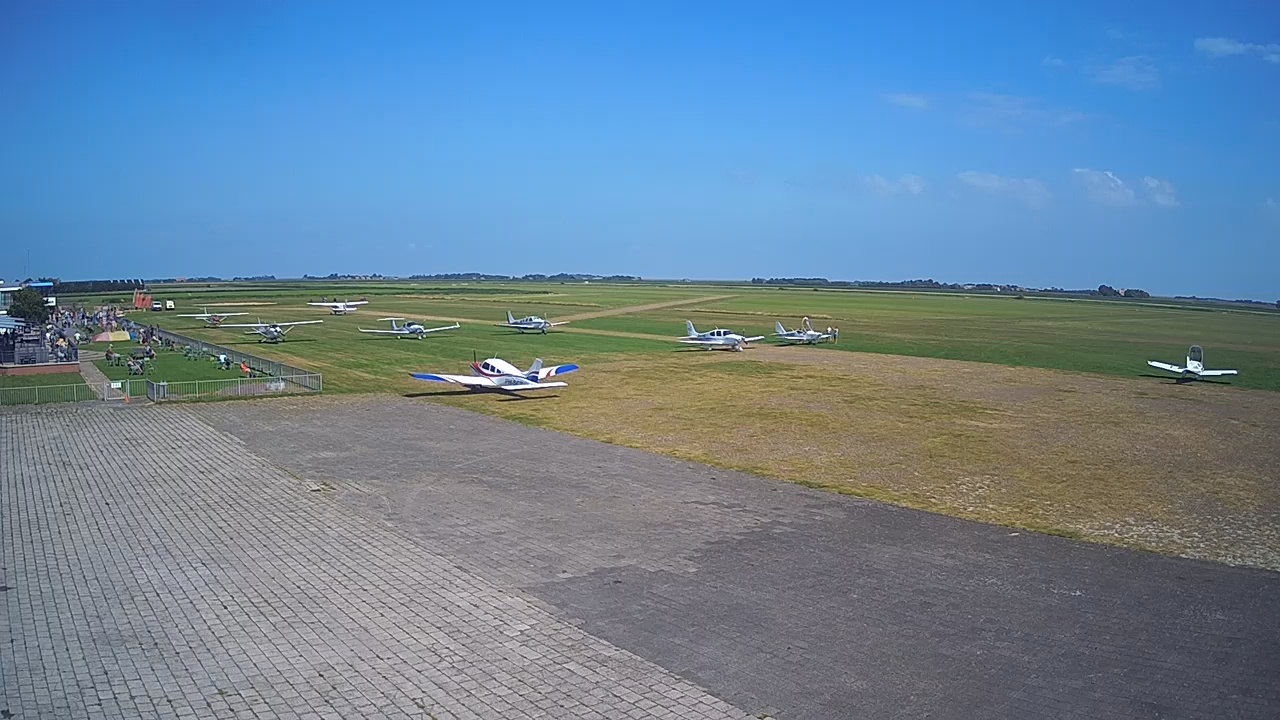}
    \includegraphics[width=.3\columnwidth]{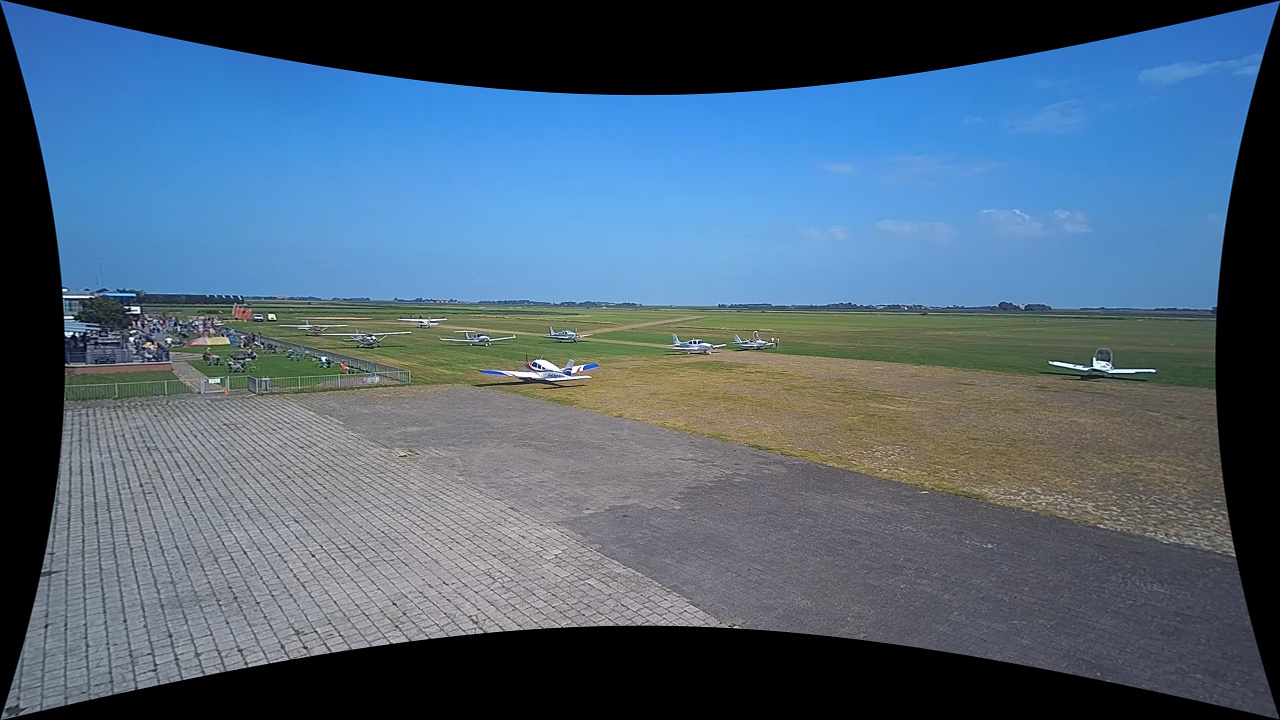}
    \includegraphics[width=.3\columnwidth]{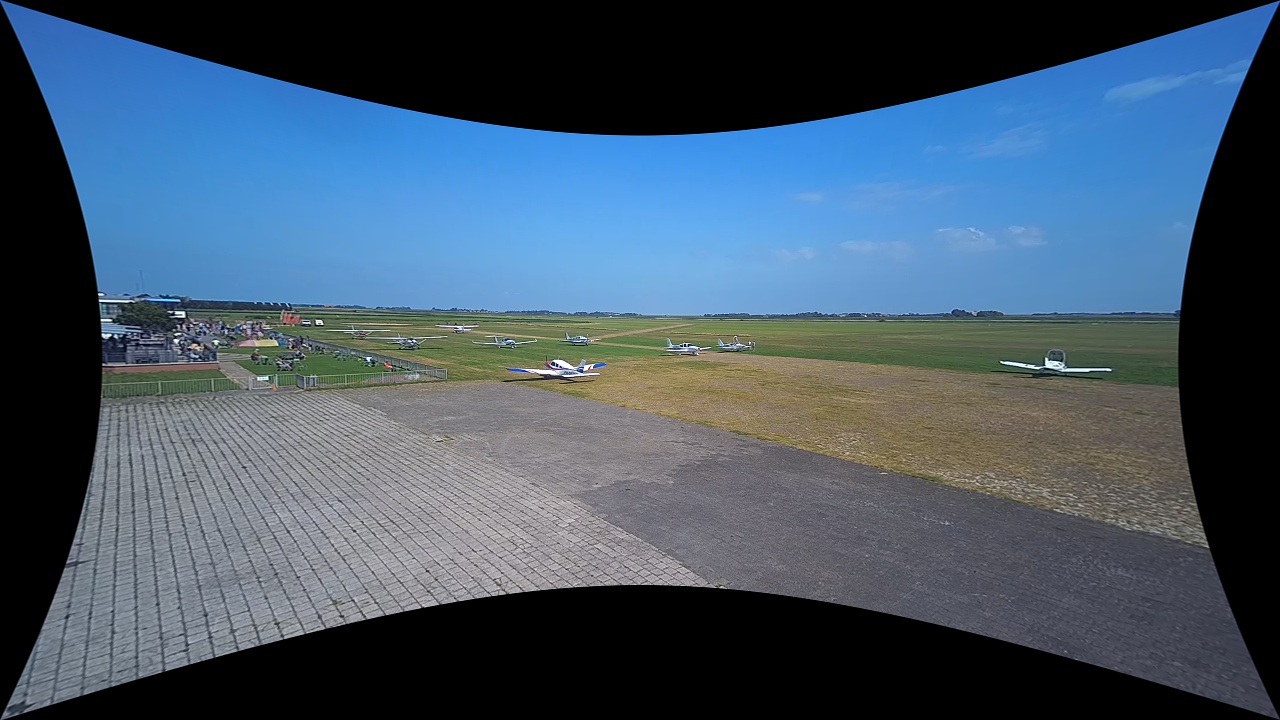}
    \caption{Example from the TexAir dataset, where the two-view method (right column) fails but the single-view method (middle column) successfully calibrates the cameras.}
    \label{fig:calib_nyc_texel}
\end{figure}

In this section, we compare the results of single and two view calibration methods, evaluate their accuracy, identify failure cases for each of them, and show how they can mutually complement each other. The estimated focal lengths and the distortion coefficients from the single- and two-view calibration can be quite different in certain scenarios (see the Appendix), yet both methods were able to recover the undistorted images successfully in most cases that we tried, as demonstrated in Figure \ref{fig:calib_samples}. An interesting situation occurs in the RottPor data, where, even with matches on  clouds (Figure \ref{fig:rott_match}), the two-view calibration is surprisingly successful. Nevertheless, the algorithms also have practical limitations. For the single-view calibration, the presence of Manhattan frames is an a-priori requirement; hence, the method breaks down if no dominant MF can be found in the scene. In addition, the quality of the calibration relies on the extraction of good features (\eg, edges) to fit the circular arcs. This can be especially challenging when the image quality is bad, such that only few features can be identified. For example, the images of the LauwHav dataset are rather hazy and there are not enough features to compute reasonable calibration parameters with the single-view method, while with the two-view method,  state-of-the-art matching algorithms like SuperPoint/SuperGlue can handle the extreme conditions rather well and are able to generate robust matches for computing the calibration (Figure \ref{fig:calib_lauwersoog}). Still, also the accuracy of the two-view method can be problematic, not only if the matching becomes unreliable, but also when the matches only cover a part of the image, such that the calibration is overfitted to that subpart (Fig. \ref{fig:calib_nyc_texel}).

After acquiring the camera parameters, the division model is converted into Brown's model \cite{duane1971close}, packaged in OpenCV library, for ease of use. The resulting parameters are applied as initial values and optimized as part of the bundle adjustment. In Table \ref{tab:calib_ba} we show the initial and optimised values of the focal length.

\begin{table}[th]
\small
    \centering
    \renewcommand{\tabcolsep}{4pt}
    \renewcommand{\arraystretch}{1}
    \begin{tabular}{c|c|c|c|c|c}
         \multicolumn{2}{c}{ } & $f^1_s$ & $f^2_s$ & $f^1_t$ & $f^2_t$ \\ \hline
\multirow{3}{*}{Croatia} & init & 1528.3 & 1070.1 & 977.0 & 652.7 \\ 
 & Opt. & 1552.8 & 1113.7 & 989.6 & 632.0 \\ 
 \hline 
\multirow{3}{*}{NewYork} & init & 1417.7 & 1083.6 & 1244.2 & 405.0 \\ 
 & Opt. & 1390.0 & 965.0 & 1394.5 & 432.4 \\ 
 \hline 
\multirow{3}{*}{TimeSq} & init & 782.2 & 888.2 & 946.8 & 885.81 \\ 
 & Opt. & 672.9 & 1289.5 & 741.6 & 1224.6 \\ 
 \hline 
\multirow{3}{*}{AalsHav} & init & 3010.7 & 1111.5 & 175.3 & 368.0 \\ 
 & Opt. & 3154.1 & 1099.9 & 190.8 & 340.9 \\ 
 \hline 
\multirow{3}{*}{BranHot} & init & 1077.0 & 1909.3 & 1172.5 & 947.8 \\ 
 & Opt. & 1070.4 & 1960.5 & 1159.3 & 984.5 \\ 
 \hline 
\multirow{3}{*}{BranPor} & init & 1352.0 & 1397.6 & 2067.1 & 1921.0 \\ 
 & Opt. & 1301.2 & 1419.3 & 2254.3 & 1979.4 \\ 
 \hline 
\multirow{3}{*}{RottPor} & init & 1573.8 & 1008.3 & 1268.4 & 1021.7 \\ 
 & Opt. & 1572.5 & 944.4 & 1316.7 & 1037.4 \\ 
 \hline 
\multirow{3}{*}{TexAir} & init & 1395.0 & 1924.7 & 822.2 & 16651.0 \\ 
 & Opt. & 1445.9 & 1733.2 & 955.4 & 15832.1 \\ 
 \hline 
\multirow{3}{*}{LauwHav} & init & - & - & 1063.9 & 1157.7 \\ 
 & Opt. & - & - & 1047.8 & 1181.9 \\ 
 \hline 

    \end{tabular}
    \caption{Resulting focal lengths before and after BA.}
    \label{tab:calib_ba}
\end{table}

\subsection{Static and dynamic scene reconstruction}
After the necessary preprocessing steps, we are ready for the final 3D reconstruction. To disentangle the impact of each feature type on the reconstruction, and to investigate the benefit of combining static and dynamic features,, we design three experiments: static-only ($SO$); static-dynamic-unsync ($SD_{un}$), using both static and dynamic features but without optimising the initial synchronisation parameters during BA; and static-dynamic-sync ($SD_{sc}$), where we using both static and dynamic features, and optimise the camera synchronisation in a data-driven manner during BA.

To evaluate accuracy of the reconstruction, 20 manual correspondences (control points) are selected for each camera pair such that they cover as large area of the image as possible. Reprojection errors are computed for those points, and for the reconstructed object trajectories.

\begin{figure*}
    \centering
    \includegraphics[width=.33\columnwidth]{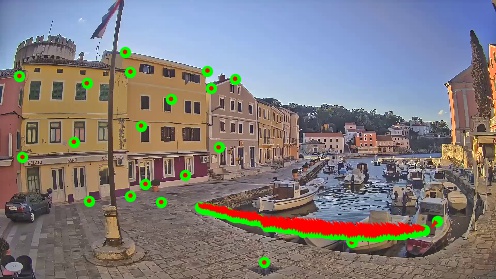}
    \includegraphics[width=.33\columnwidth]{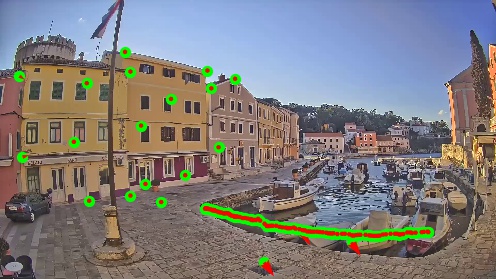}
    \includegraphics[width=.33\columnwidth]{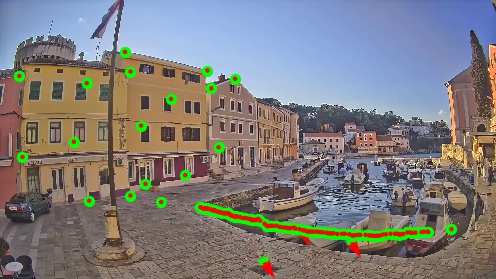}
    \includegraphics[width=.33\columnwidth]{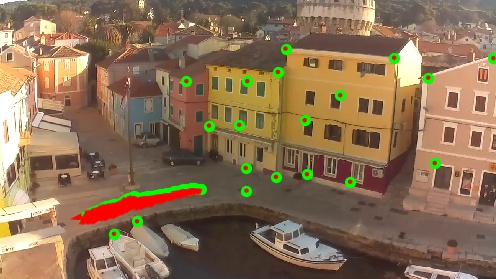}
    \includegraphics[width=.33\columnwidth]{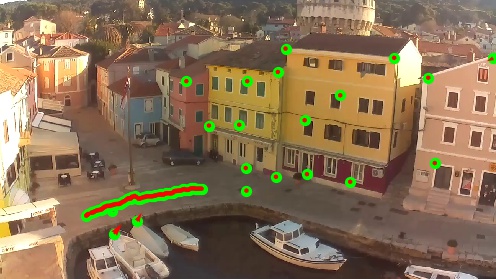}
    \includegraphics[width=.33\columnwidth]{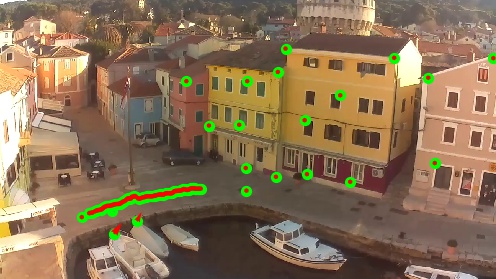}
    \includegraphics[width=.33\columnwidth]{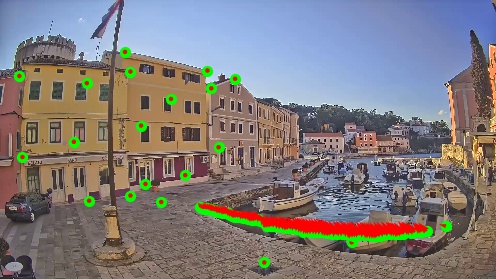}
    \includegraphics[width=.33\columnwidth]{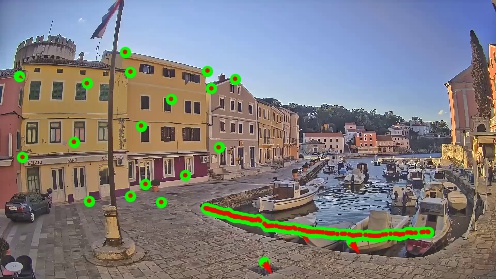}
    \includegraphics[width=.33\columnwidth]{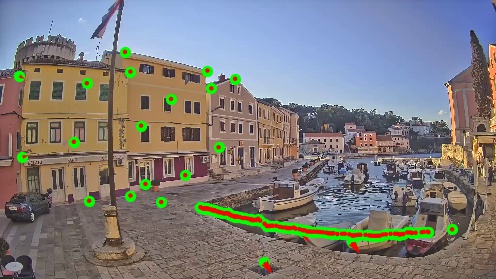}
    \includegraphics[width=.33\columnwidth]{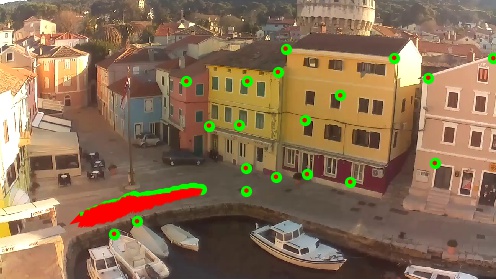}
    \includegraphics[width=.33\columnwidth]{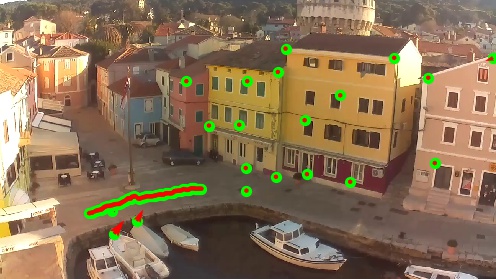}
    \includegraphics[width=.33\columnwidth]{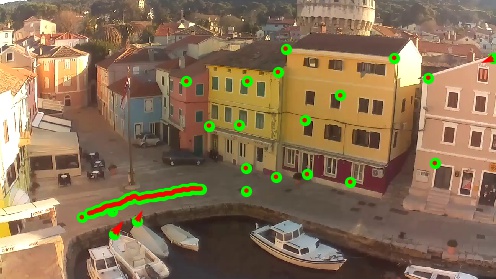}
    \caption{The reconstruction results of Croatia data in the static-only (left), static-dynamic-sync (middle) and static-dynamic-unsync (right) setting initialized with single- (top) and two-view (bottom) calibration. The red arrows indicate the reprojection error magnitudes.}
    \label{fig:recon_croatia}
\end{figure*}

\begin{figure*}[!htbp]
    \centering
    \includegraphics[width=.33\columnwidth]{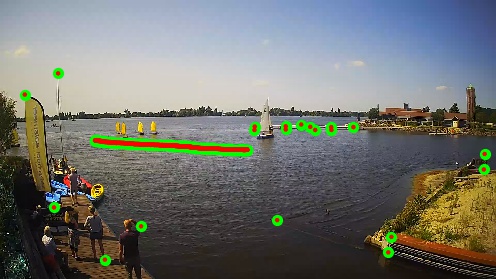}
    \includegraphics[width=.33\columnwidth]{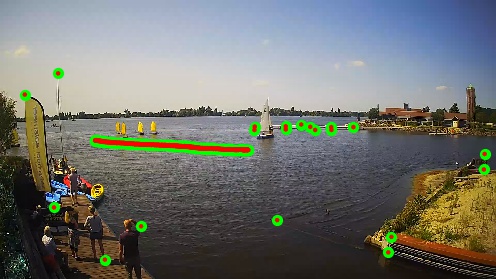}
    \includegraphics[width=.33\columnwidth]{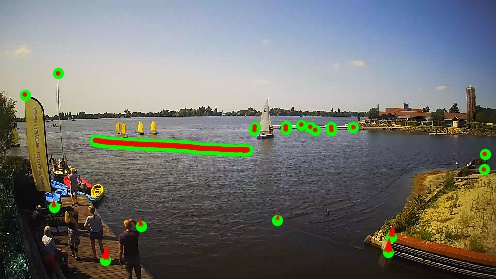}
    \includegraphics[width=.33\columnwidth]{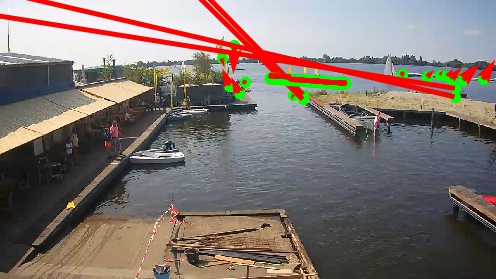}
    \includegraphics[width=.33\columnwidth]{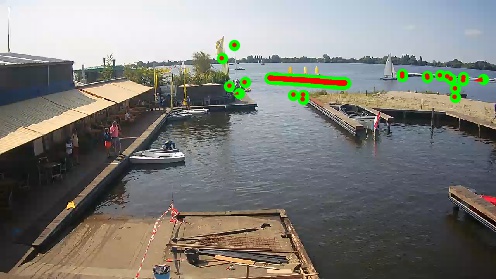}
    \includegraphics[width=.33\columnwidth]{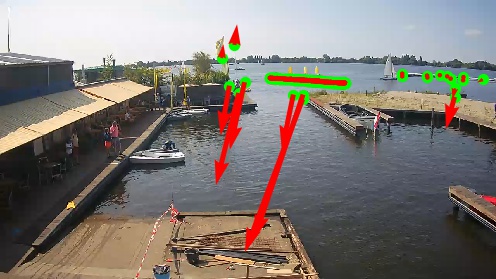}
    \includegraphics[width=.33\columnwidth]{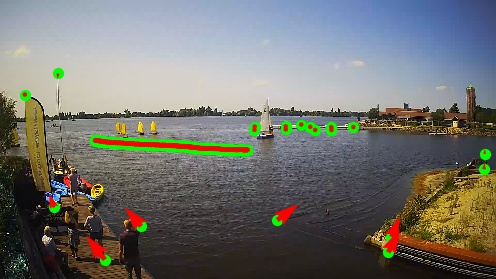}
    \includegraphics[width=.33\columnwidth]{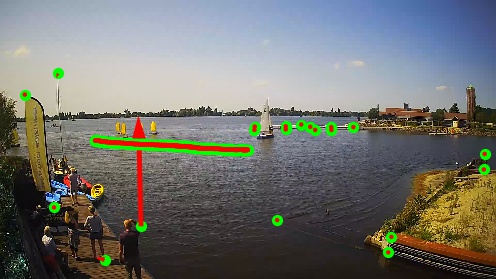}
    \includegraphics[width=.33\columnwidth]{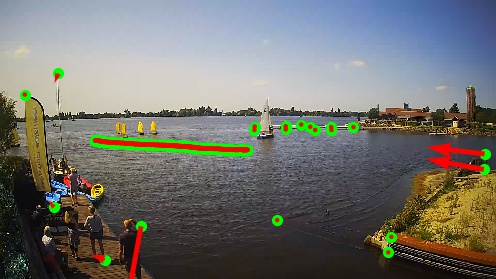}
    \includegraphics[width=.33\columnwidth]{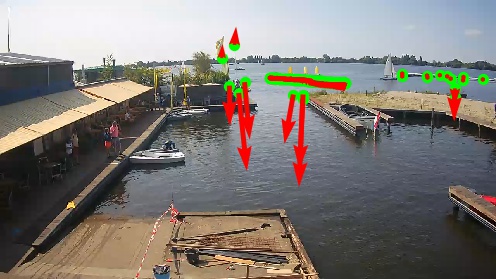}
    \includegraphics[width=.33\columnwidth]{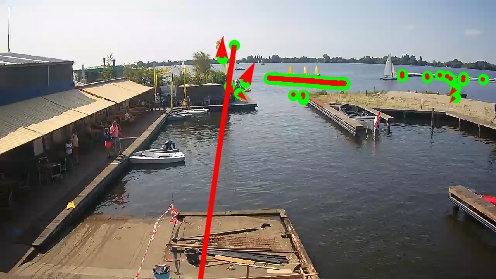}
    \includegraphics[width=.33\columnwidth]{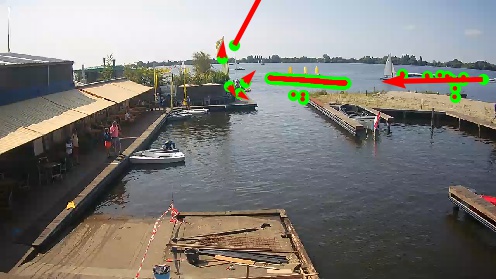}
    \caption{An example of 3D reconstruction failure case on AalsHav. The top row shows the result initialized with single-view calibration, the bottom row the one with two-view calibration. Red arrows indicate the magnitude of the reprojection errors.}
    \label{fig:recon_aalsmeer}
\end{figure*}

Table \ref{tab:recon_results} summarizes the reprojection errors of the annotated control points $e_{cp}$ and of the reconstructed trajectories $e_{traj}$ under different experiment settings, and with two different initial estimates of the calibration parameters. Based on quantitative and qualitative examination of the reconstruction, we find that our pipeline produces reasonable reconstructions for 8 out of 9 datasets (\eg, Figure \ref{fig:recon_croatia}); while largely failing on the last dataset (Figure \ref{fig:recon_aalsmeer}). Although preliminary, some insights can be inferred by comparing the results of different setups. In the $SO$ setup, the control point errors $e_{cp}$ are comparatively small, whereas the trajectory errors $e_{traj}$ are often large, especially when the matched static features are clustered in some parts of the images (like building facades), such that the reconstruction is overfitted to that particular region. With additional information introduced by the dynamic features in the $SD_{sc}$ and $SD_{un}$ setups, $e_{traj}$ dramatically drops to much smaller values, while the $e_{cp}$ stay at a similar level or sometimes increase a little, indicating that the combination of both static and dynamic features constrains the overall geometry better. Moreover, by allowing the pipeline to refine the synchronisation between cameras, $SD_{sc}$ achieves a small, but consistent improvement of the trajectory error $e_{traj}$. We note that our evaluation may not be representative \wrt the effect of synchronisation, because the time offsets between the video streams in our dataset are rather small compared to the object velocities, whereas one would expect $SD_{sc}$ to have a bigger impact for large de-synchronisation and/or fast-moving objects.

\begin{table*}
    \centering
    \renewcommand{\tabcolsep}{4pt}
    \renewcommand{\arraystretch}{1}
    \begin{tabular}{c|c|c|c|c|c|c|c|c|c|c|c|c}
          & \multicolumn{3}{c|}{$e^1_{cp}$}  & \multicolumn{3}{c|}{$e^1_{traj}$}  & \multicolumn{3}{c|}{$e^2_{cp}$}    & \multicolumn{3}{c}{$e^2_{traj}$} \\ \cline{2-13} 
          & $SO$     & $SD_{sc}$ & $SD_{un}$      & $SO$     & $SD_{sc}$ & $SD_{un}$        & $SO$     & $SD_{sc}$ & $SD_{un}$         & $SO$     & $SD_{sc}$ & $SD_{un}$     \\
         \hline
\rowcolor{green!50}
Croatia & 2.9 & 12.7 & 11.9 & 66.3 & 2.6 & 2.9 & 2.3 & 7.0 & 7.0 & 54.4 & 1.5 & 1.7 \\ 
\hline 
\rowcolor{green!50}
NewYork & 0.2 & 0.2 & 0.2 & 1.8 & 0.1 & 0.1 & 0.6 & 0.6 & 0.7 & 5.0 & 0.4 & 0.4 \\ 
\hline 
\rowcolor{green!50}
TimeSq & 2.7 & 0.9 & 0.8 & 22.5 & 0.4 & 0.5 & 3.8 & 0.7 & 0.5 & 34.5 & 0.4 & 0.5 \\ 
\hline 
\rowcolor{red!50}
AalsHav & 34.7 & 29.0 & 32.7 & 7.2 & 0.3 & 0.6 & 54.8 & 41.5 & 20.2 & 12.7 & 0.6 & 0.8 \\ 
\hline 
\rowcolor{green!50}
BranHot & 0.3 & 0.4 & 0.5 & 1.0 & 0.5 & 0.5 & 0.3 & 0.4 & 0.5 & 1.1 & 0.5 & 0.6 \\ 
\hline 
\rowcolor{green!50}
BranPor & 1.0 & 2.3 & 2.2 & 0.7 & 0.2 & 0.2 & 1.1 & 2.3 & 2.2 & 0.9 & 0.3 & 0.3 \\ 
\hline 
\rowcolor{green!50}
RottPor  & 1.3 & 1.2 & 1.2 & 2.1 & 0.2 & 0.2 & 1.8 & 1.7 & 1.6 & 3.2 & 0.3 & 0.3 \\ 
\hline 
\rowcolor{green!50}
TexAir & 2.0 & 2.0 & 2.4 & 5.9 & 1.1 & 1.3 & 0.0 & 0.2 & 0.2 & 0.1 & 0.1 & 0.1 \\ 
\hline 
\rowcolor{green!50}
LauwHav  & 0.5 & 0.5 & 0.5 & 0.7 & 0.3 & 0.3 & 0.5 & 0.6 & 0.6 & 0.6 & 0.3 & 0.3 \\ 
 \hline 
    \end{tabular}
    \caption{Reprojection errors of static control points and of trajectories in both cameras under the static-only, static-dynamic-sync, and static-dynamic-unsync settings, initialized with two-view calibration. Success cases (average error \textless7 pixels) are marked in green, failure case marked in red. }
    \label{tab:recon_results}
\end{table*}

In most cases, initialising the camera parameters with the single-view or two-view calibration does not make a big difference for the final reconstruction. For the success cases, the reprojection errors of both approaches often end up being similar, indicating that the pipeline is relatively robust against fluctuations of the initial values within reasonable bounds. For the failure cases, however, the algorithm would likely fail with both calibration methods: the crucial bottleneck, as so often in 3D vision, is to find a sufficient set of correct correspondences, and the factors that prevent this from happening, like extreme viewpoint differences or lack of texture, equally affect both initialisation methods. 

As expected, there are cases where the pipeline fails to reconstruct the 3D scen geometry. Figure \ref{fig:recon_aalsmeer} illustrates the most challenging scene in our collection, where the viewpoints of the two cameras are wildly different and the overlap region between the views is small and virtually featureless. In this example, both the static and the dynamic features that the respective algorithms manage to extract are distributed along a narrow horizontal band in the images with very small vertical extent, thus rendering the pose estimation ill-posed, which in turn leads to large reprojection errors at the control points.


\section{Conclusion} \label{sec:conclusion}
We have found that 3D reconstruction from public webcams, perhaps surprisingly, differs from conventional SfM in a number of crucial details, such that conventional SfM pipelines fail despite their considerable maturity. Starting from that observation, we have shown that 3D reconstruction from public webcams is indeed possible, if appropriate cutting edge computer vision methods are added to the pipeline, including in particular camera calibration from only one or two images, deep feature point detection and matching to handle large viewpoint changes and challenging, in-the-wild imaging conditions, and dedicated mechanisms to track and reconstruct moving objects. With those extensions, we have obtained convincing reconstructions of camera poses, scene structure and 3D object trajectories. We have also introduced a new dataset of wide-baseline stereo videos from public webcams, with varying degrees of difficulty. Our code and data will be made available online to encourage and support further research on the topic.




{\small
\bibliographystyle{ieee_fullname}
\bibliography{egbib}
}

\newpage
\appendix
\onecolumn

\section{Appendix}
   Here we present the full sets of results for the feature matching, camera calibration with the single- and two-view approach, comparison of results obtained by RANSAC and MAGSAC, as well as the 3D reconstruction.
\subsection{Feature matching}
\begin{figure*}[!htbp]
    \centering
    \begin{tabular}{ccc}
          & SIFT & SuperPoint + SuperGlue \\
         \rotname{Croatia} & \includegraphics[width=.47\columnwidth]{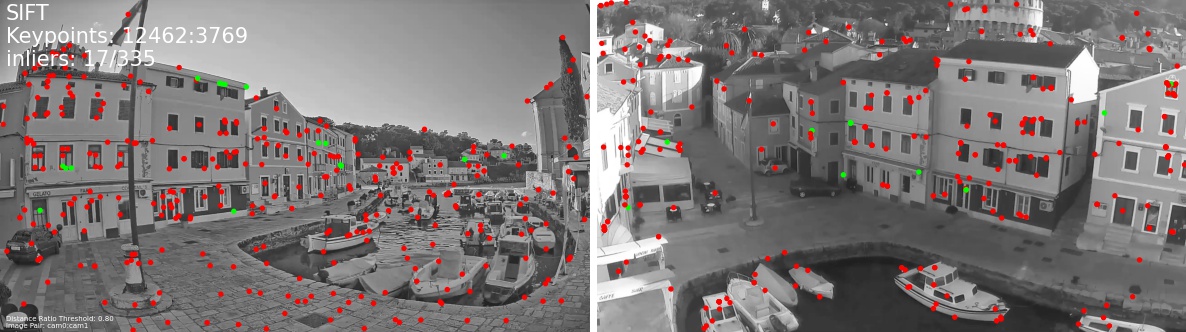} & \includegraphics[width=.47\columnwidth]{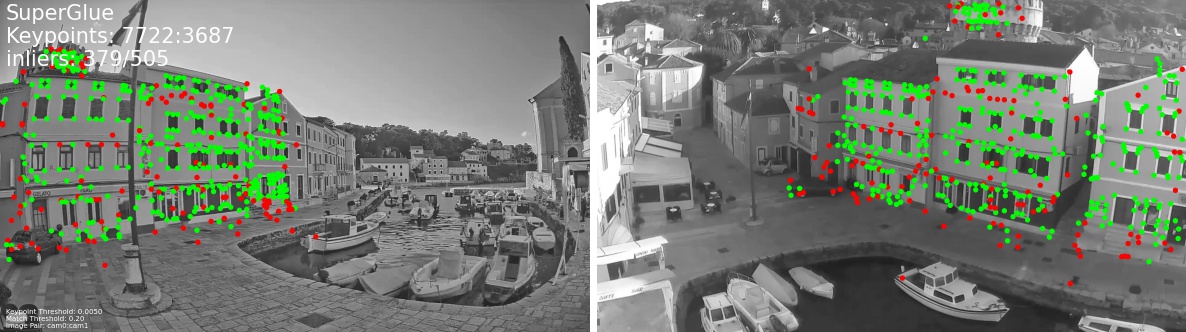} \\
         \rotname{NewYork} & \includegraphics[width=.47\columnwidth]{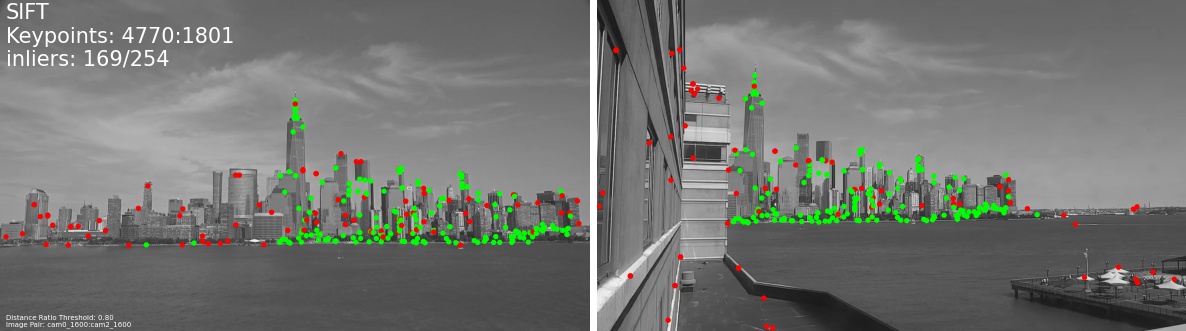} & \includegraphics[width=.47\columnwidth]{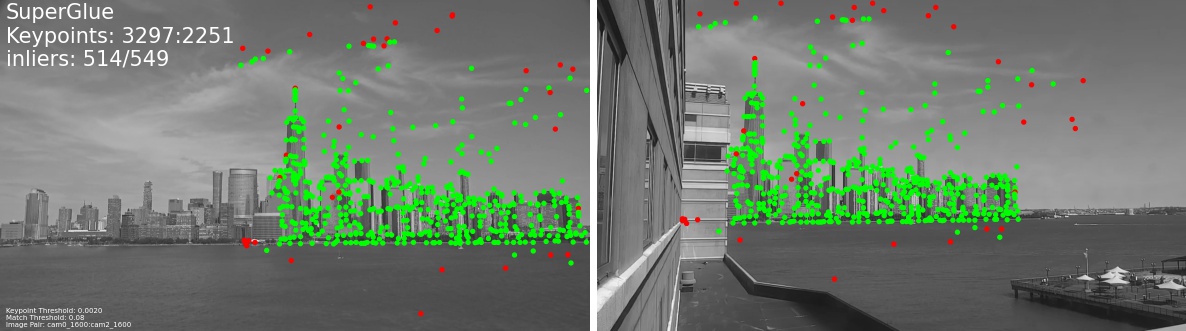} \\
         \rotname{TimeSq} & \includegraphics[width=.47\columnwidth]{images/img_match/inliers_no_lines_large/time_square_datasets_cam0_cam1_sift_inlier.jpg} & \includegraphics[width=.47\columnwidth]{images/img_match/inliers_no_lines_large/time_square_datasets_cam0_cam1_inlier.jpg} \\
         \rotname{AalsHav} & \includegraphics[width=.47\columnwidth]{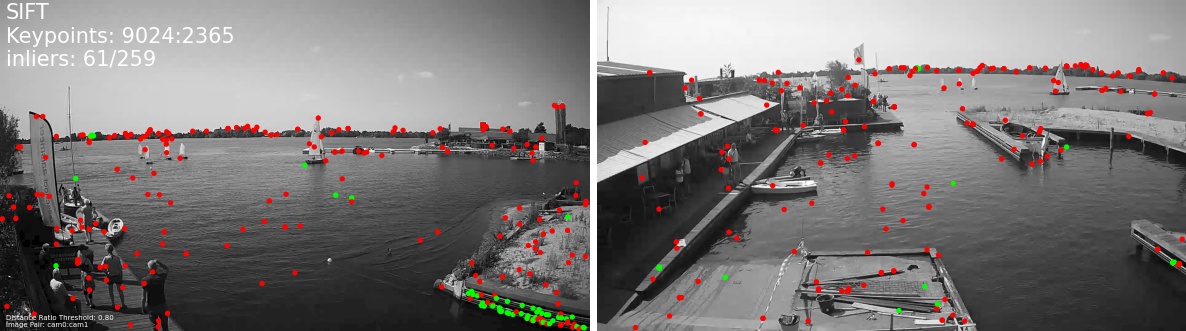} & \includegraphics[width=.47\columnwidth]{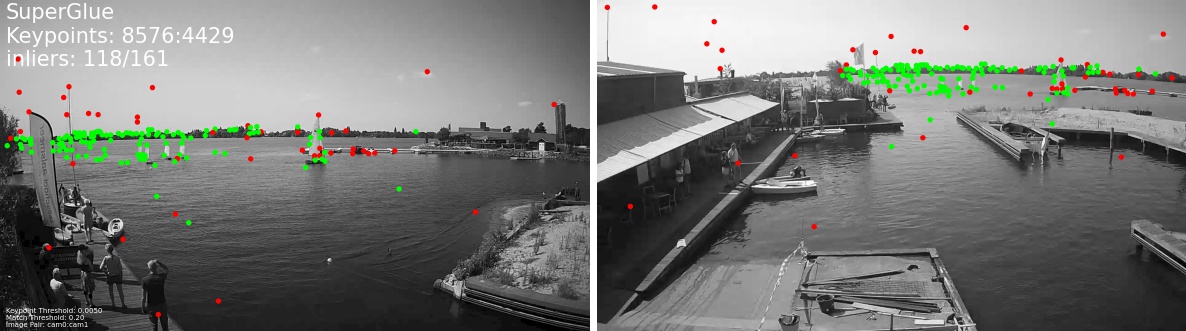} \\
         \rotname{BranHot} & \includegraphics[width=.47\columnwidth]{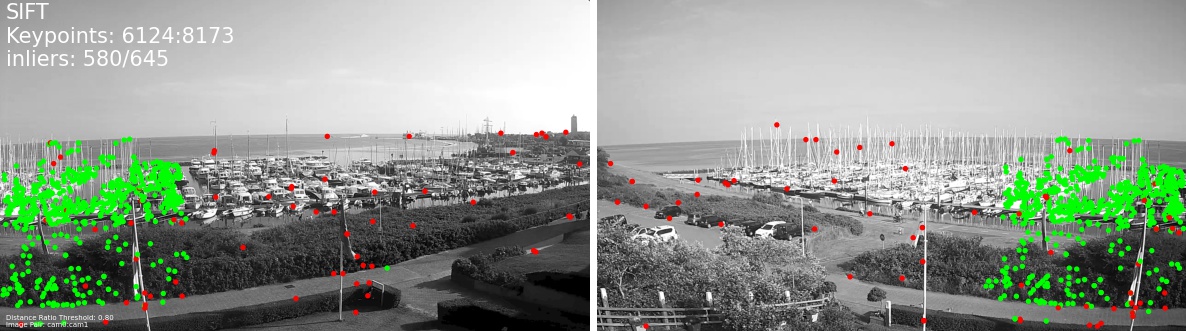} & \includegraphics[width=.47\columnwidth]{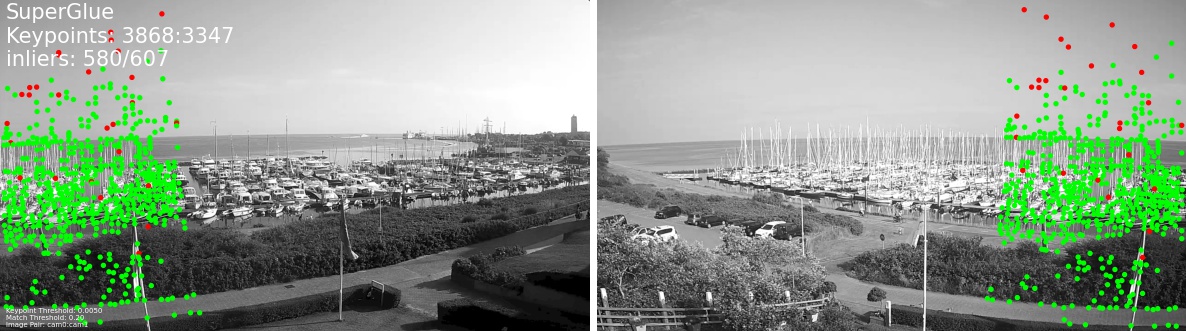} \\
         \rotname{BranPor} & \includegraphics[width=.47\columnwidth]{images/img_match/inliers_no_lines_large/brandaris_port_datasets_cam0_cam1_sift_inlier.jpg} & \includegraphics[width=.47\columnwidth]{images/img_match/inliers_no_lines_large/brandaris_port_datasets_cam0_cam1_inlier.jpg} \\
         \rotname{RottPor} & \includegraphics[width=.47\columnwidth]{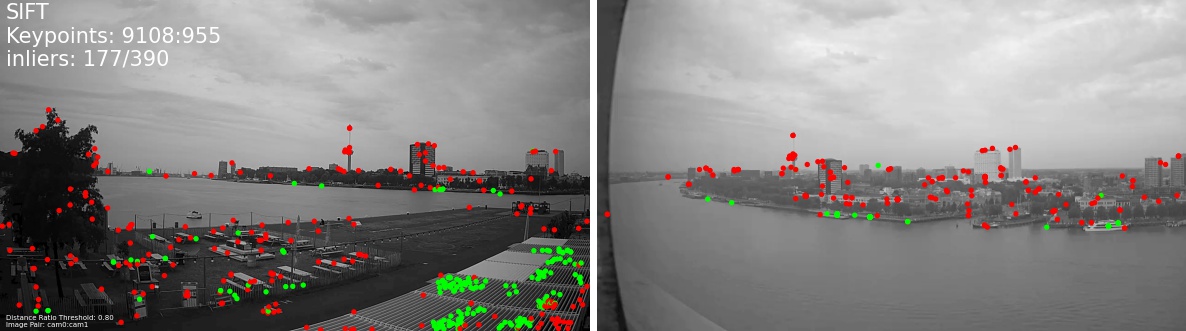} & \includegraphics[width=.47\columnwidth]{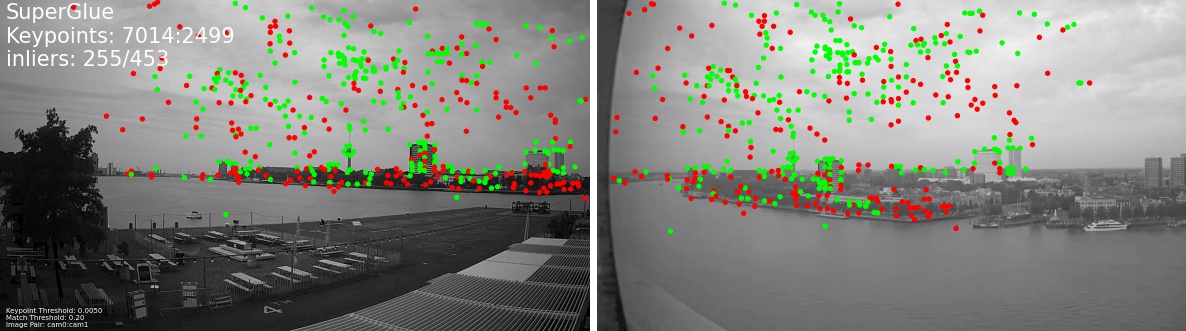} \\
         \rotname{TexAir} & \includegraphics[width=.47\columnwidth]{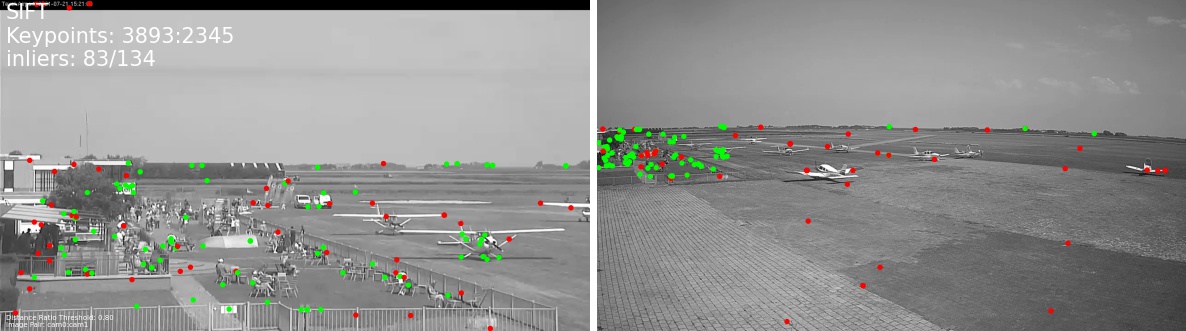} & \includegraphics[width=.47\columnwidth]{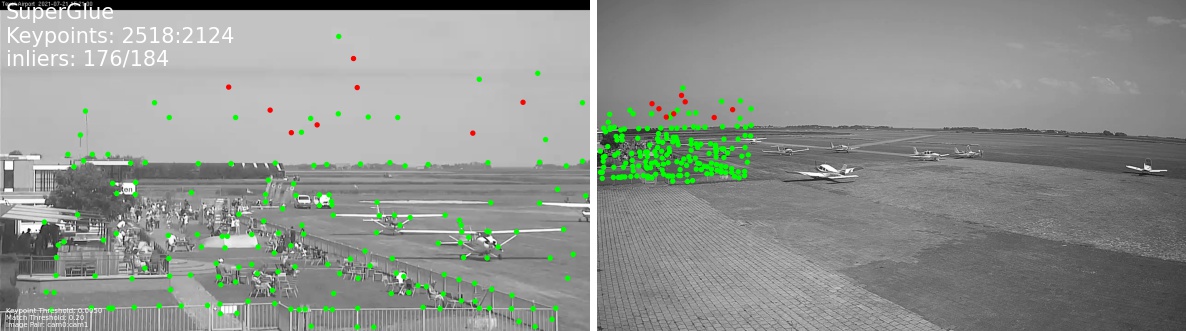} \\
         \rotname{LauwHav} & \includegraphics[width=.47\columnwidth]{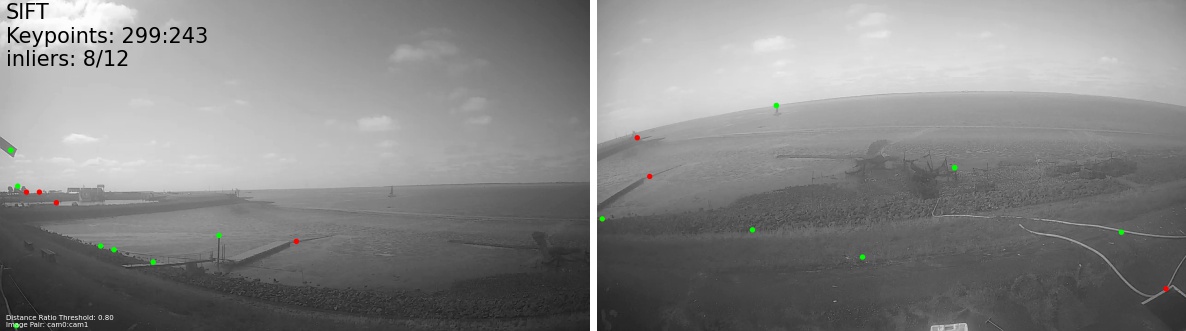} & \includegraphics[width=.47\columnwidth]{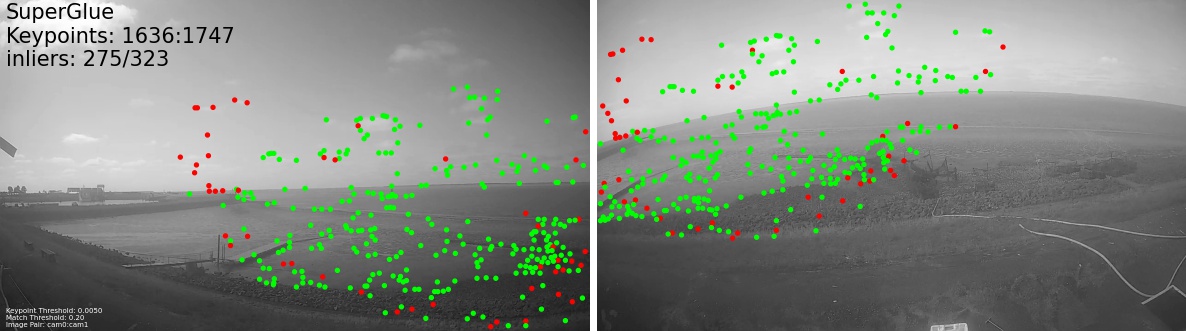}
    \end{tabular}
    \caption{The matching results with SIFT vs. SuperPoint + SuperGlue. The inliers are computed from the undistorted features and fundamental matrix with MAGSAC. The inlier matches are in \textcolor{green}{green}, and outlier matches in \textcolor{red}{red}.}
    \label{fig:feat_match_full}
\end{figure*}

We evaluated the matching results of SuperPoint + SuperGlue and compared it with matches generated using exhaustive SIFT matching. In Figure~\ref{fig:feat_match_full} the tentative matches are shown in red and the inliers computed from the tentative matches using MAGSAC are shown in green. In most cases, SuperPoint + SuperGlue provides significantly more matches and less outliers. Notice that in many cases (NewYork, RottPor, LauwHav) SuperPoint + SuperGlue is able to successfully identify correspondences not only on the structure but also in the clouds.

\subsection{Single- and two-view calibration results}
\begin{figure*}[!htbp]
    \centering
    \begin{tabular}{c@{ }c@{ }c@{ }c@{ }c@{ }c@{ }c@{ }}
     & Cam 0 & Single-view & Two-view & Cam 1 & Single-view & Two-view \\
    \rotname{Croatia} & \includegraphics[width=.16\columnwidth]{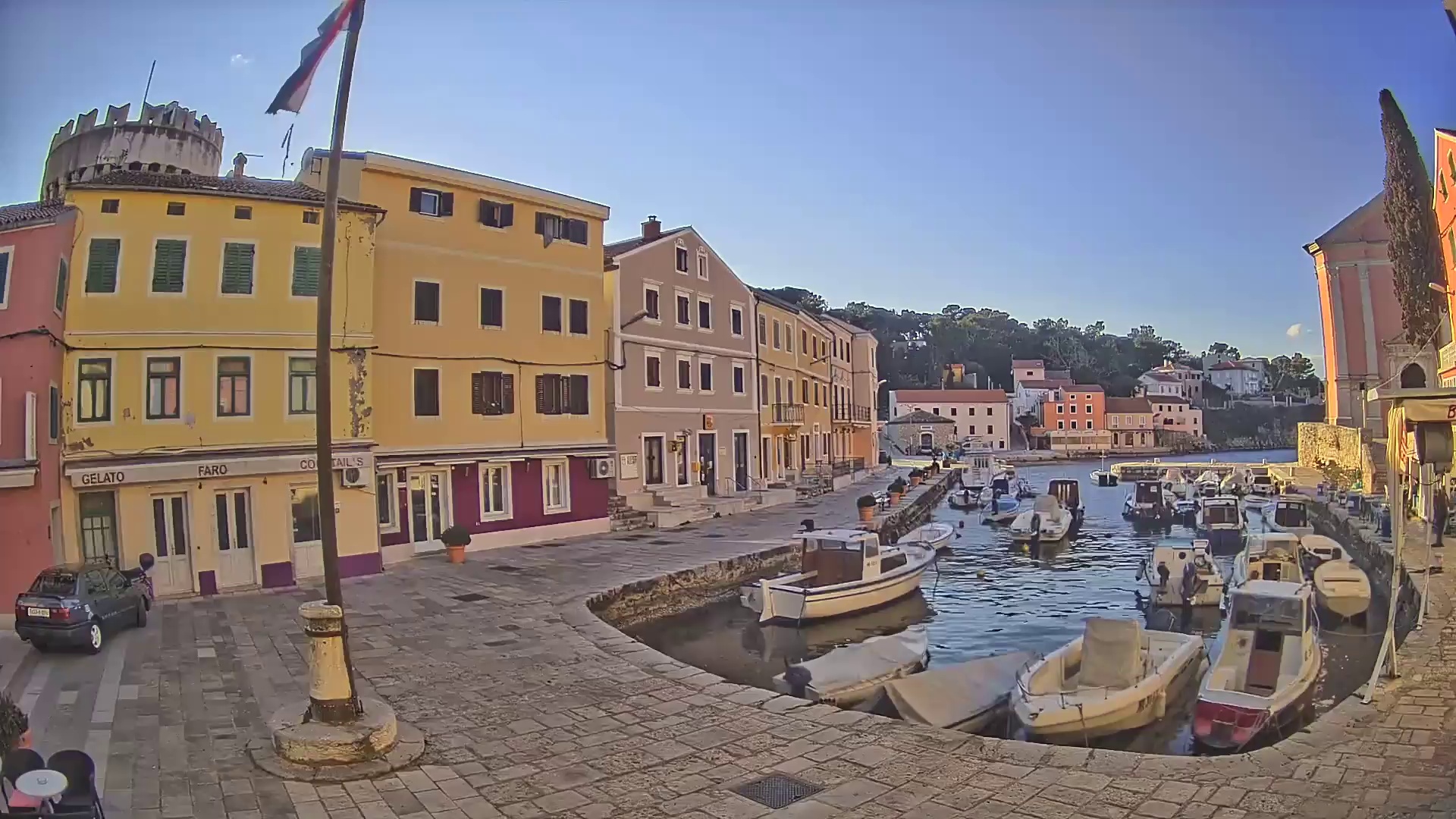} & \includegraphics[width=.16\columnwidth]{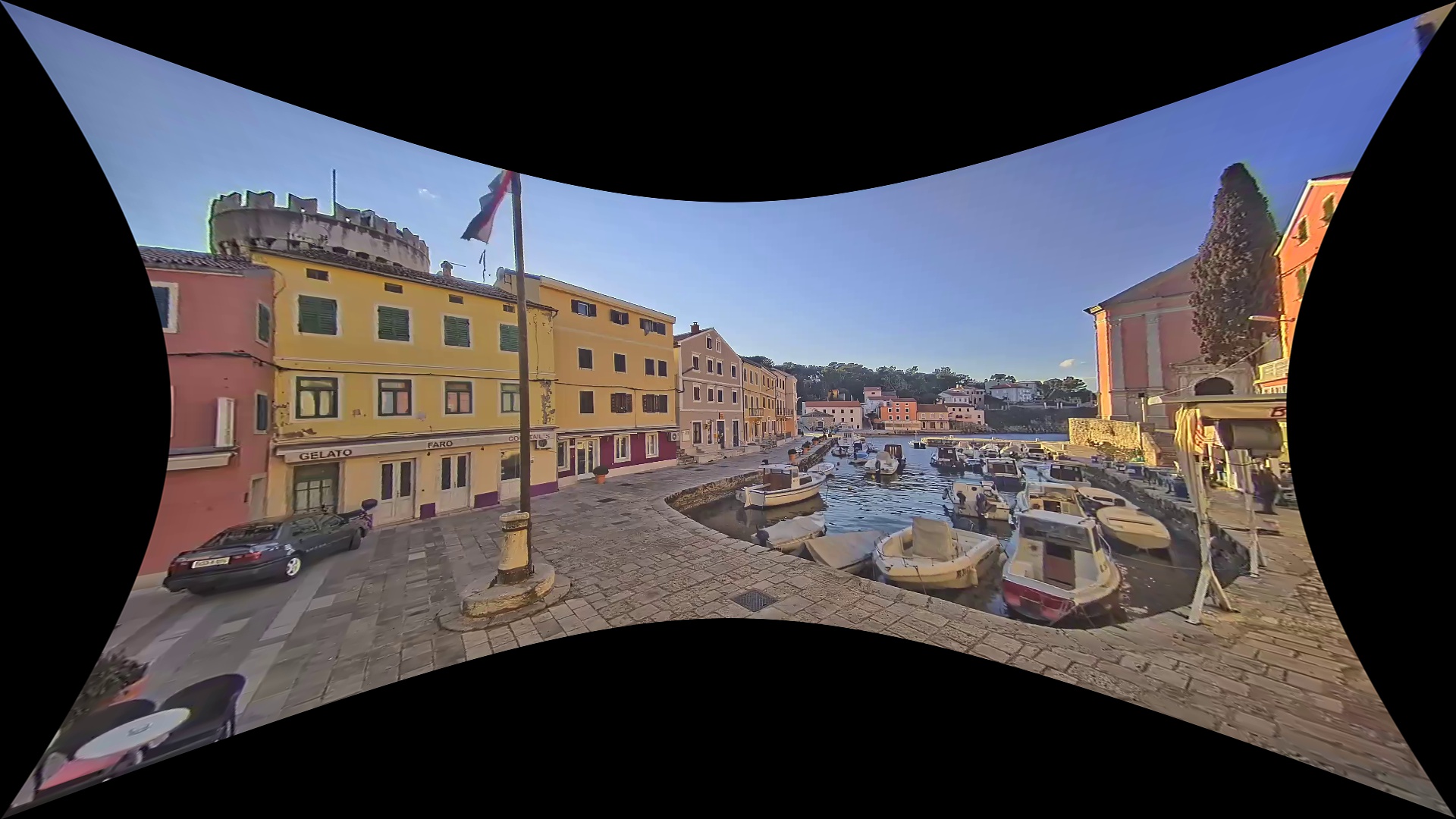} & \includegraphics[width=.16\columnwidth]{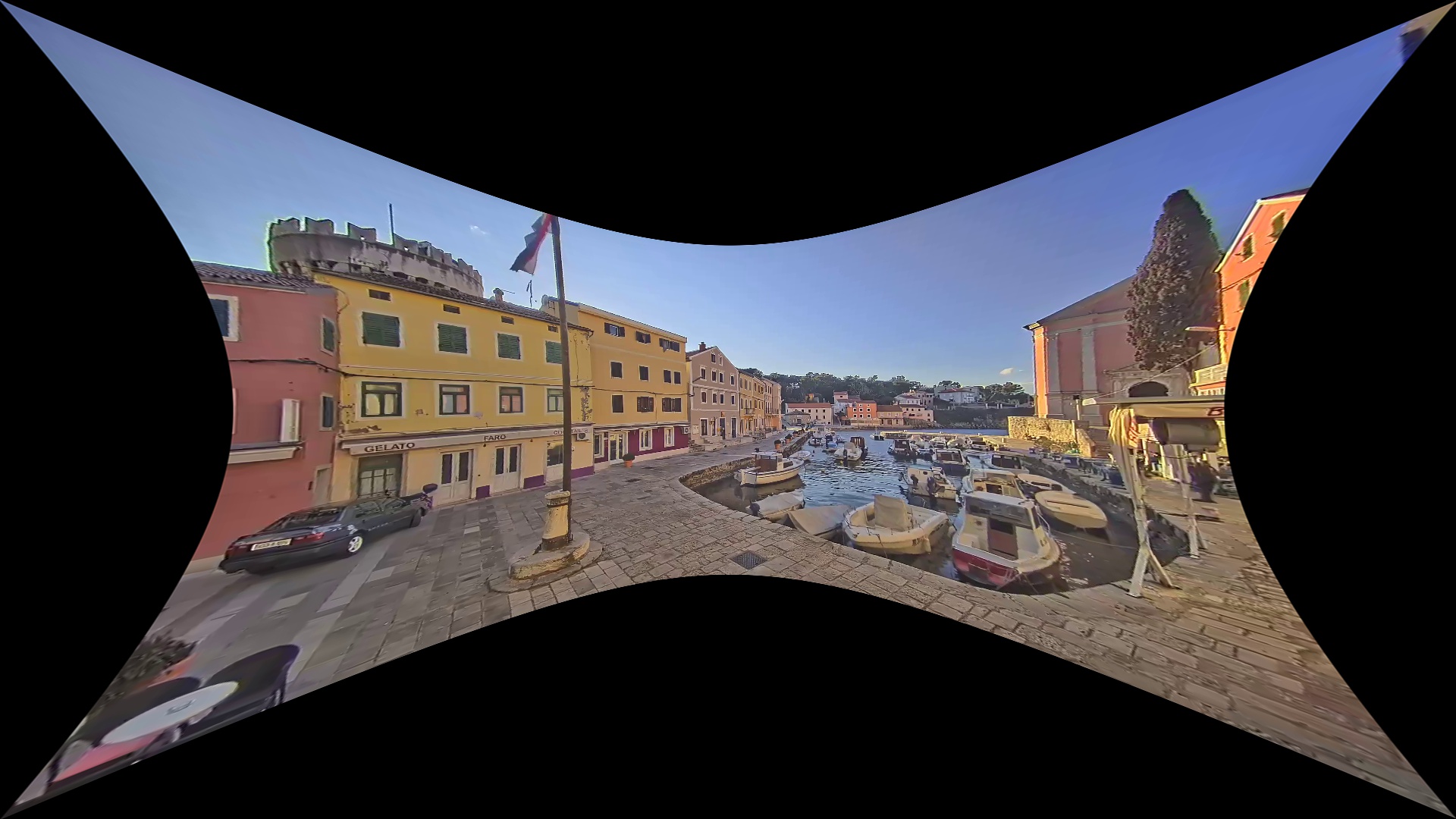} & \includegraphics[width=.16\columnwidth]{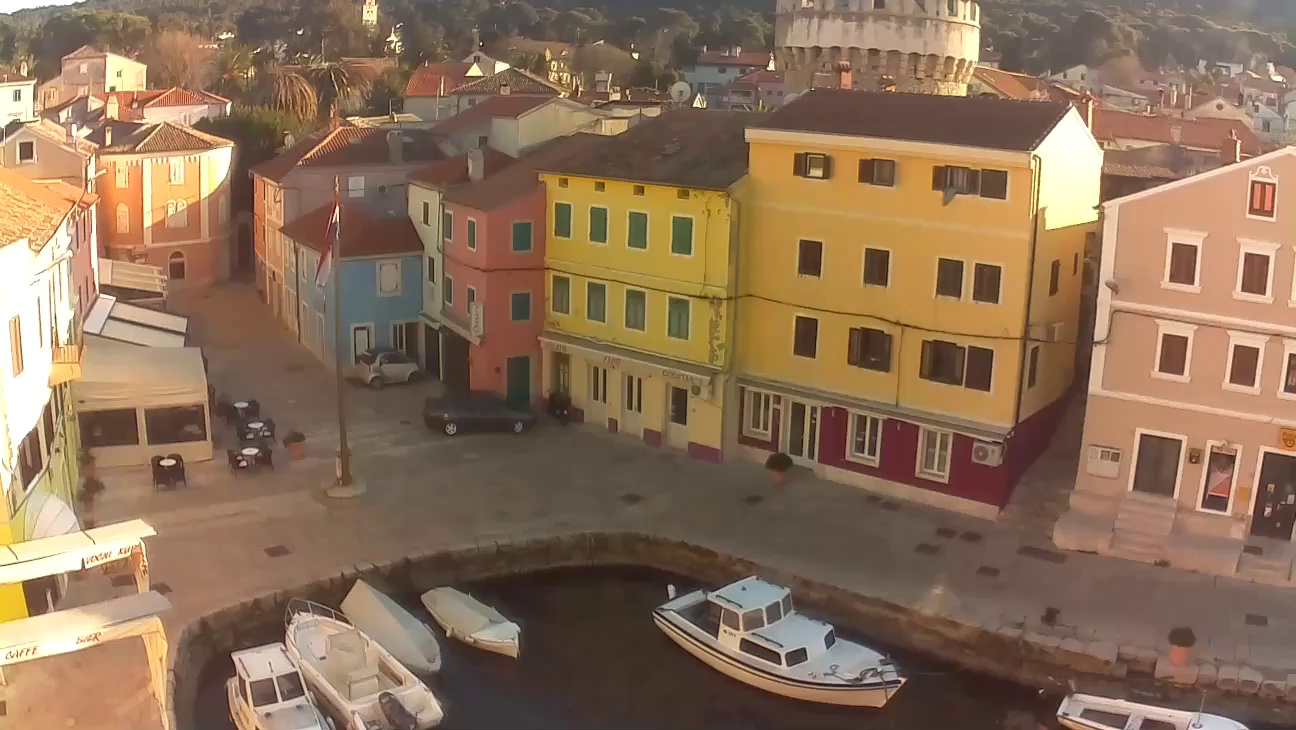} & \includegraphics[width=.16\columnwidth]{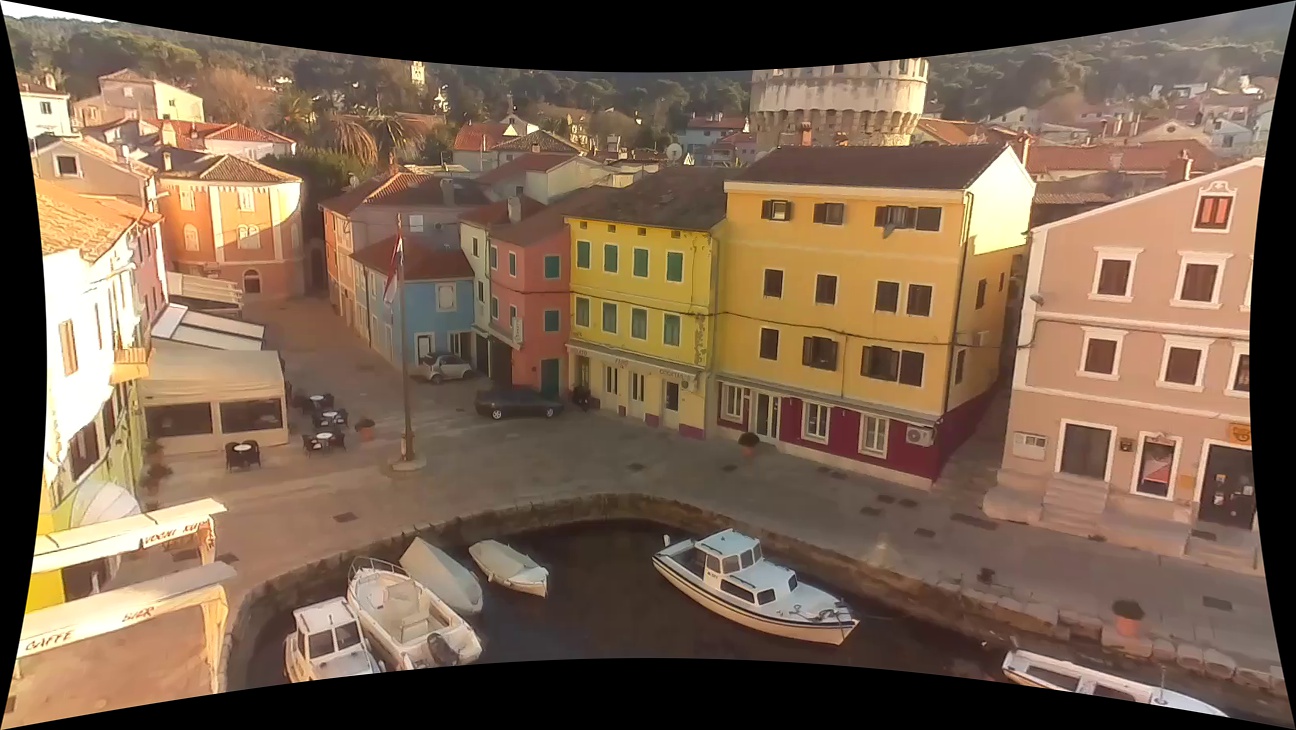} & \includegraphics[width=.16\columnwidth]{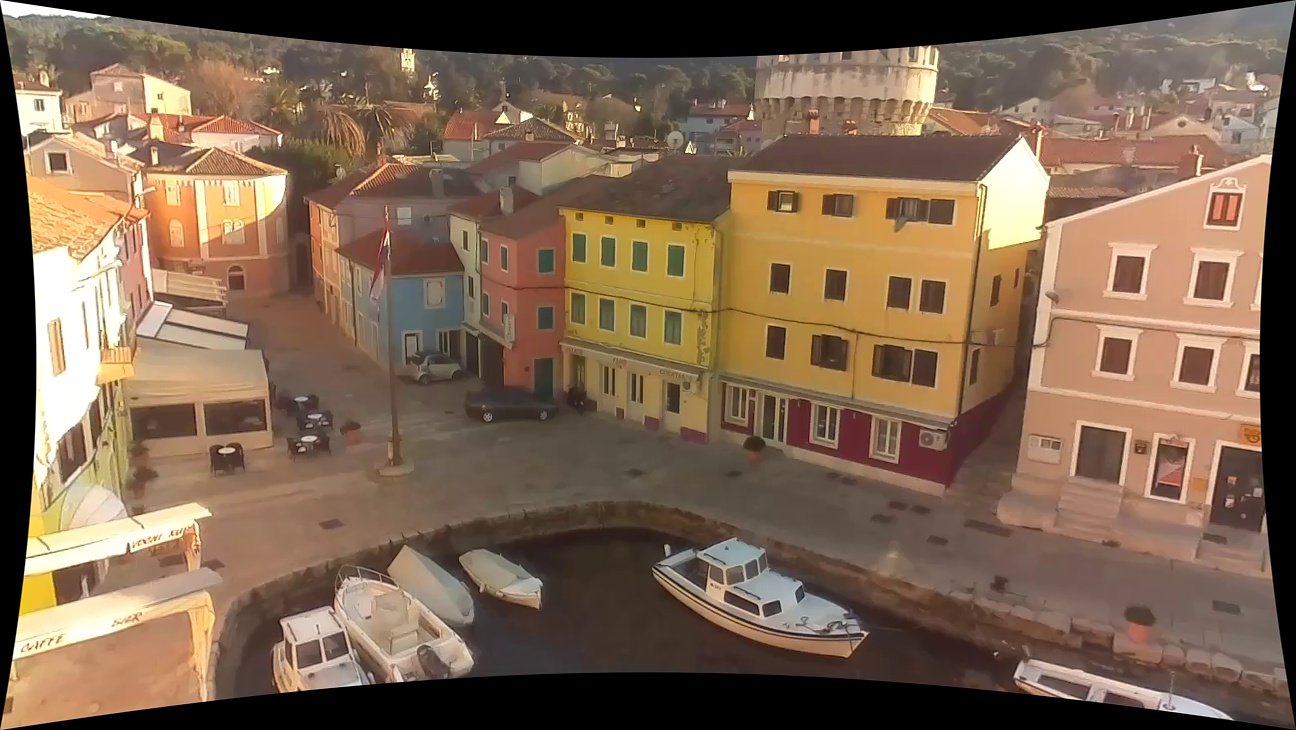} \\
    \rotname{NewYork} & \includegraphics[width=.16\columnwidth]{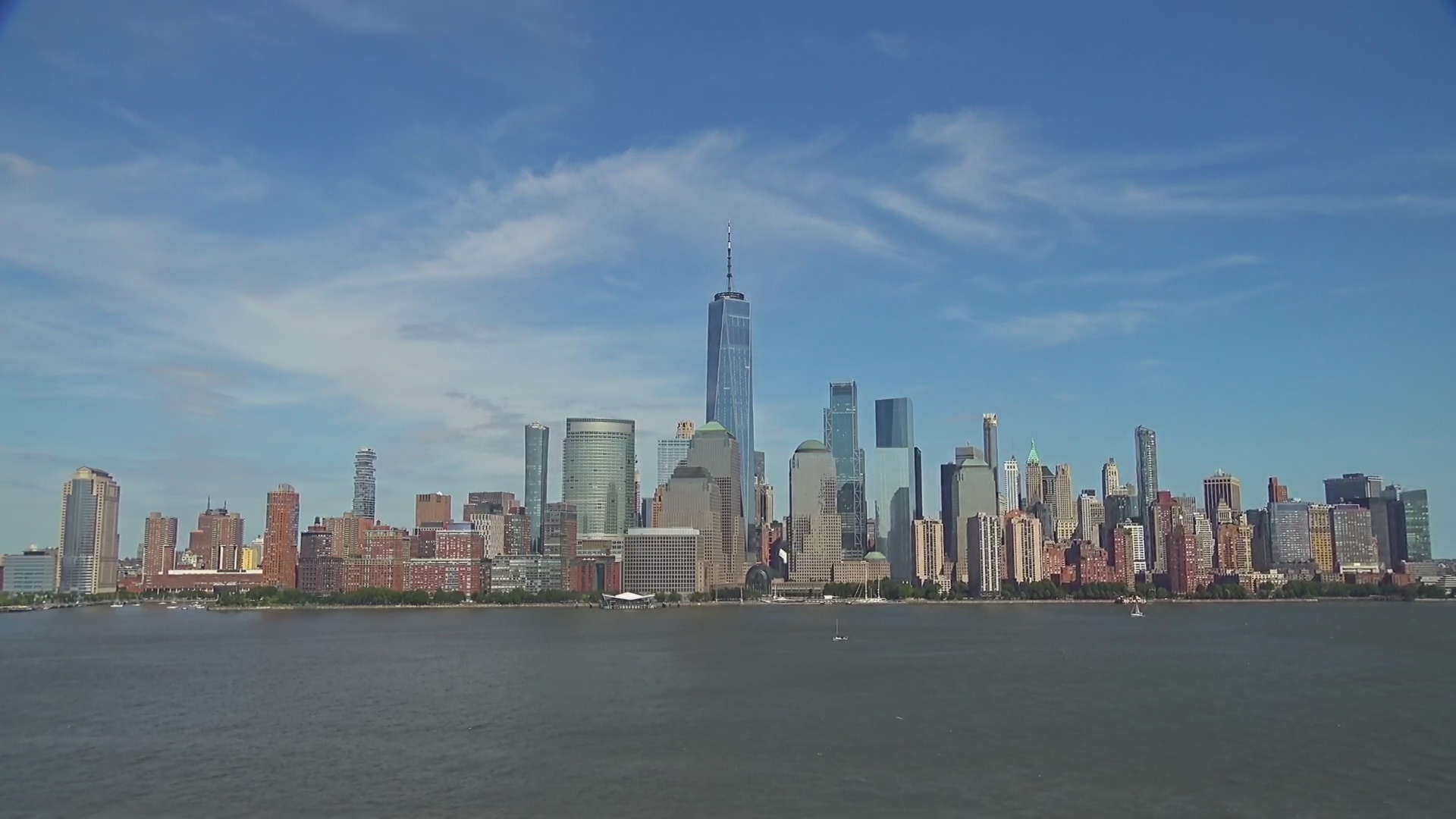} & \includegraphics[width=.16\columnwidth]{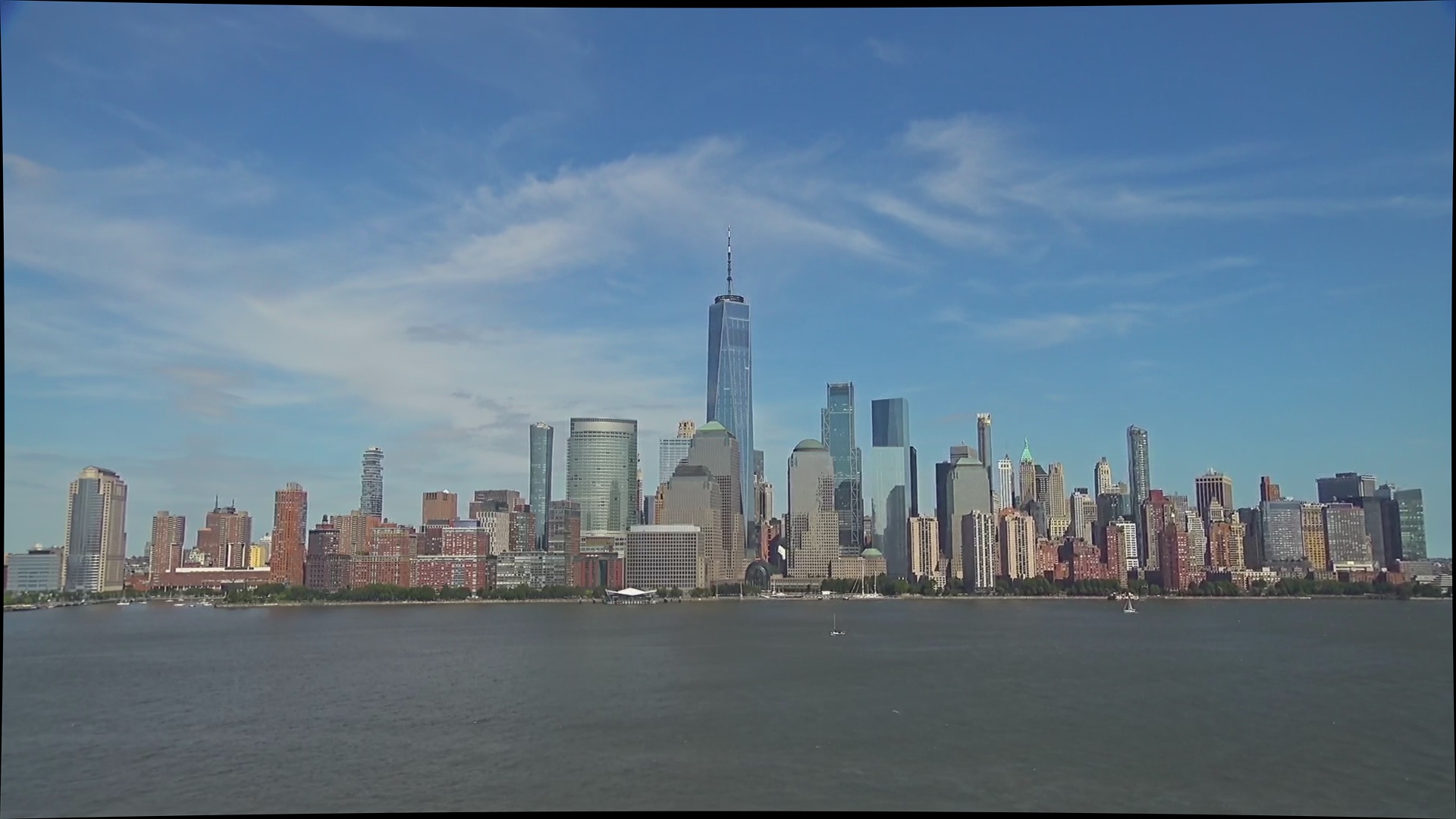} & \includegraphics[width=.16\columnwidth]{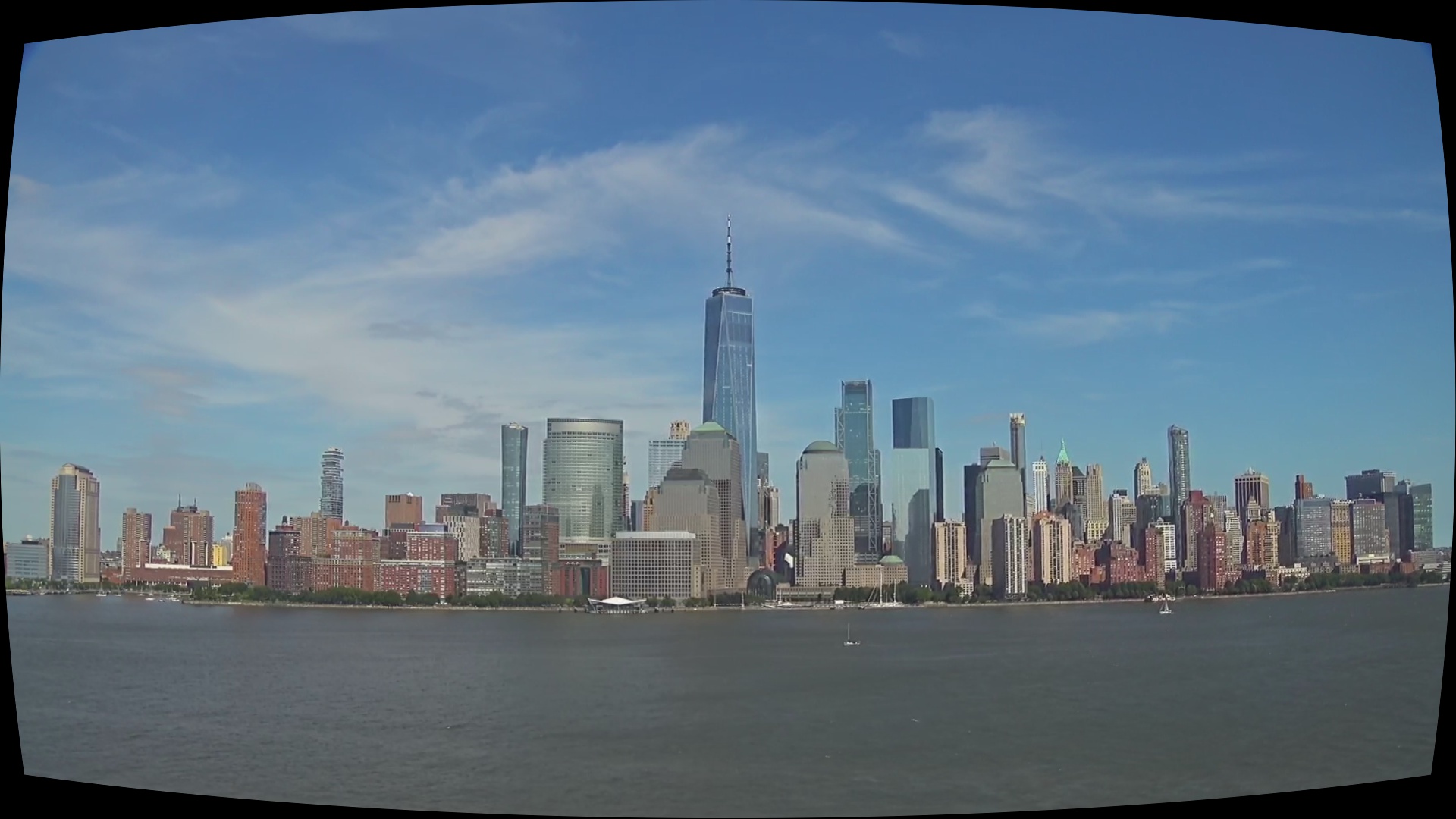} & \includegraphics[width=.16\columnwidth]{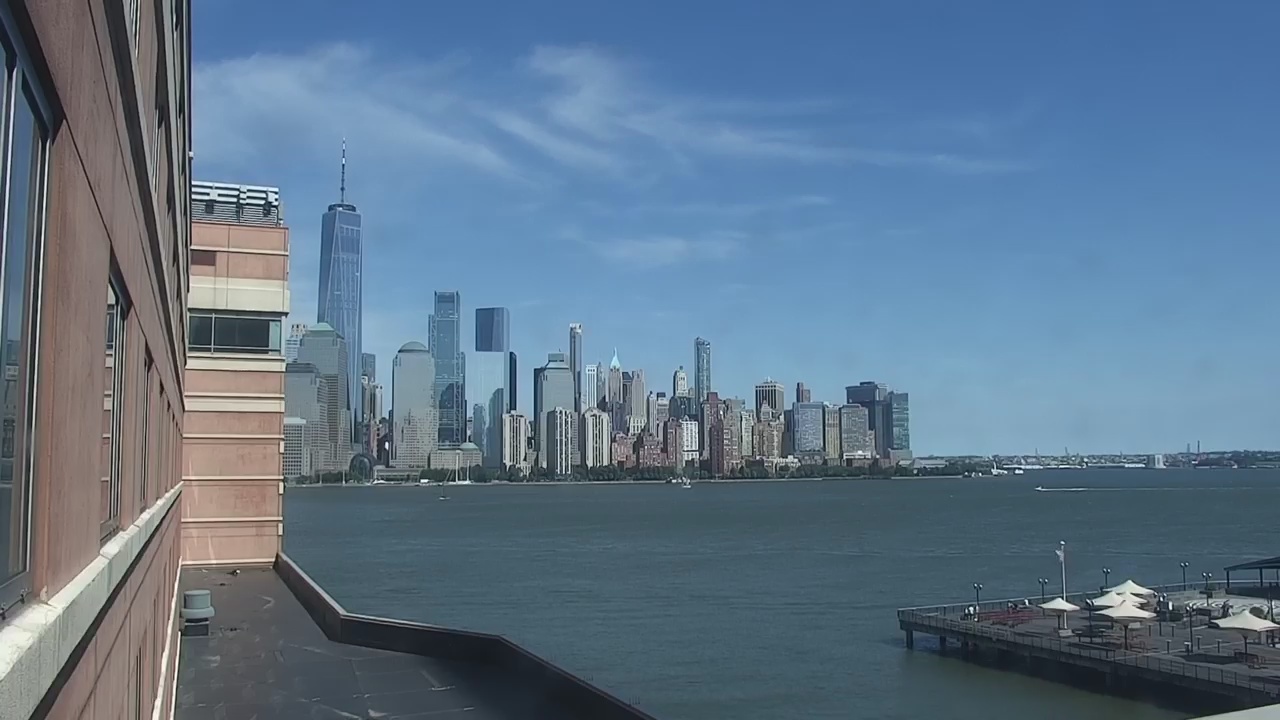} & \includegraphics[width=.16\columnwidth]{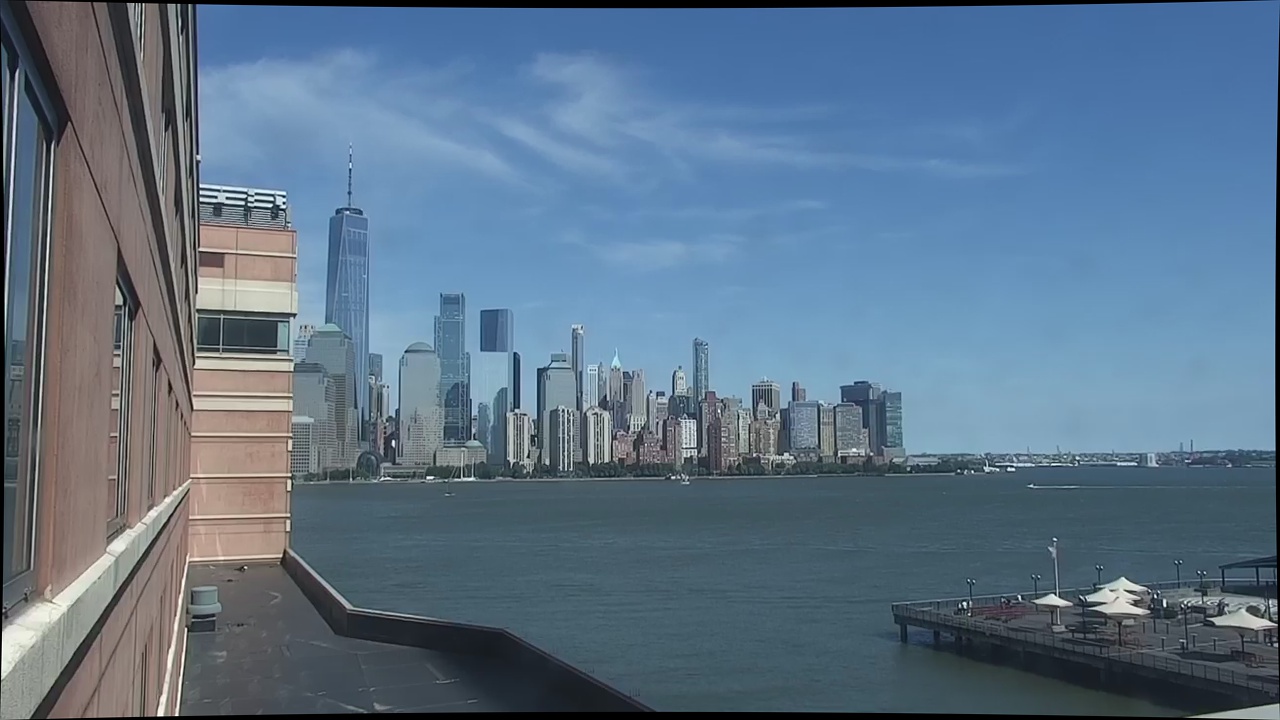} & \includegraphics[width=.16\columnwidth]{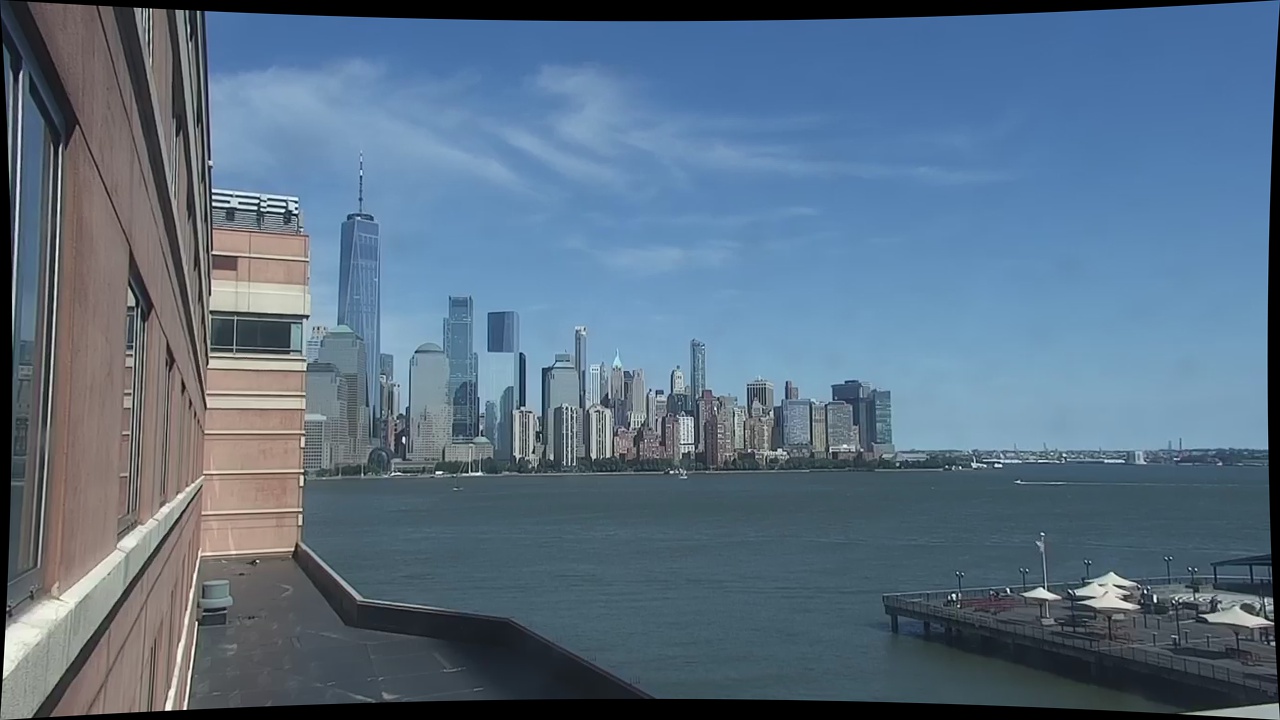} \\
    \rotname{TimeSq} & \includegraphics[width=.16\columnwidth]{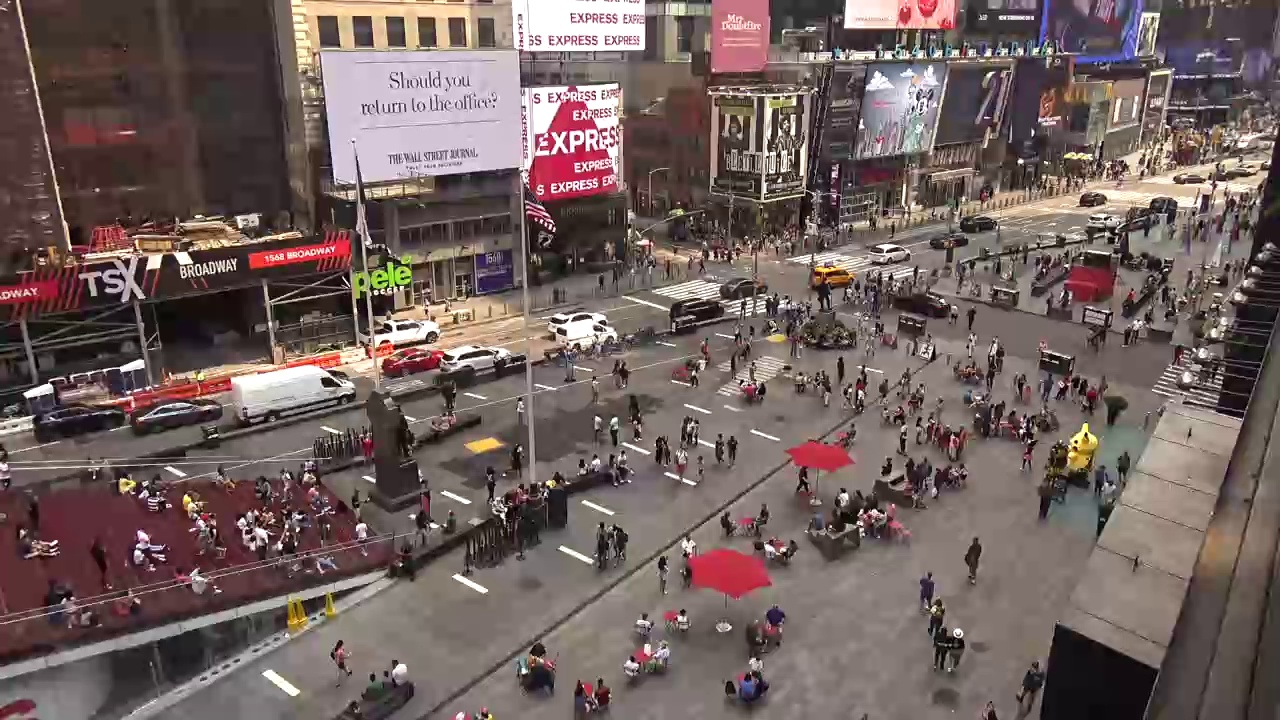} & \includegraphics[width=.16\columnwidth]{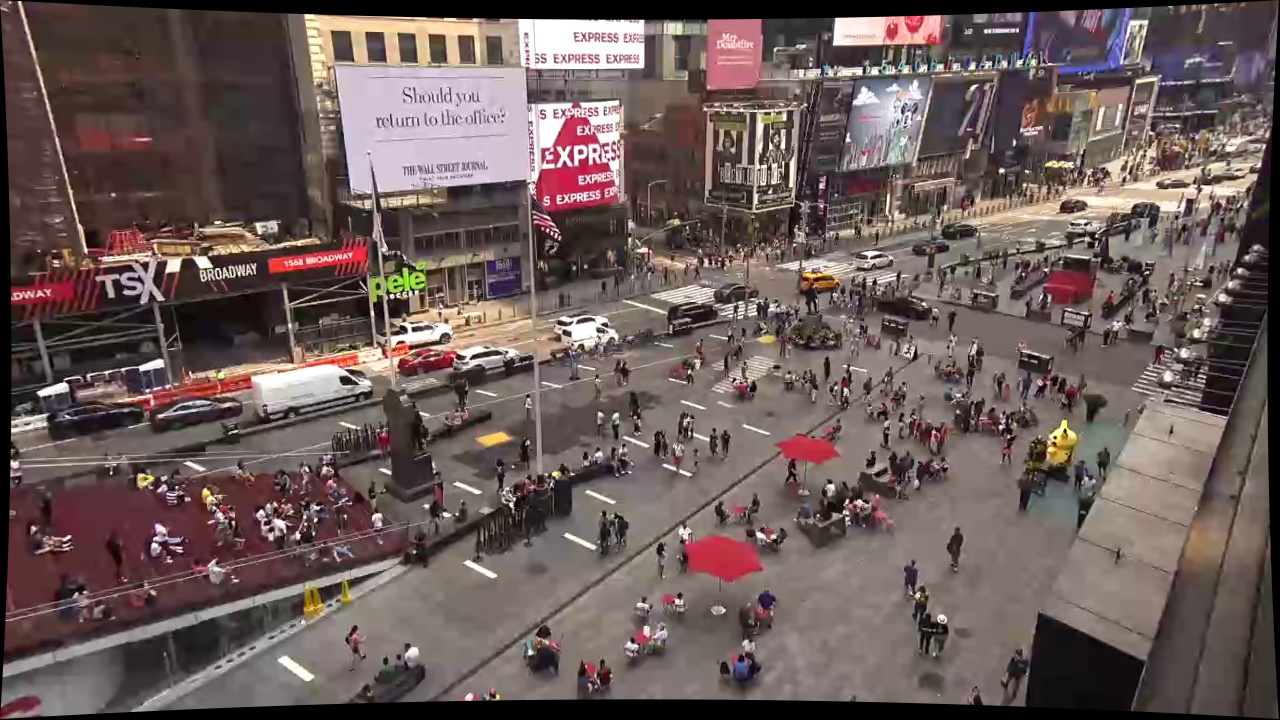} & \includegraphics[width=.16\columnwidth]{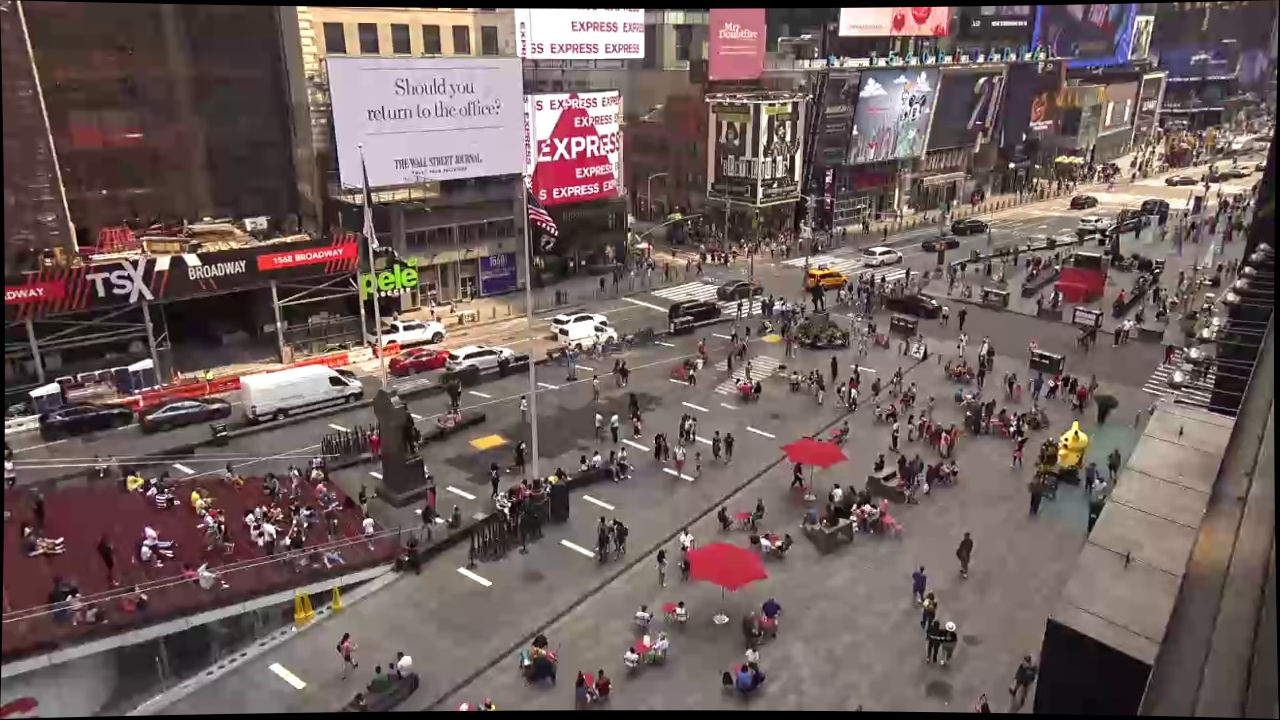} & \includegraphics[width=.16\columnwidth]{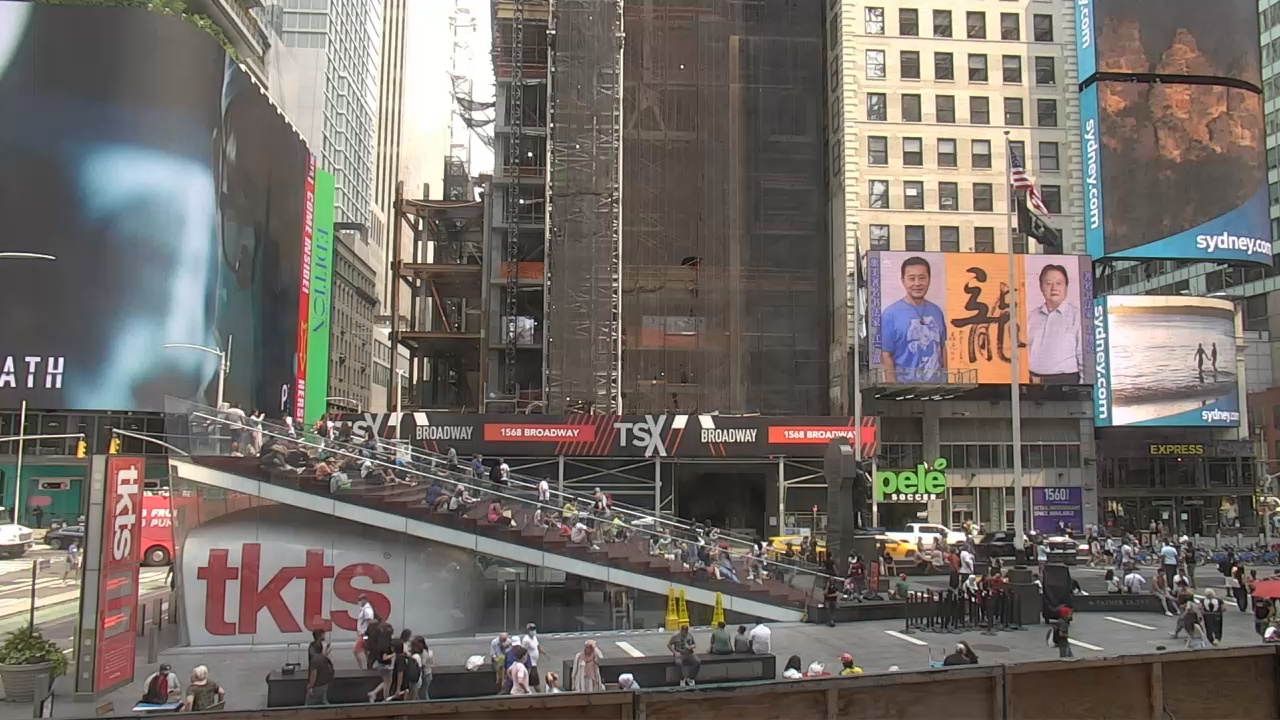} & \includegraphics[width=.16\columnwidth]{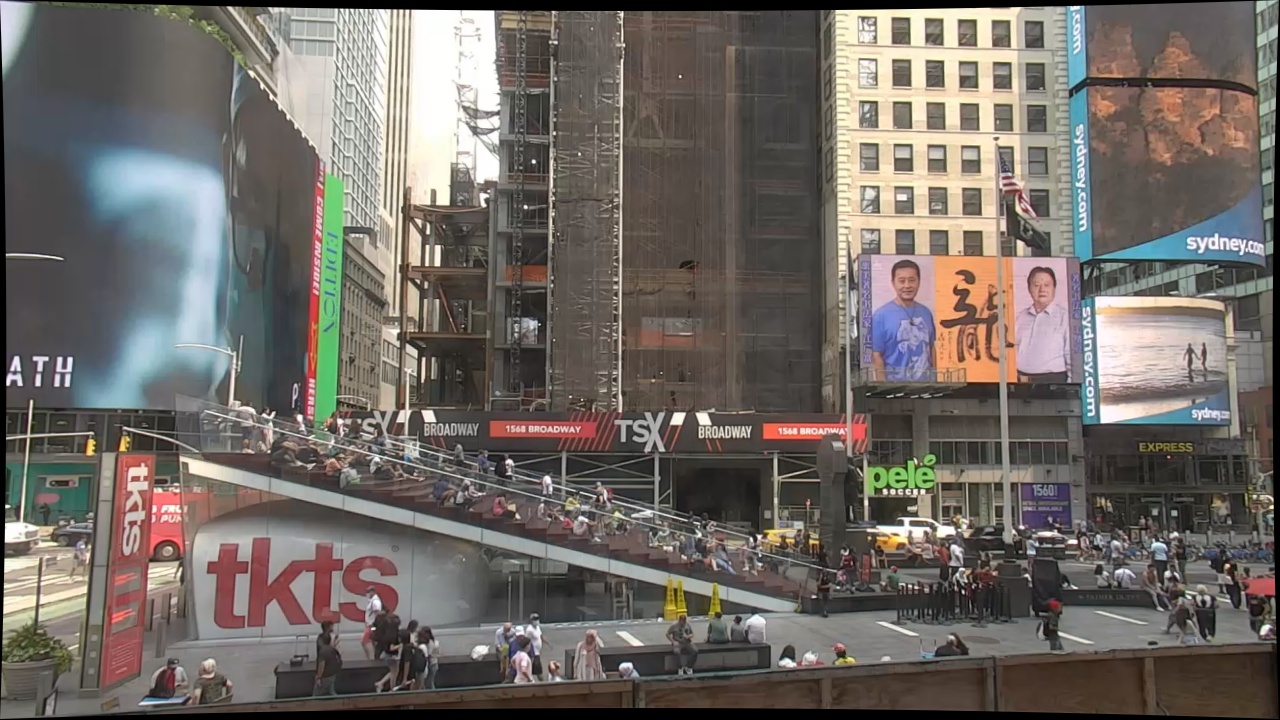} & \includegraphics[width=.16\columnwidth]{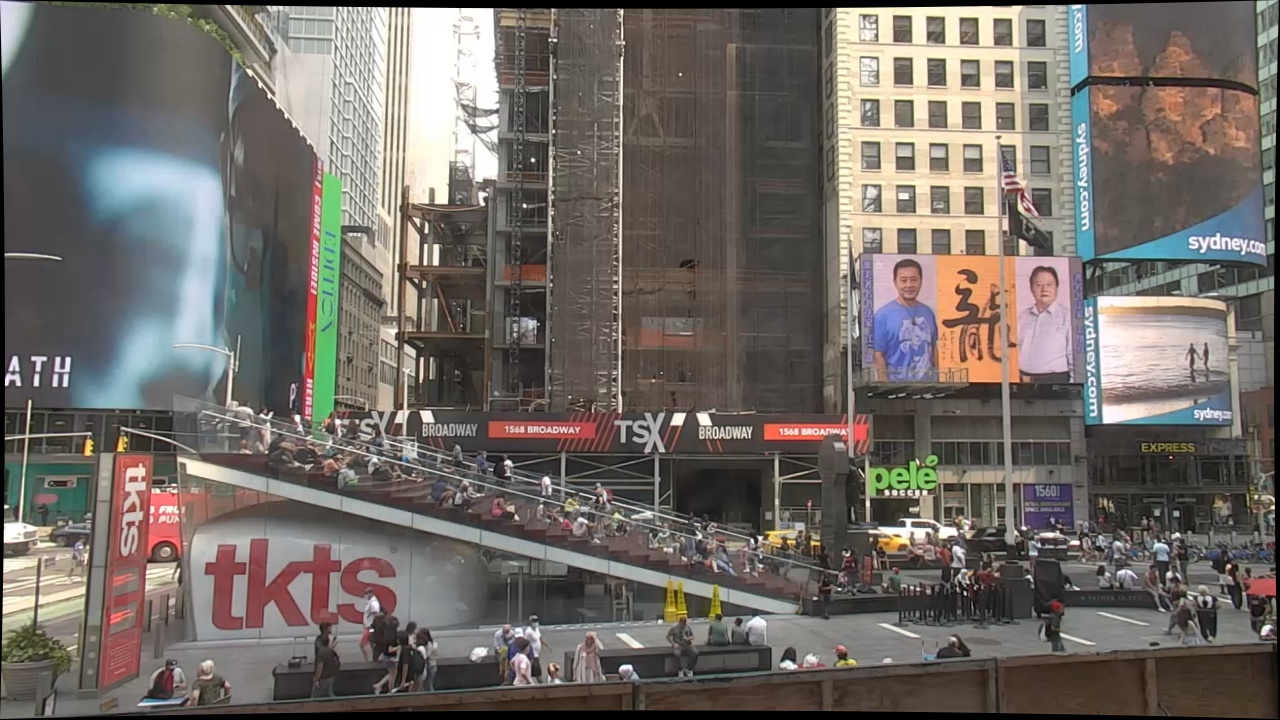} \\
    \rotname{AalsHav} & \includegraphics[width=.16\columnwidth]{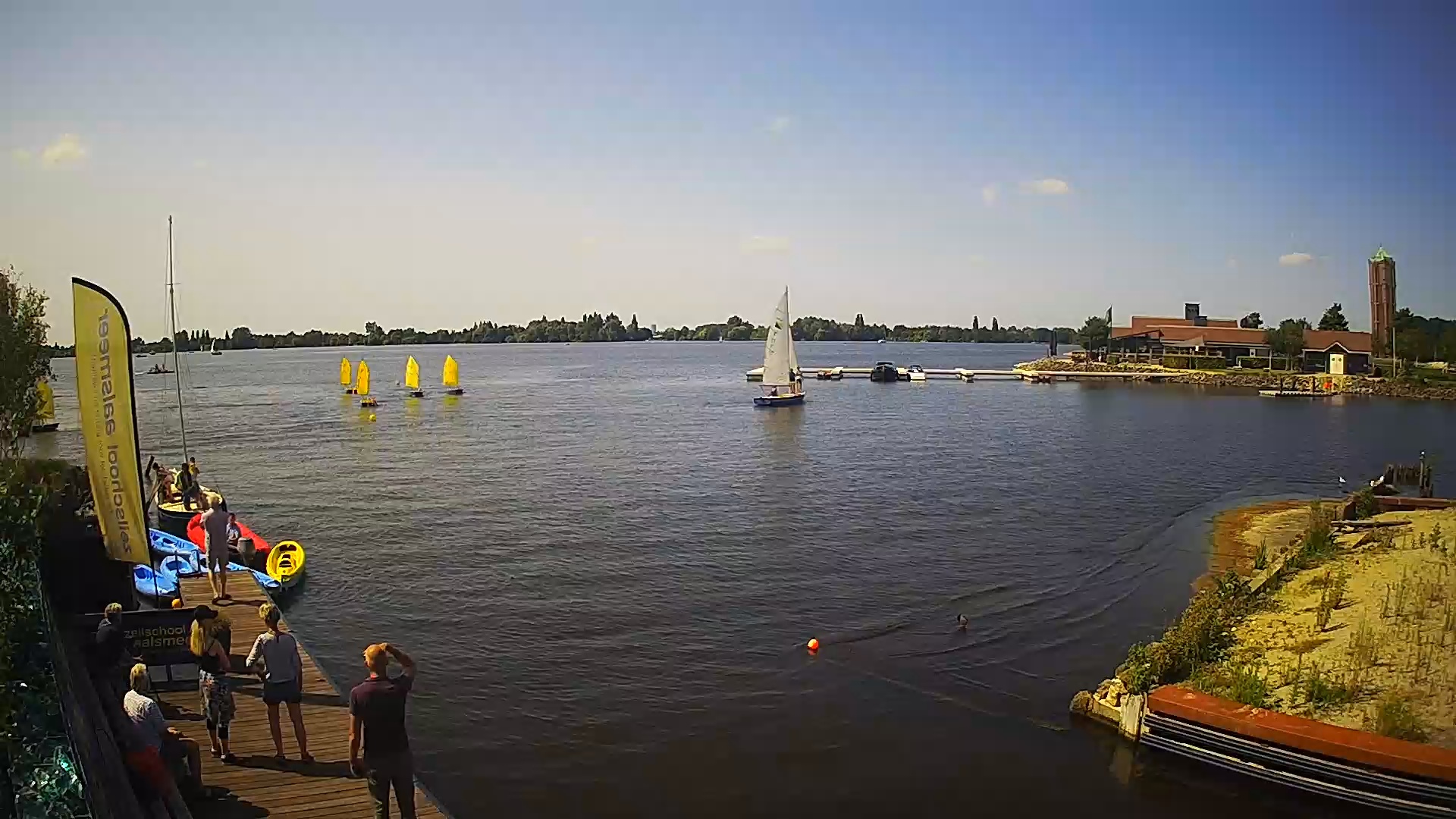} & \includegraphics[width=.16\columnwidth]{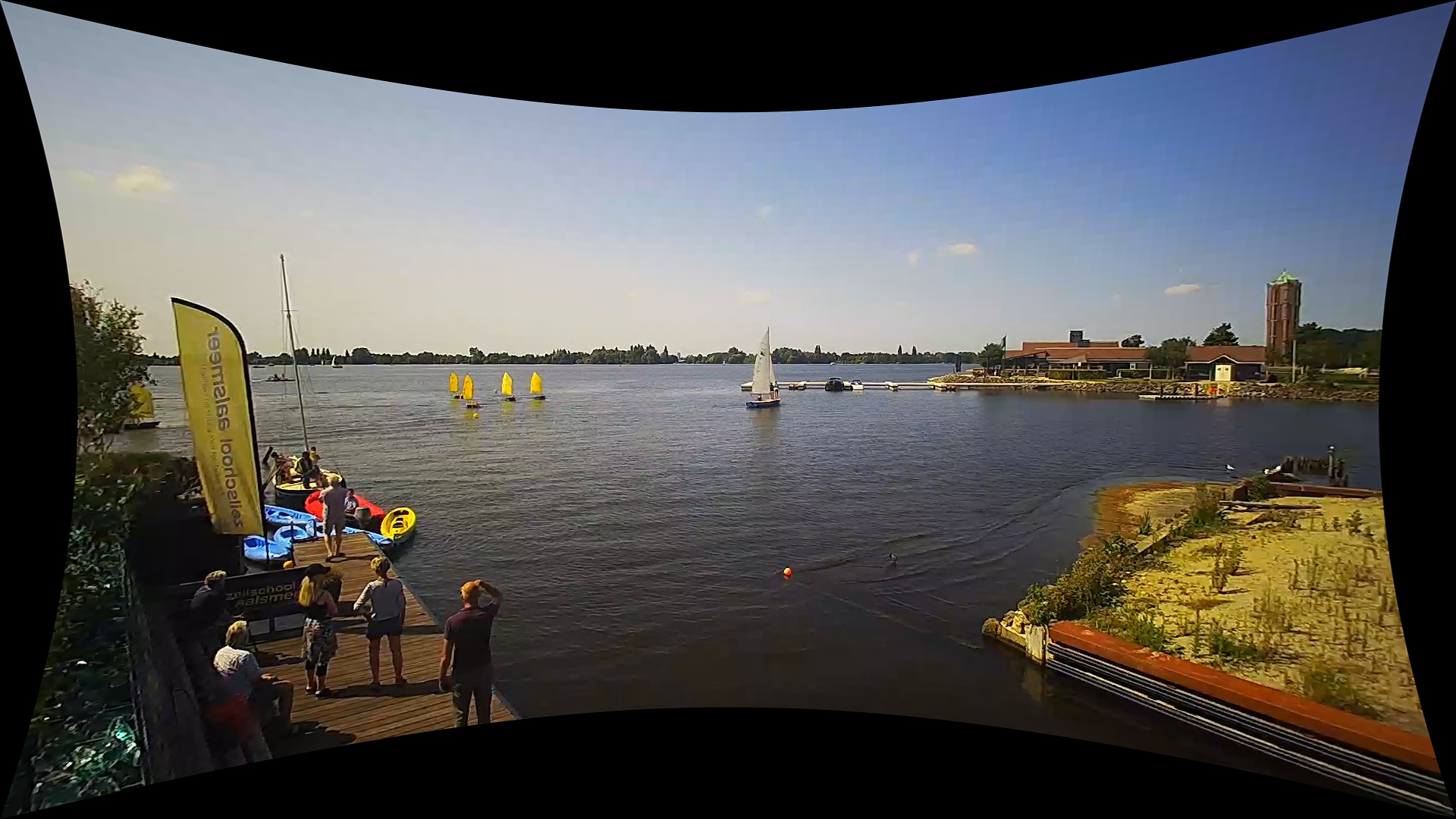} & \includegraphics[width=.16\columnwidth]{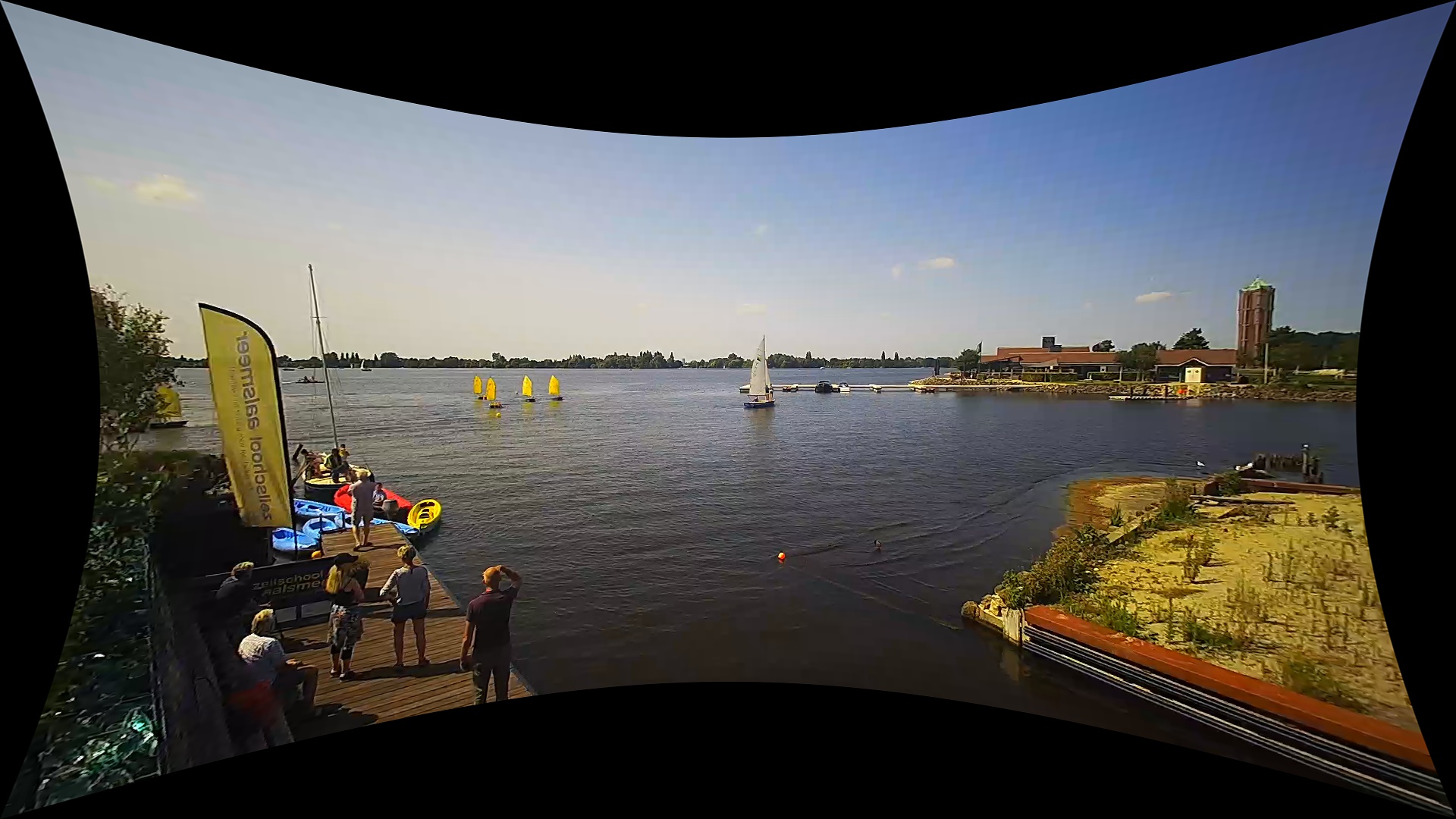} & \includegraphics[width=.16\columnwidth]{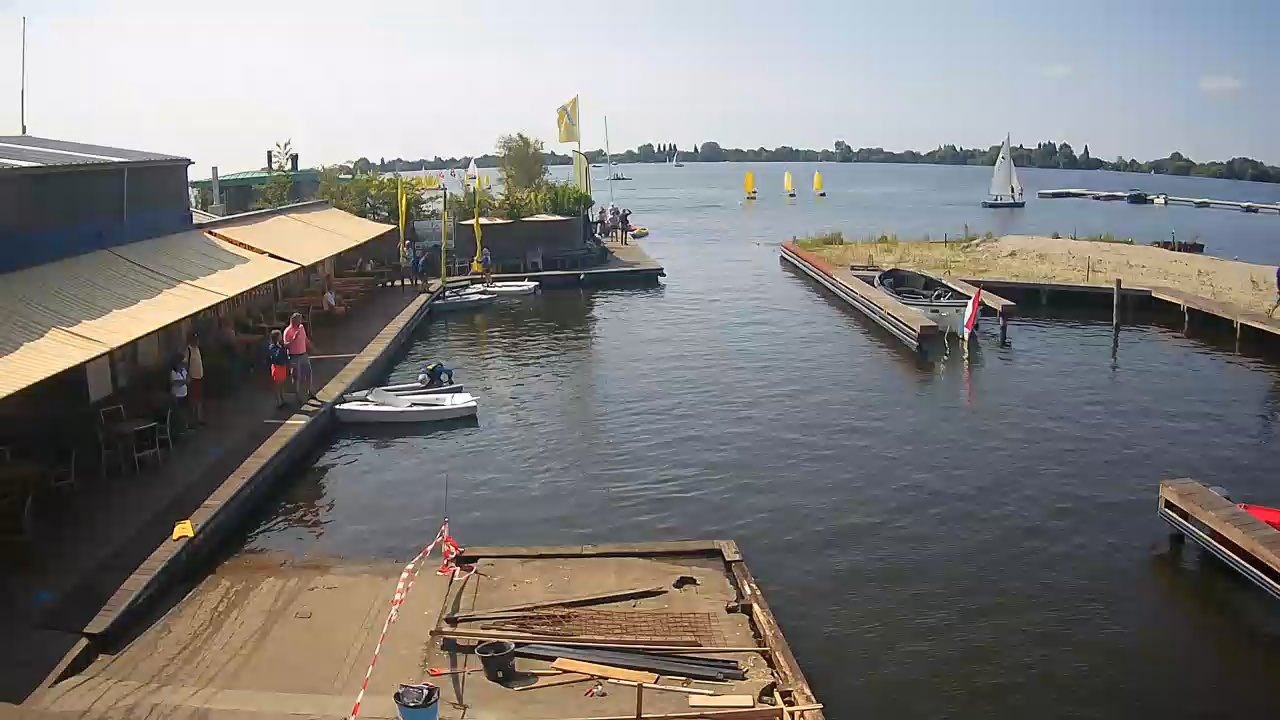} & \includegraphics[width=.16\columnwidth]{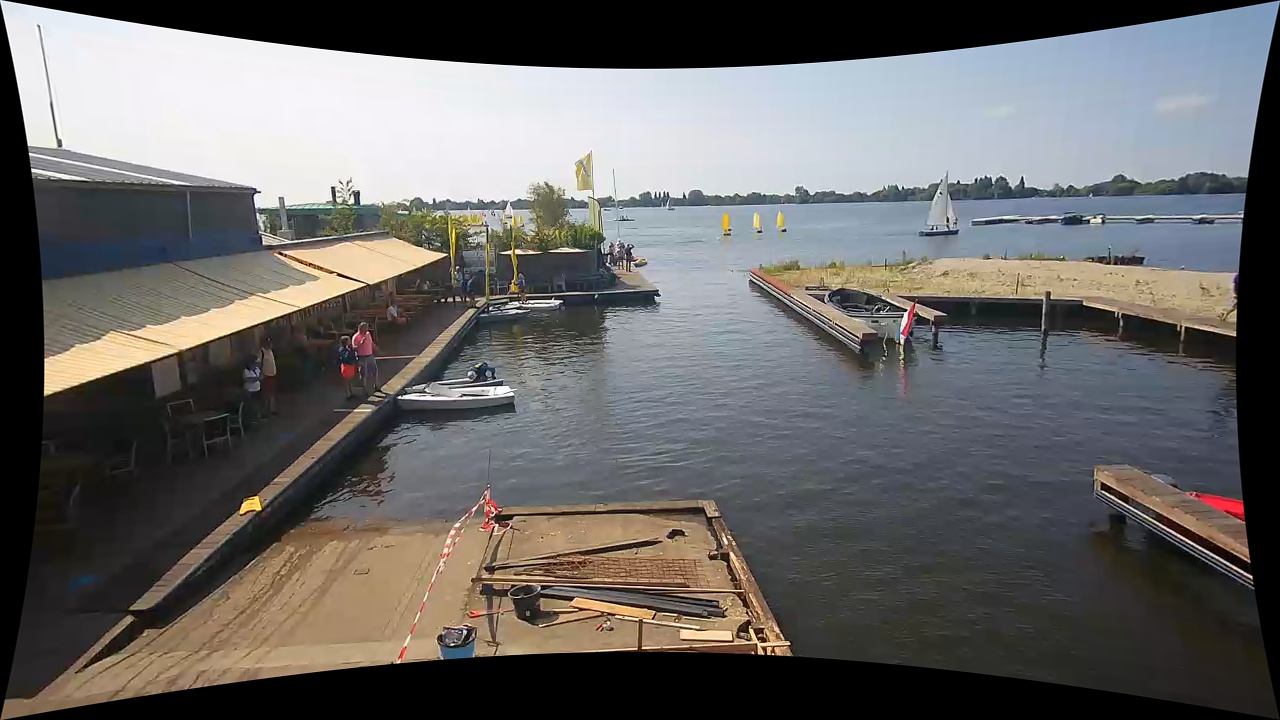} & \includegraphics[width=.16\columnwidth]{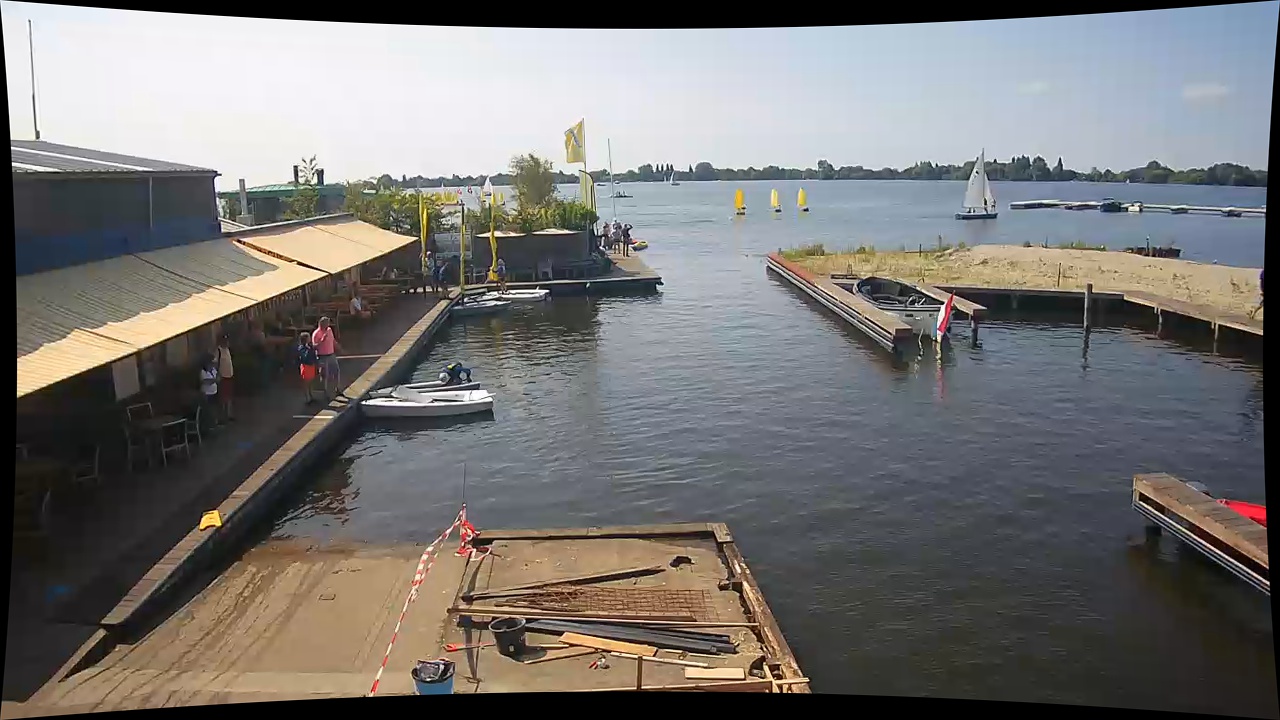} \\
    \rotname{BranHot} & \includegraphics[width=.16\columnwidth]{images/cam_calib/brandaris_hotel/cam0.jpg} & \includegraphics[width=.16\columnwidth]{images/cam_calib/brandaris_hotel/cam0_undistort.jpg} & \includegraphics[width=.16\columnwidth]{images/cam_calib/brandaris_hotel/cam0_undistort_2v.jpg} & \includegraphics[width=.16\columnwidth]{images/cam_calib/brandaris_hotel/cam1.jpg} & \includegraphics[width=.16\columnwidth]{images/cam_calib/brandaris_hotel/cam1_undistort.jpg} & \includegraphics[width=.16\columnwidth]{images/cam_calib/brandaris_hotel/cam1_undistort_2v.jpg} \\
    \rotname{BranPor} & \includegraphics[width=.16\columnwidth]{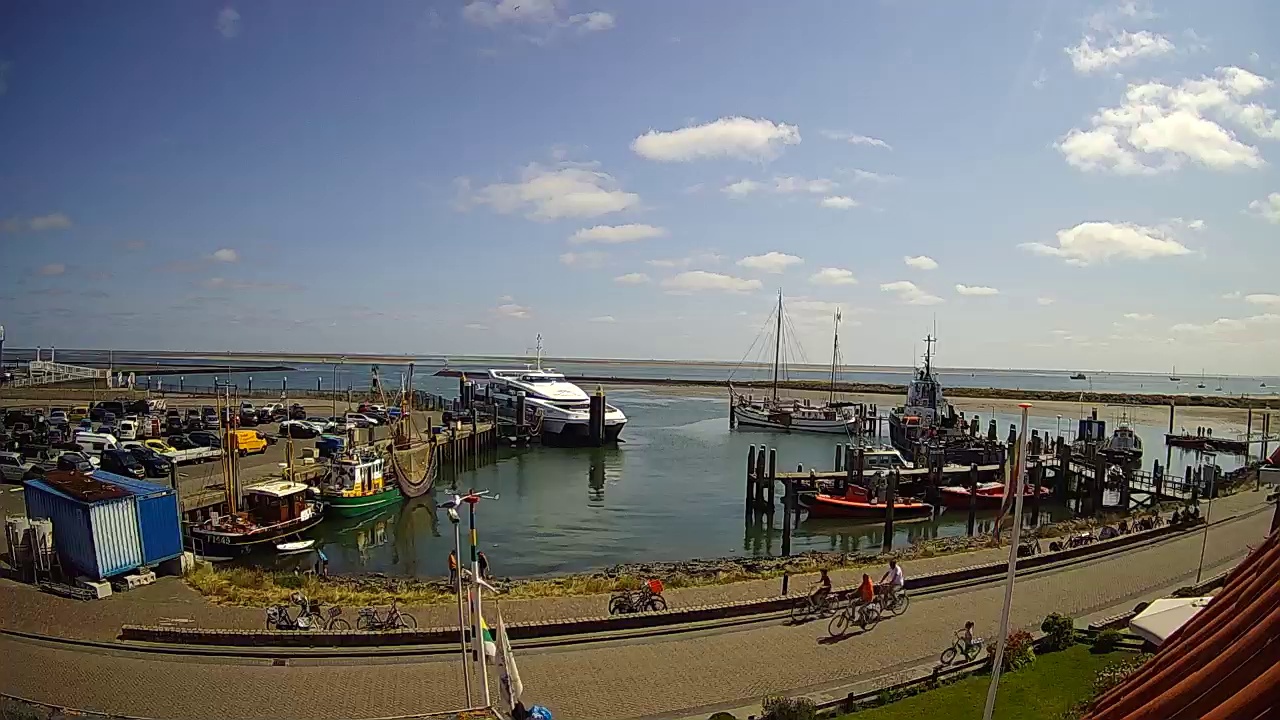} & \includegraphics[width=.16\columnwidth]{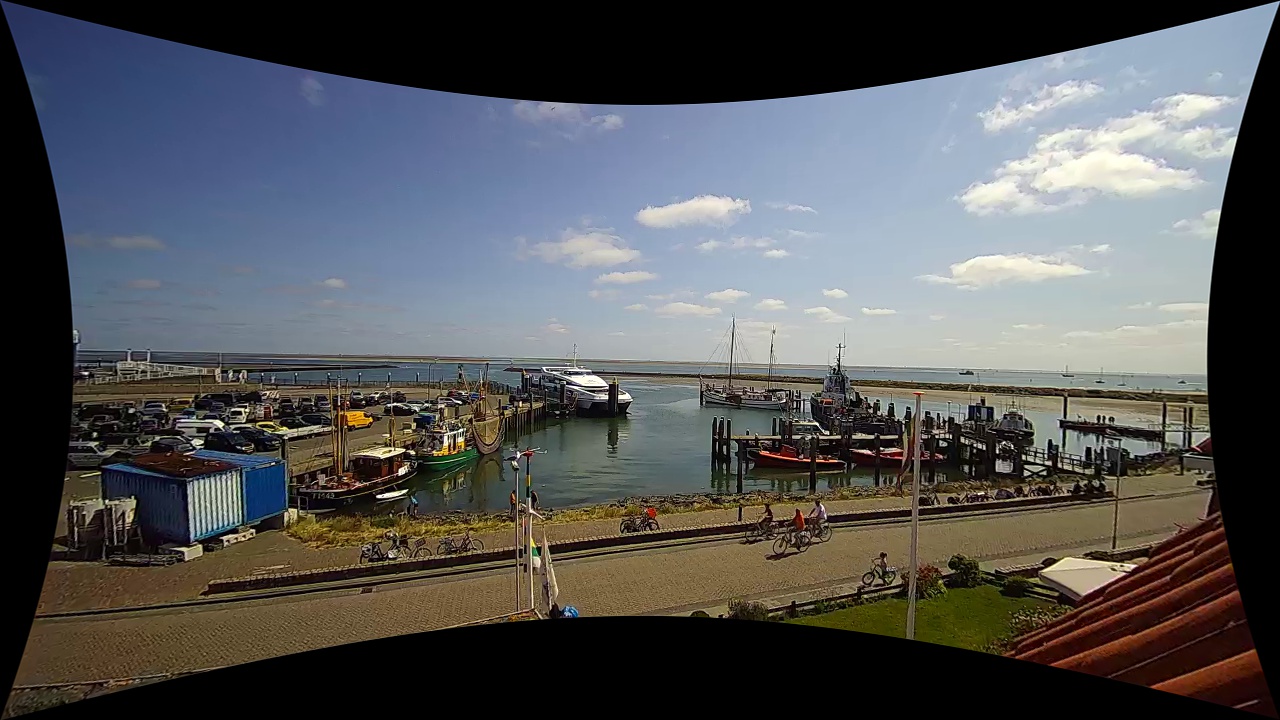} & \includegraphics[width=.16\columnwidth]{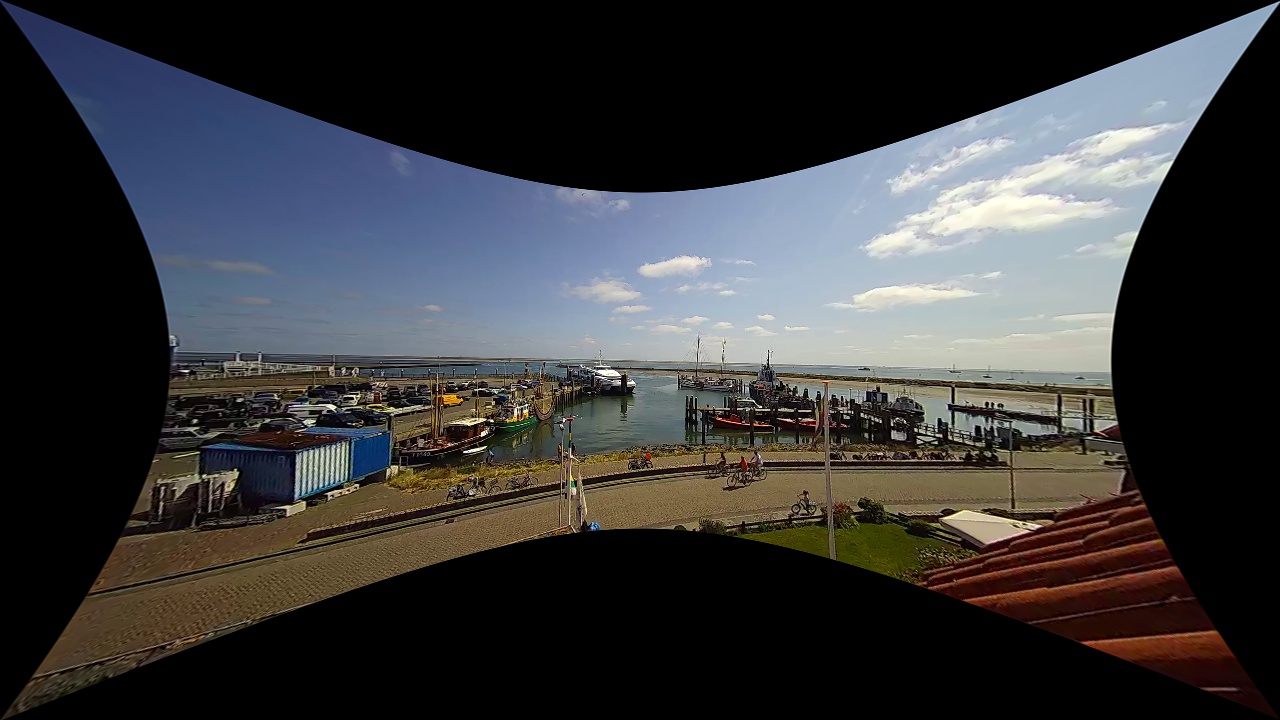} & \includegraphics[width=.16\columnwidth]{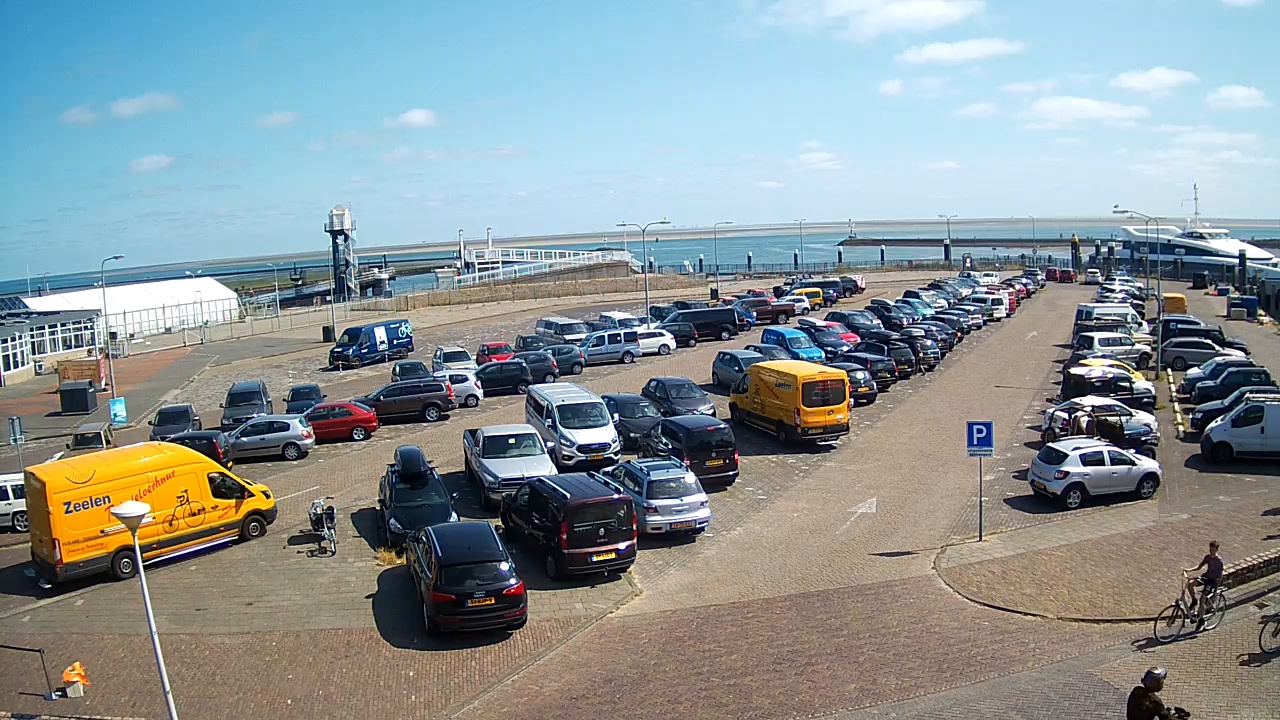} & \includegraphics[width=.16\columnwidth]{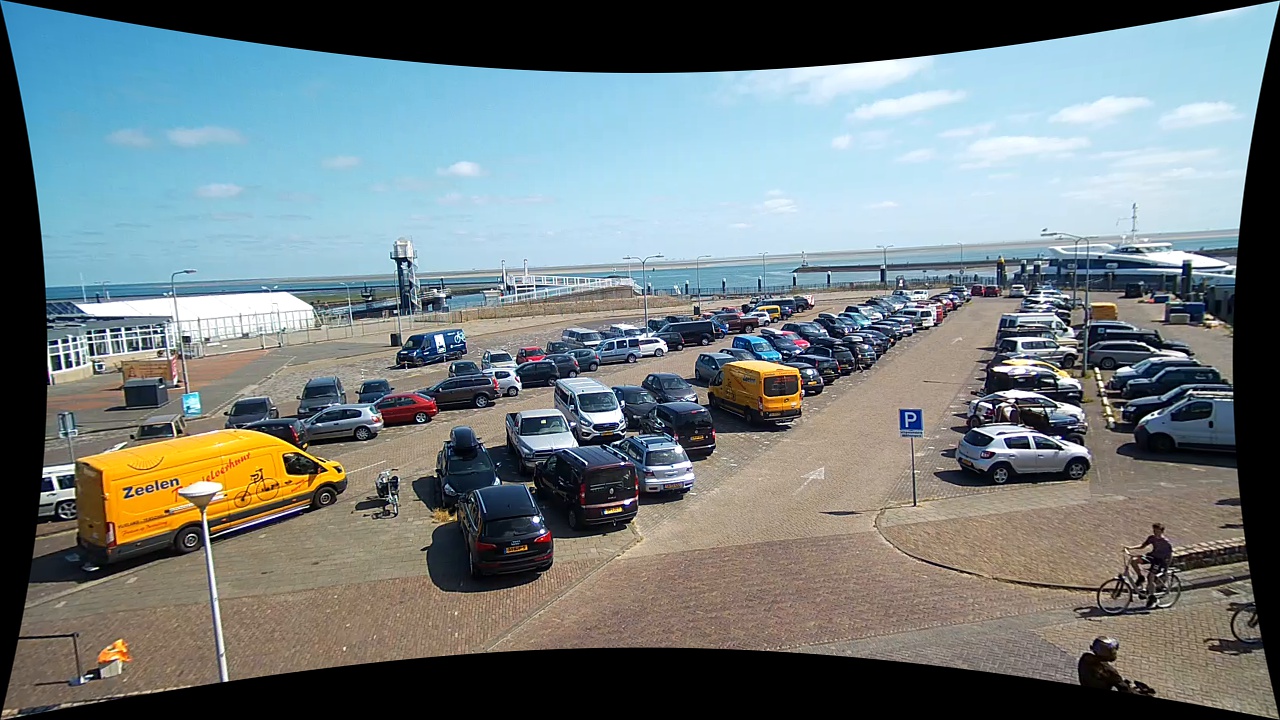} & \includegraphics[width=.16\columnwidth]{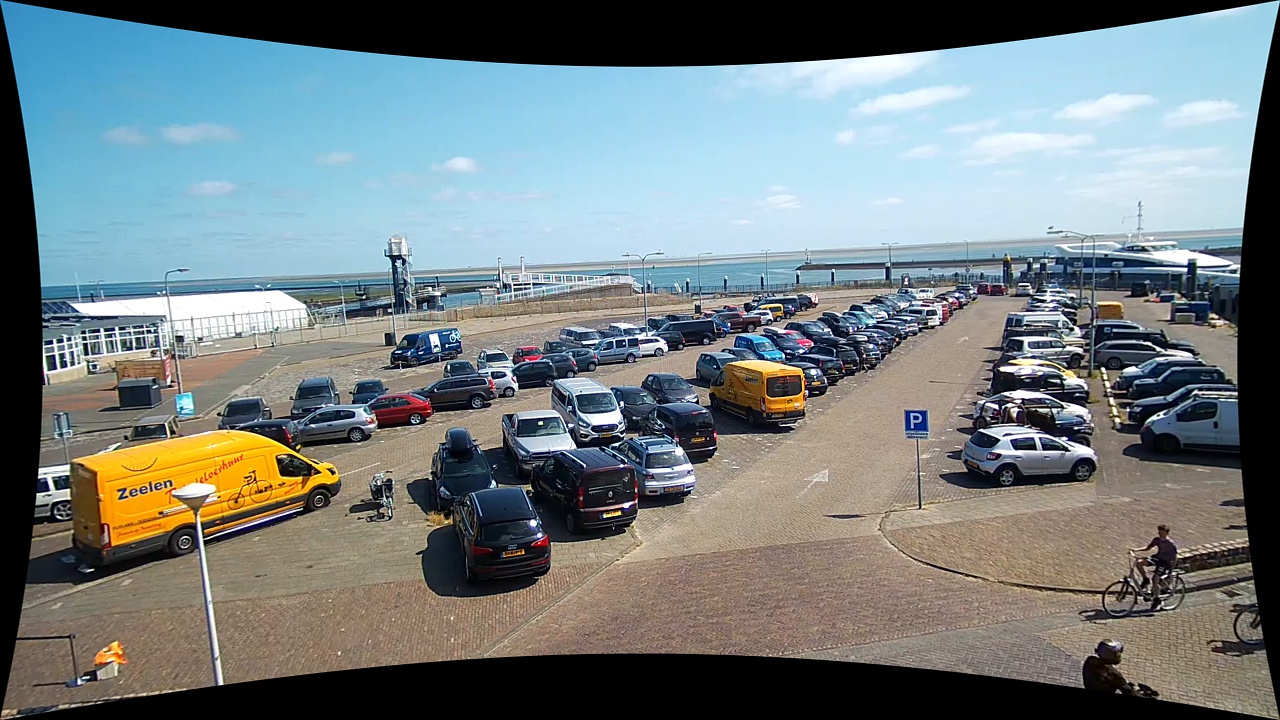} \\
    \rotname{RottPor} & \includegraphics[width=.16\columnwidth]{images/cam_calib/port_rotterdam/cam0_0721.jpg} & \includegraphics[width=.16\columnwidth]{images/cam_calib/port_rotterdam/cam0_0721_undistort_div.jpg} & \includegraphics[width=.16\columnwidth]{images/cam_calib/port_rotterdam/cam0_0721_undistort_div_2v.jpg} & \includegraphics[width=.16\columnwidth]{images/cam_calib/port_rotterdam/cam1_0721.jpg} & \includegraphics[width=.16\columnwidth]{images/cam_calib/port_rotterdam/cam1_0721_undistort_div.jpg} & \includegraphics[width=.16\columnwidth]{images/cam_calib/port_rotterdam/cam1_0721_undistort_div_2v.jpg} \\
    \rotname{TexAir} & \includegraphics[width=.16\columnwidth]{images/cam_calib/texel_airport/cam0.jpg} & \includegraphics[width=.16\columnwidth]{images/cam_calib/texel_airport/cam0_undistort_div.jpg} & \includegraphics[width=.16\columnwidth]{images/cam_calib/texel_airport/cam0_undistort_2v.jpg} & \includegraphics[width=.16\columnwidth]{images/cam_calib/texel_airport/cam1.jpg} & \includegraphics[width=.16\columnwidth]{images/cam_calib/texel_airport/cam1_undistort_div.jpg} & \includegraphics[width=.16\columnwidth]{images/cam_calib/texel_airport/cam1_undistort_div_2v.jpg}
    \end{tabular}
    \caption{The calibration results of all tested datasets with the single- and two-view calibration approach. From left to right: original image, single-view calibration, two-view calibration}
    \label{fig:calib_res_full}
\end{figure*}

\begin{figure*}[!htbp]
    \centering
    \begin{tabular}{c@{ }c@{ }c@{ }c@{ }}
         \includegraphics[width=.16\columnwidth]{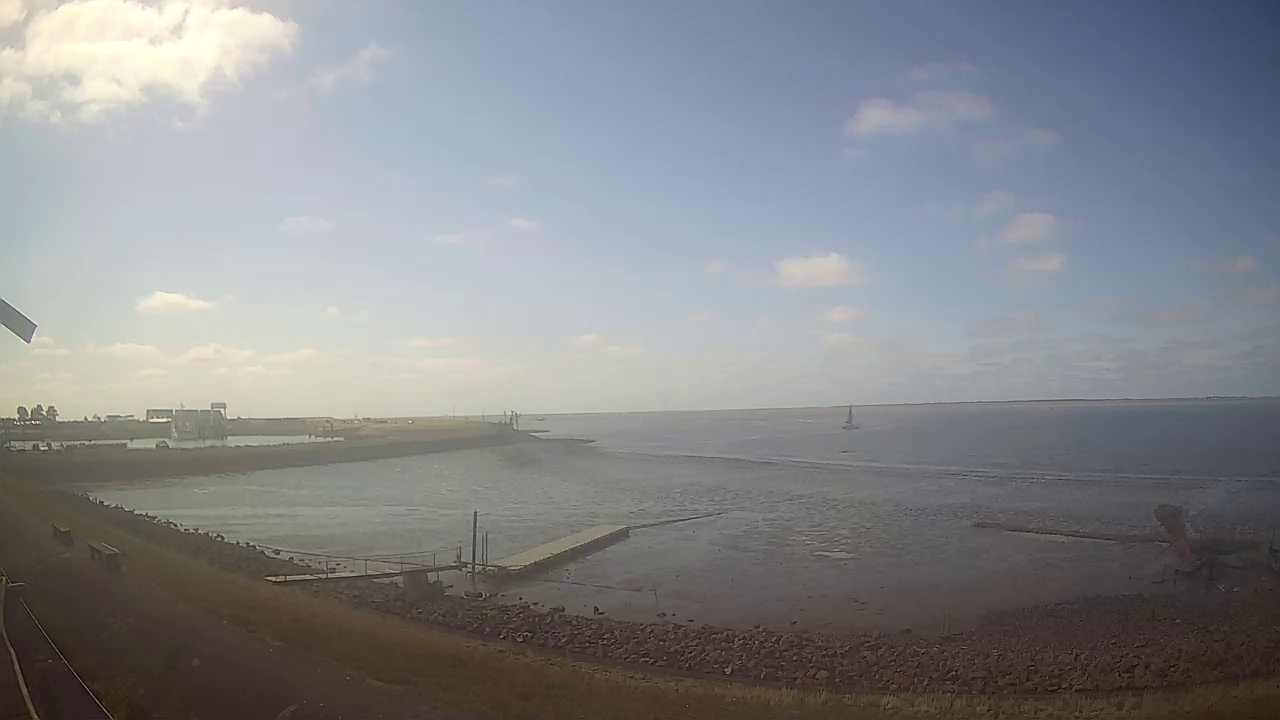} & \includegraphics[width=.16\columnwidth]{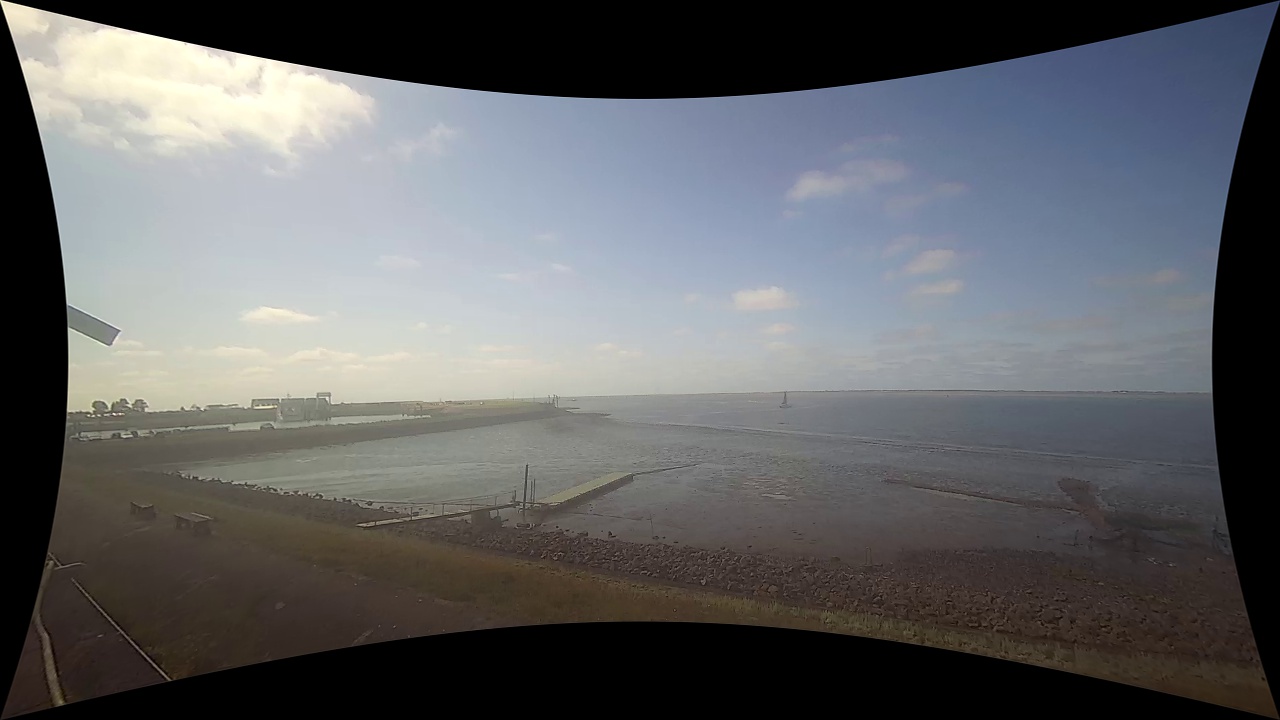} & \includegraphics[width=.16\columnwidth]{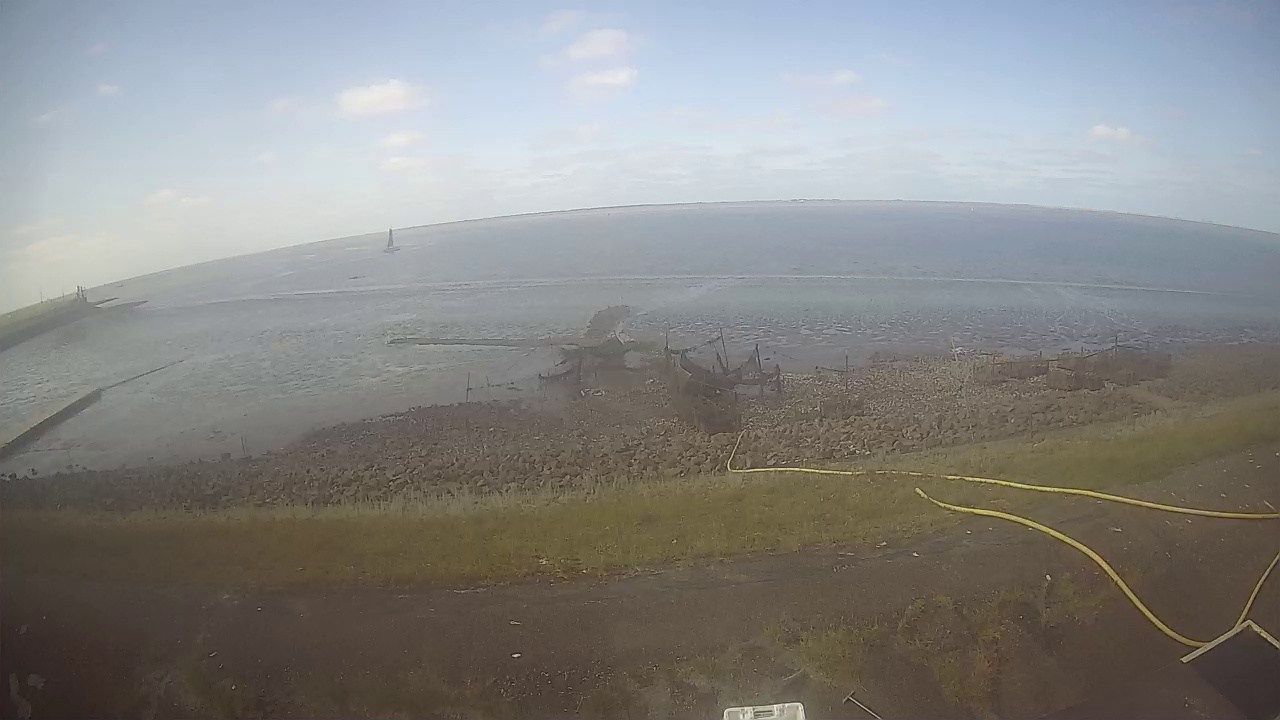} & \includegraphics[width=.16\columnwidth]{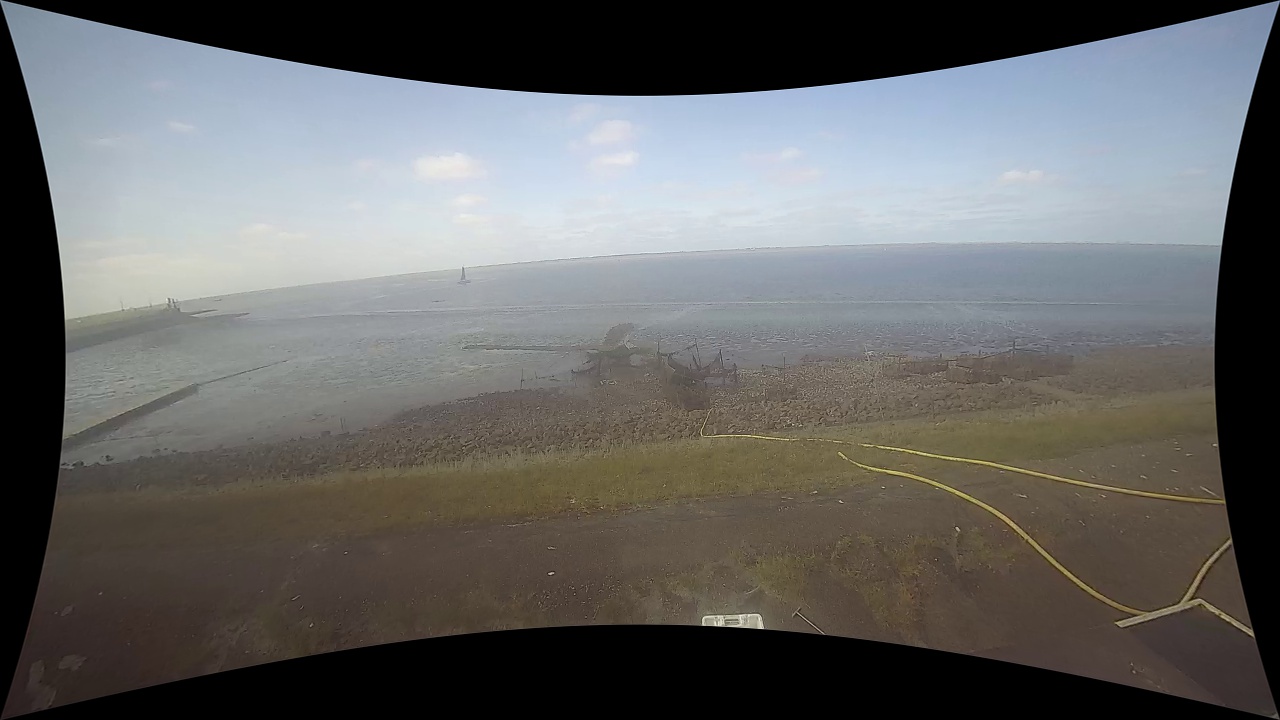}
    \end{tabular}
    \caption{The calibration results of LauwHav with two-view calibration approach.}
    \label{fig:calib_lauw}
\end{figure*}

Here we present the complete results of the single- and two-view calibration approach on all nine datasets. The undistorted images with the estimated camera parameters are shown in Figure \ref{fig:calib_res_full} and Table~\ref{tab:calib_results_full} respectively. Note that for the dataset LauwHav, the single-view approach failed to compute the parameters due to the lack of features; hence, only the calibration results with the two-view approach are listed here (Figure \ref{fig:calib_lauw}).

\subsection{3D reconstruction}
\begin{figure*}
    \centering
    \begin{tabular}{c@{ }c@{ }c@{ }c@{ }c@{ }c@{ }c@{ }}
          & $SO$ & $SD_{sc}$ & $SD_{un}$ & $SO$ & $SD_{sc}$ & $SD_{un}$ \\
          \multirow{2}{*}{\rotname{Croatia}} & \includegraphics[width=.16\columnwidth]{images/reconstruction/croatia/magsac/single/static_only_cam0_ground_truth_reprojection.jpg} & \includegraphics[width=.16\columnwidth]{images/reconstruction/croatia/magsac/single/static_dynamic_sync_cam0_ground_truth_reprojection.jpg} & \includegraphics[width=.16\columnwidth]{images/reconstruction/croatia/magsac/single/static_dynamic_unsync_cam0_ground_truth_reprojection.jpg} & \includegraphics[width=.16\columnwidth]{images/reconstruction/croatia/magsac/single/static_only_cam1_ground_truth_reprojection.jpg} & \includegraphics[width=.16\columnwidth]{images/reconstruction/croatia/magsac/single/static_dynamic_sync_cam1_ground_truth_reprojection.jpg} & \includegraphics[width=.16\columnwidth]{images/reconstruction/croatia/magsac/single/static_dynamic_unsync_cam1_ground_truth_reprojection.jpg} \\
           & \includegraphics[width=.16\columnwidth]{images/reconstruction/croatia/magsac/2view/static_only_cam0_ground_truth_reprojection.jpg} & \includegraphics[width=.16\columnwidth]{images/reconstruction/croatia/magsac/2view/static_dynamic_sync_cam0_ground_truth_reprojection.jpg} & \includegraphics[width=.16\columnwidth]{images/reconstruction/croatia/magsac/2view/static_dynamic_unsync_cam0_ground_truth_reprojection.jpg} & \includegraphics[width=.16\columnwidth]{images/reconstruction/croatia/magsac/2view/static_only_cam1_ground_truth_reprojection.jpg} & \includegraphics[width=.16\columnwidth]{images/reconstruction/croatia/magsac/2view/static_dynamic_sync_cam1_ground_truth_reprojection.jpg} & \includegraphics[width=.16\columnwidth]{images/reconstruction/croatia/magsac/2view/static_dynamic_unsync_cam1_ground_truth_reprojection.jpg} \\
          \multirow{2}{*}{\rotname{NewYork}} & \includegraphics[width=.16\columnwidth]{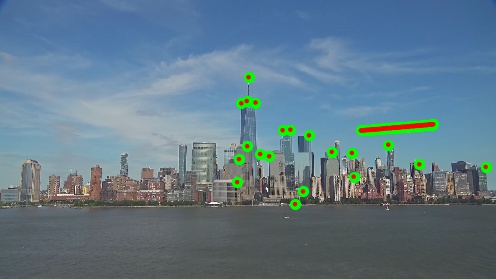} & \includegraphics[width=.16\columnwidth]{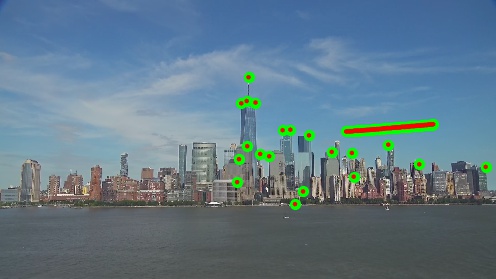} & \includegraphics[width=.16\columnwidth]{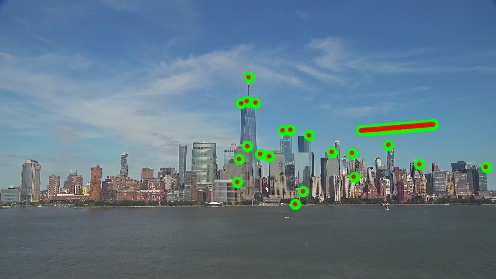} & \includegraphics[width=.16\columnwidth]{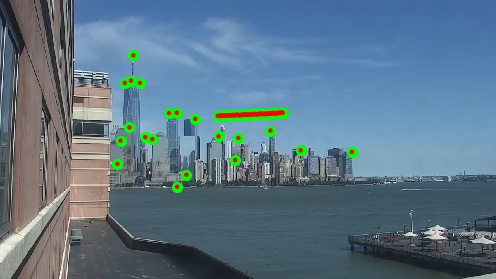} & \includegraphics[width=.16\columnwidth]{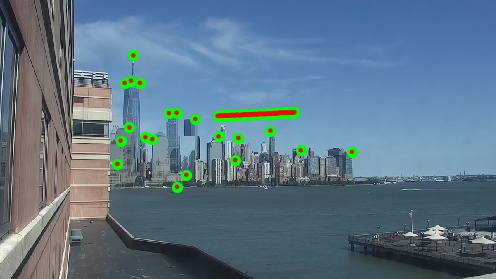} & \includegraphics[width=.16\columnwidth]{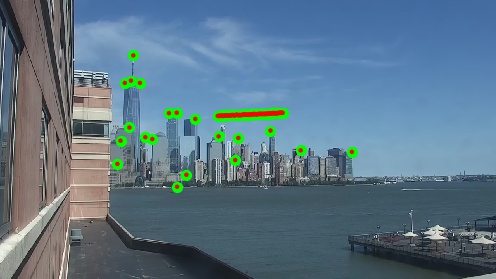} \\
          & \includegraphics[width=.16\columnwidth]{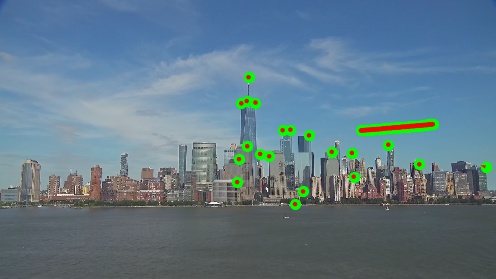} & \includegraphics[width=.16\columnwidth]{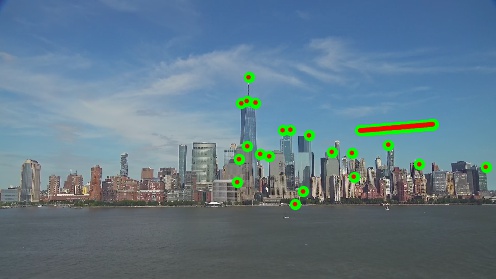} & \includegraphics[width=.16\columnwidth]{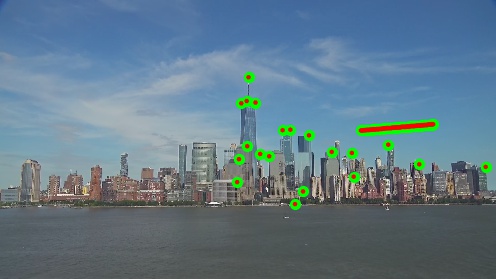} & \includegraphics[width=.16\columnwidth]{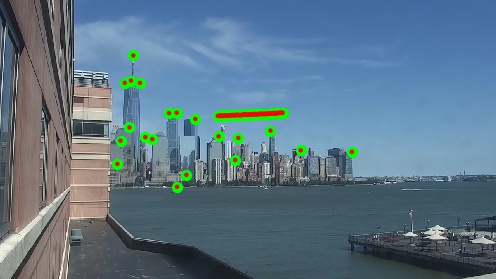} & \includegraphics[width=.16\columnwidth]{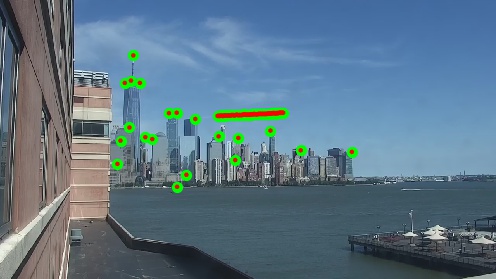} & \includegraphics[width=.16\columnwidth]{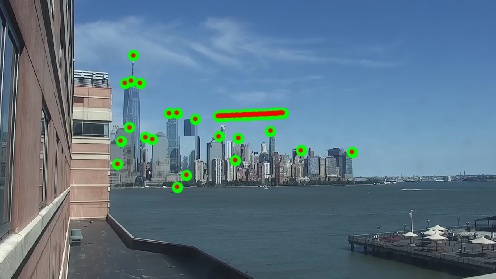} \\
          \multirow{2}{*}{\rotname{TimeSq}} & \includegraphics[width=.16\columnwidth]{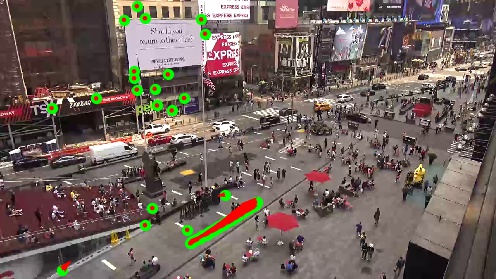} & \includegraphics[width=.16\columnwidth]{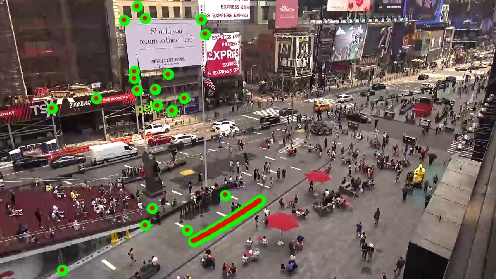} & \includegraphics[width=.16\columnwidth]{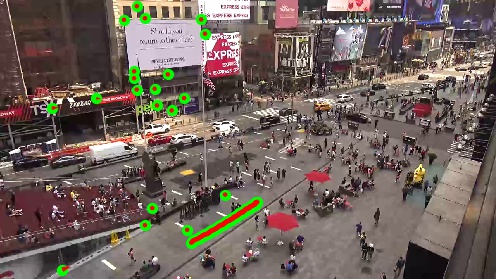} & \includegraphics[width=.16\columnwidth]{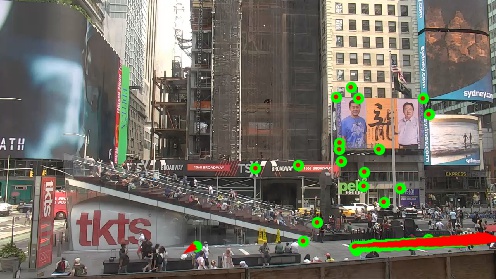} & \includegraphics[width=.16\columnwidth]{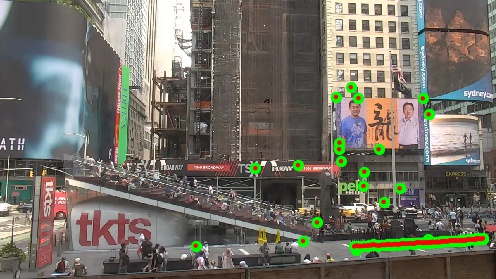} & \includegraphics[width=.16\columnwidth]{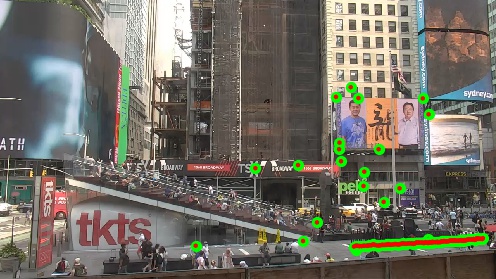} \\
          & \includegraphics[width=.16\columnwidth]{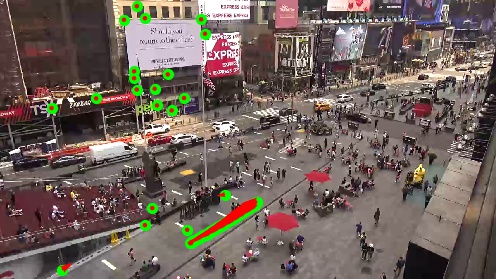} & \includegraphics[width=.16\columnwidth]{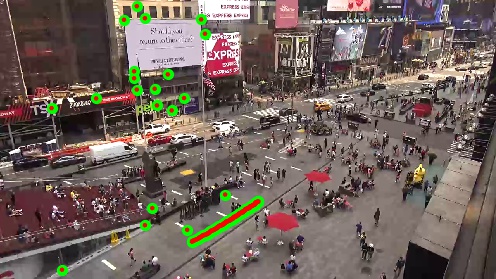} & \includegraphics[width=.16\columnwidth]{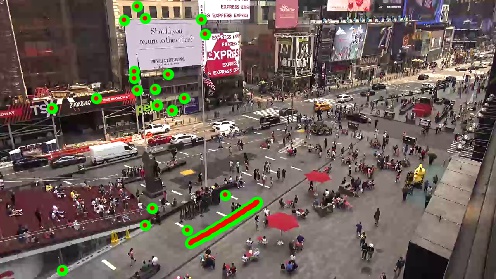} & \includegraphics[width=.16\columnwidth]{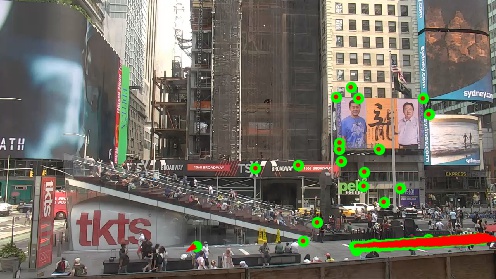} & \includegraphics[width=.16\columnwidth]{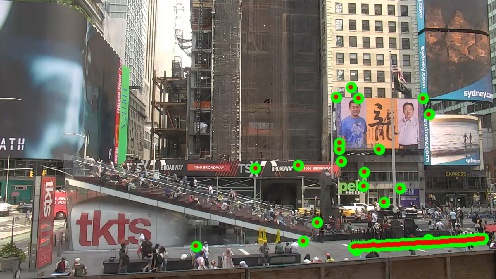} & \includegraphics[width=.16\columnwidth]{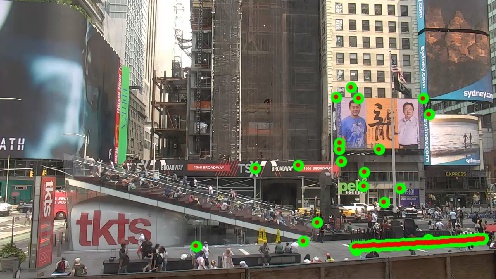} \\
          \multirow{2}{*}{\rotname{AalsHav}} & \includegraphics[width=.16\columnwidth]{images/reconstruction/aalsmeer/magsac/single/static_only_cam0_ground_truth_reprojection.jpg} & \includegraphics[width=.16\columnwidth]{images/reconstruction/aalsmeer/magsac/single/static_dynamic_sync_cam0_ground_truth_reprojection.jpg} & \includegraphics[width=.16\columnwidth]{images/reconstruction/aalsmeer/magsac/single/static_dynamic_unsync_cam0_ground_truth_reprojection.jpg} & \includegraphics[width=.16\columnwidth]{images/reconstruction/aalsmeer/magsac/single/static_only_cam1_ground_truth_reprojection.jpg} & \includegraphics[width=.16\columnwidth]{images/reconstruction/aalsmeer/magsac/single/static_dynamic_sync_cam1_ground_truth_reprojection.jpg} & \includegraphics[width=.16\columnwidth]{images/reconstruction/aalsmeer/magsac/single/static_dynamic_unsync_cam1_ground_truth_reprojection.jpg} \\
          & \includegraphics[width=.16\columnwidth]{images/reconstruction/aalsmeer/magsac/2view/static_only_cam0_ground_truth_reprojection.jpg} & \includegraphics[width=.16\columnwidth]{images/reconstruction/aalsmeer/magsac/2view/static_dynamic_sync_cam0_ground_truth_reprojection.jpg} & \includegraphics[width=.16\columnwidth]{images/reconstruction/aalsmeer/magsac/2view/static_dynamic_unsync_cam0_ground_truth_reprojection.jpg} & \includegraphics[width=.16\columnwidth]{images/reconstruction/aalsmeer/magsac/2view/static_only_cam1_ground_truth_reprojection.jpg} & \includegraphics[width=.16\columnwidth]{images/reconstruction/aalsmeer/magsac/2view/static_dynamic_sync_cam1_ground_truth_reprojection.jpg} & \includegraphics[width=.16\columnwidth]{images/reconstruction/aalsmeer/magsac/2view/static_dynamic_unsync_cam1_ground_truth_reprojection.jpg} \\
          \multirow{2}{*}{\rotname{BranHot}} & \includegraphics[width=.16\columnwidth]{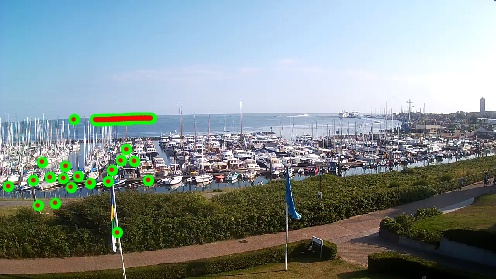} & \includegraphics[width=.16\columnwidth]{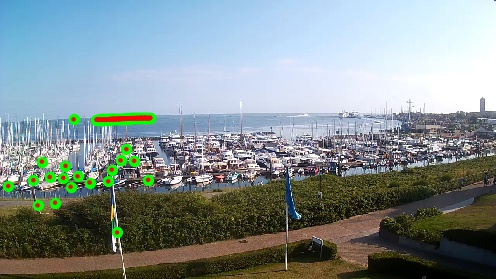} & \includegraphics[width=.16\columnwidth]{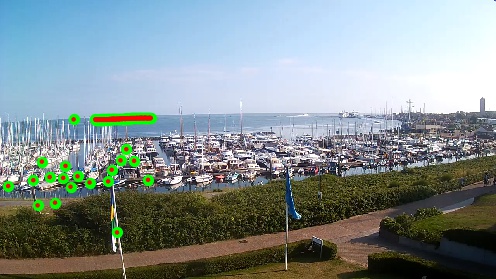} & \includegraphics[width=.16\columnwidth]{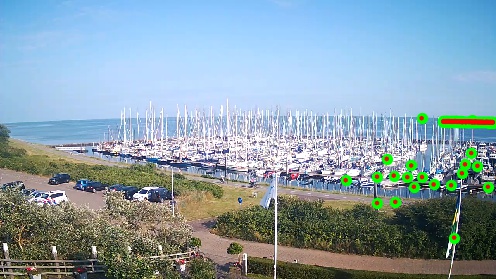} & \includegraphics[width=.16\columnwidth]{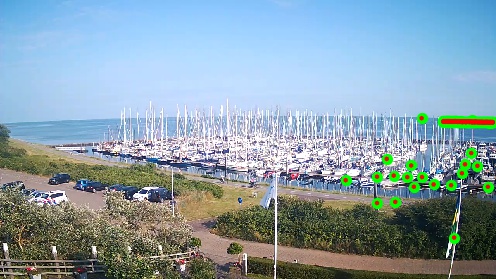} & \includegraphics[width=.16\columnwidth]{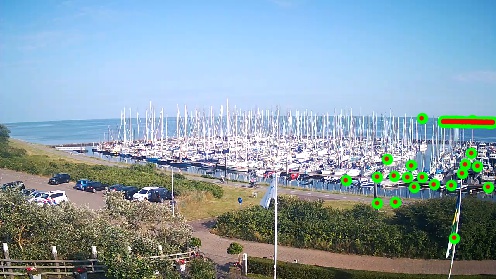} \\
          & \includegraphics[width=.16\columnwidth]{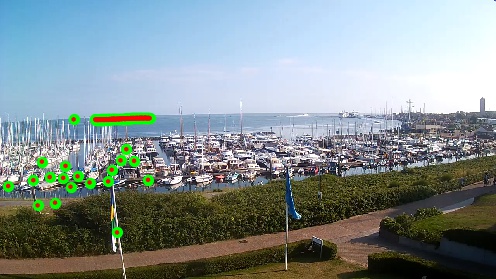} & \includegraphics[width=.16\columnwidth]{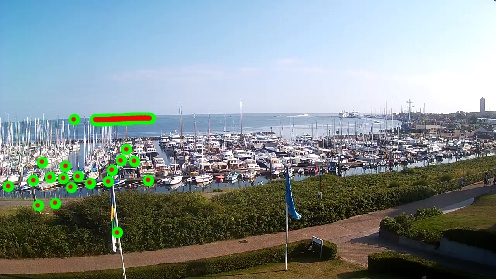} & \includegraphics[width=.16\columnwidth]{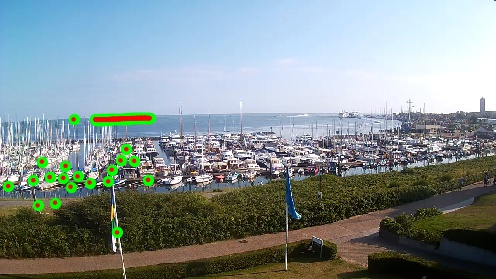} & \includegraphics[width=.16\columnwidth]{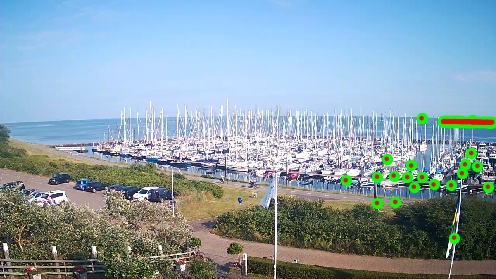} & \includegraphics[width=.16\columnwidth]{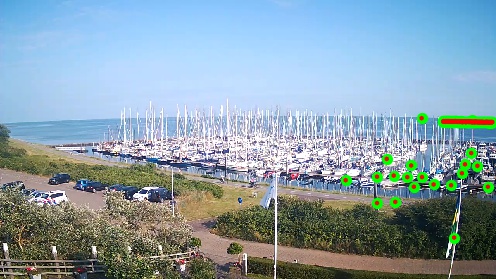} & \includegraphics[width=.16\columnwidth]{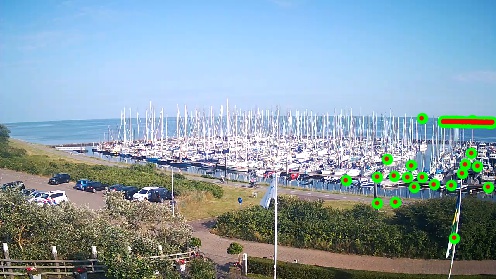} \\
          \multirow{2}{*}{\rotname{BranPor}} & \includegraphics[width=.16\columnwidth]{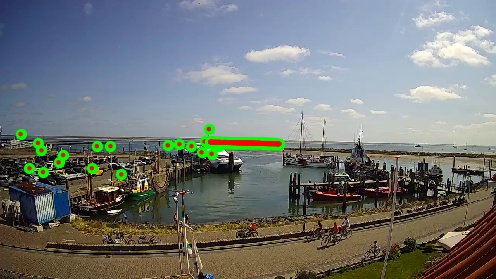} & \includegraphics[width=.16\columnwidth]{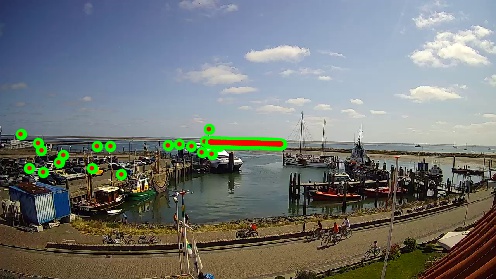} & \includegraphics[width=.16\columnwidth]{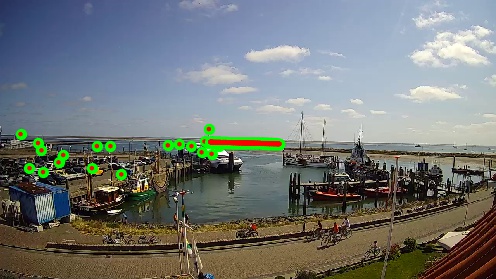} & \includegraphics[width=.16\columnwidth]{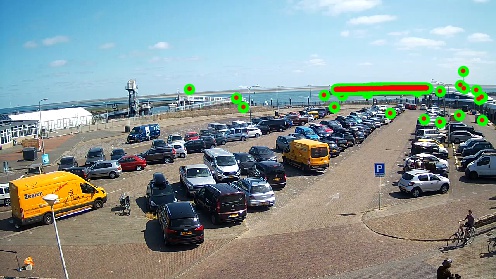} & \includegraphics[width=.16\columnwidth]{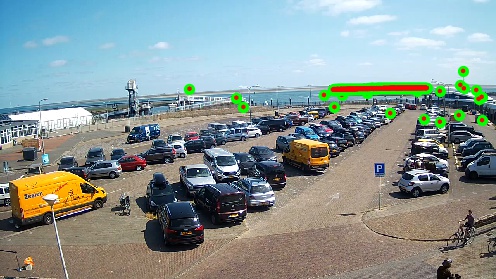} & \includegraphics[width=.16\columnwidth]{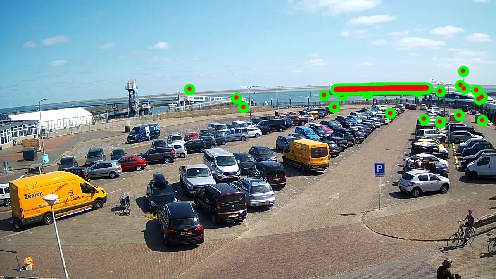} \\
          & \includegraphics[width=.16\columnwidth]{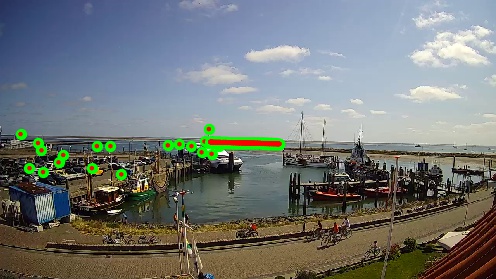} & \includegraphics[width=.16\columnwidth]{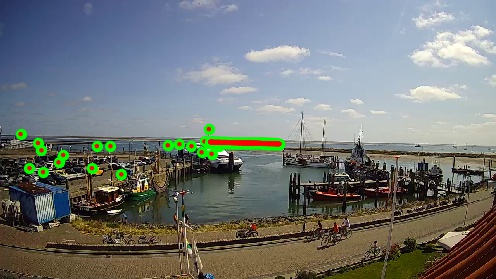} & \includegraphics[width=.16\columnwidth]{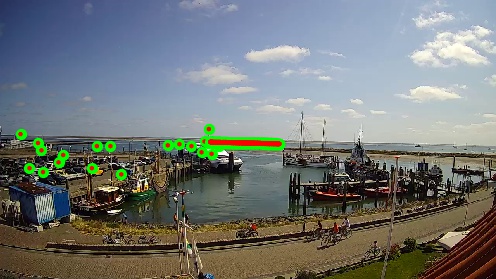} & \includegraphics[width=.16\columnwidth]{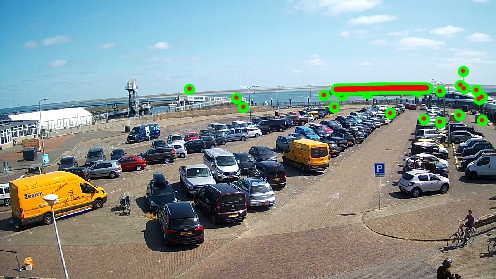} & \includegraphics[width=.16\columnwidth]{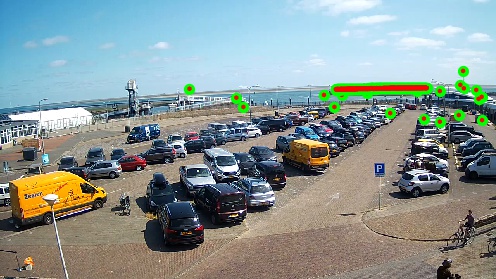} & \includegraphics[width=.16\columnwidth]{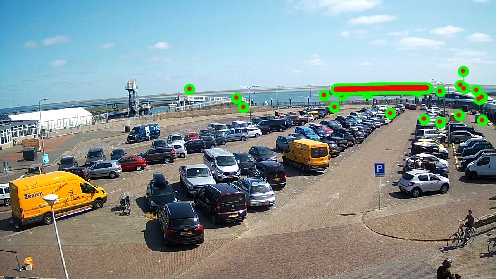}
    \end{tabular}
    \caption{The reconstruction results of the control points and trajectories in the static-only ($SO$), static-dynamic-sync ($SD_{sc}$) and static-dynamic-unsync ($SD_{un}$) setting initialized with single- (top) and two-view (bottom) calibration. The red arrows indicate the reprojection error magnitudes.}
    \label{fig:recon_res1}
\end{figure*}

\begin{figure*}
    \centering
    \begin{tabular}{c@{ }c@{ }c@{ }c@{ }c@{ }c@{ }c@{ }}
          & $SO$ & $SD_{sc}$ & $SD_{un}$ & $SO$ & $SD_{sc}$ & $SD_{un}$ \\
          \multirow{2}{*}{\rotname{RottPor}} & \includegraphics[width=.16\columnwidth]{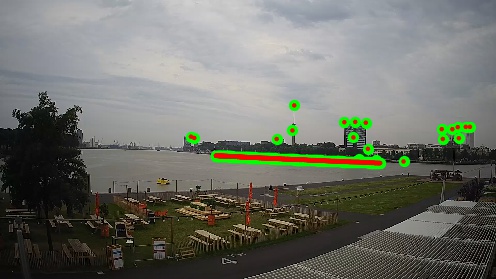} & \includegraphics[width=.16\columnwidth]{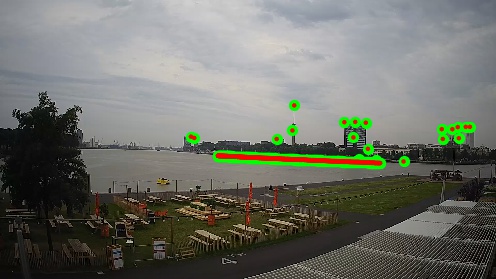} & \includegraphics[width=.16\columnwidth]{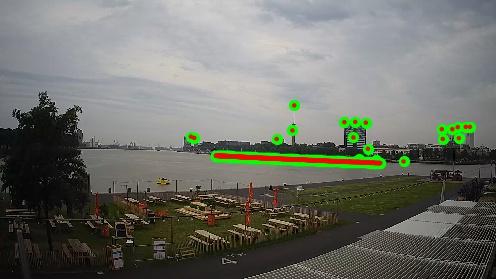} & \includegraphics[width=.16\columnwidth]{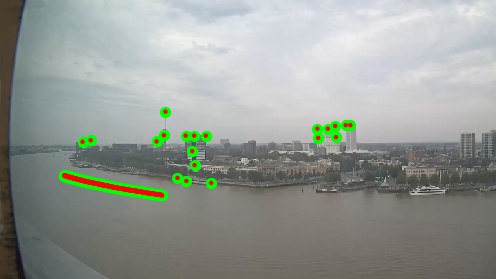} & \includegraphics[width=.16\columnwidth]{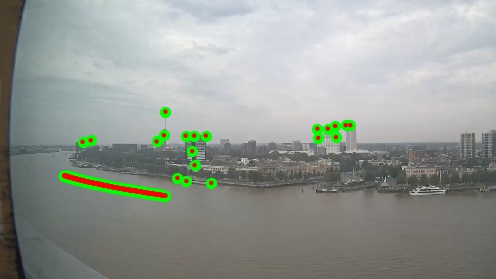} & \includegraphics[width=.16\columnwidth]{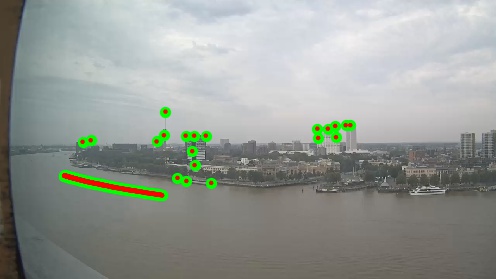} \\
          & \includegraphics[width=.16\columnwidth]{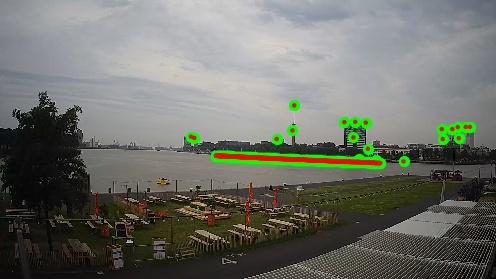} & \includegraphics[width=.16\columnwidth]{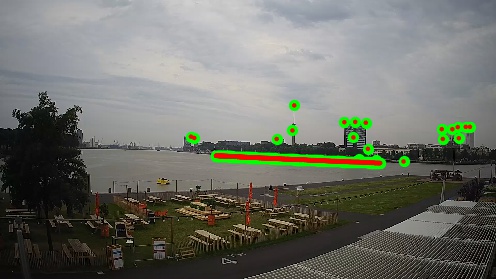} & \includegraphics[width=.16\columnwidth]{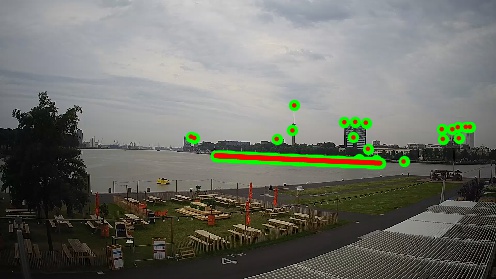} & \includegraphics[width=.16\columnwidth]{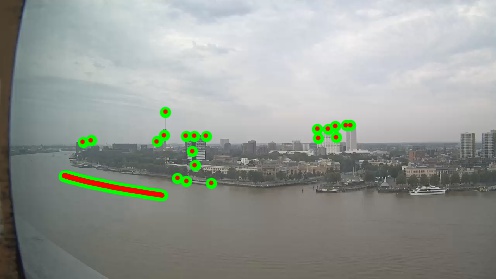} & \includegraphics[width=.16\columnwidth]{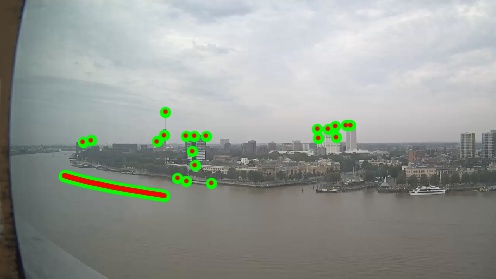} & \includegraphics[width=.16\columnwidth]{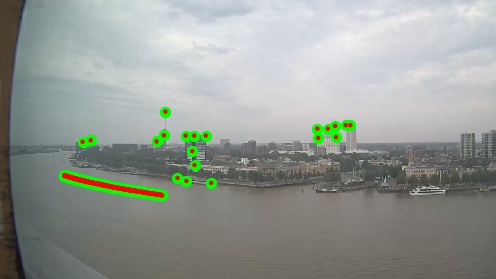} \\
          \multirow{2}{*}{\rotname{TexAir}} & \includegraphics[width=.16\columnwidth]{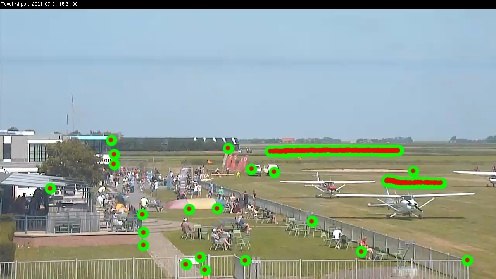} &  \includegraphics[width=.16\columnwidth]{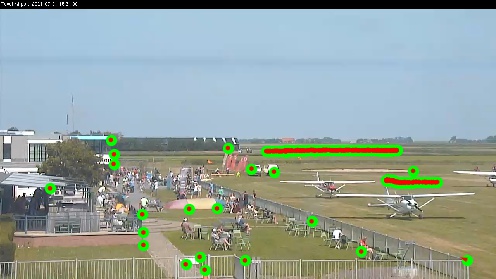} &  \includegraphics[width=.16\columnwidth]{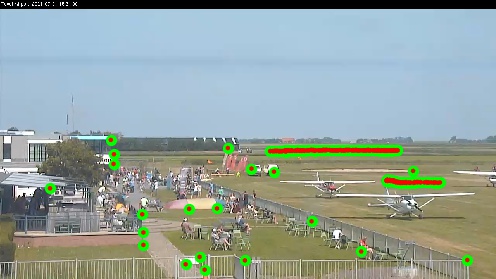} &  \includegraphics[width=.16\columnwidth]{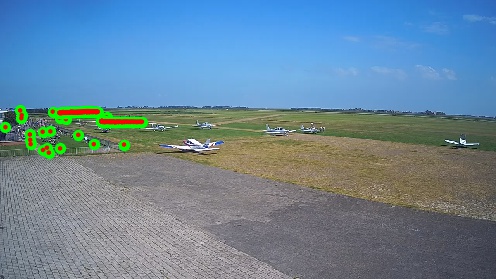} &  \includegraphics[width=.16\columnwidth]{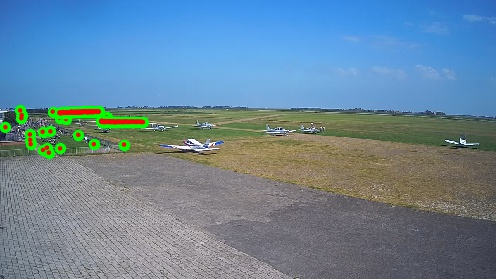} &  \includegraphics[width=.16\columnwidth]{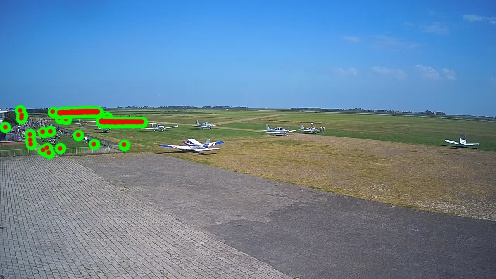} \\
          & \includegraphics[width=.16\columnwidth]{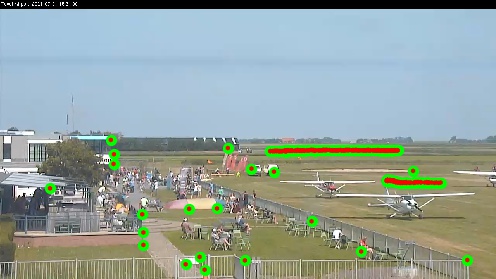} & \includegraphics[width=.16\columnwidth]{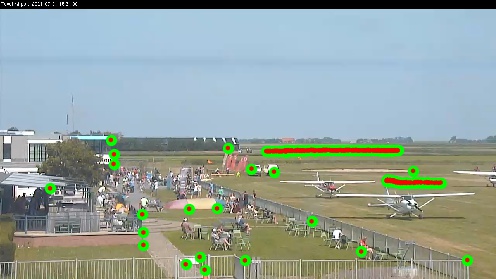} & \includegraphics[width=.16\columnwidth]{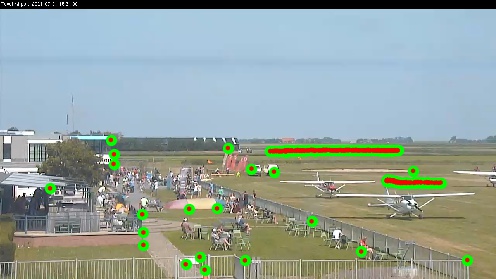} & \includegraphics[width=.16\columnwidth]{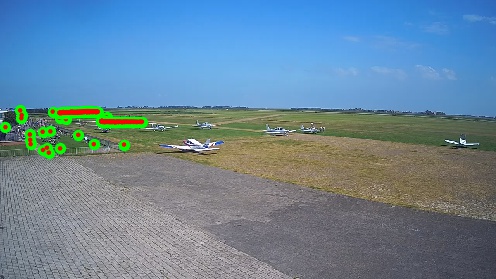} & \includegraphics[width=.16\columnwidth]{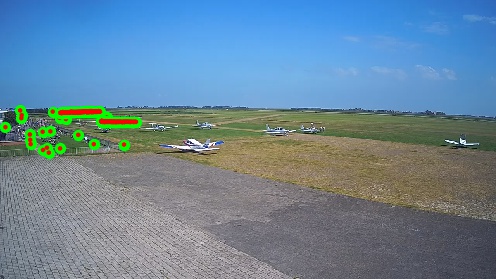} & \includegraphics[width=.16\columnwidth]{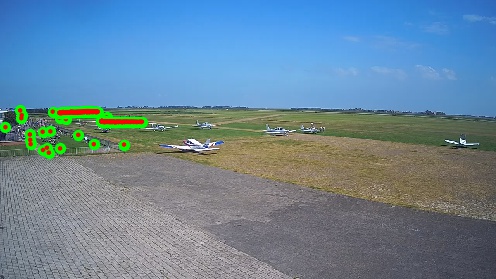}
    \end{tabular}
    \caption{The reconstruction results of the control points and trajectories (Continued).}
    \label{fig:recon_res2}
\end{figure*}

\begin{figure*}
    \centering
    \begin{tabular}{c@{ }c@{ }c@{ }c@{ }c@{ }c@{ }}
          $SO$ & $SD_{sc}$ & $SD_{un}$ & $SO$ & $SD_{sc}$ & $SD_{un}$ \\
          \includegraphics[width=.16\columnwidth]{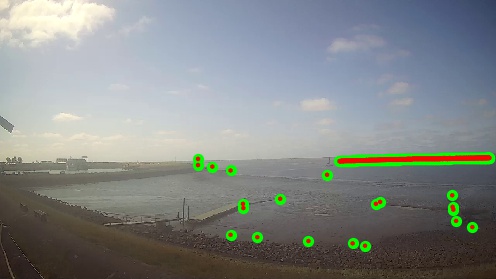} & \includegraphics[width=.16\columnwidth]{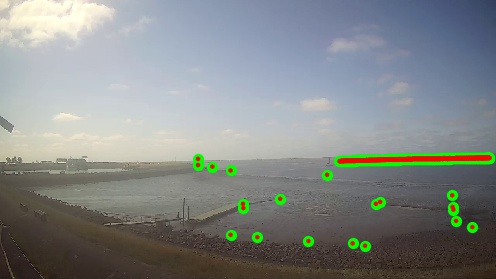} & \includegraphics[width=.16\columnwidth]{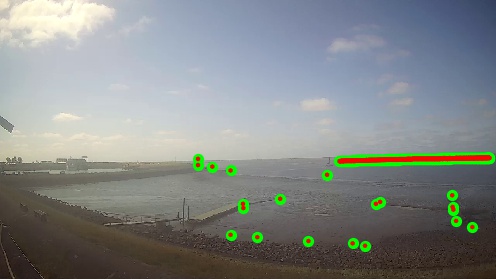} & \includegraphics[width=.16\columnwidth]{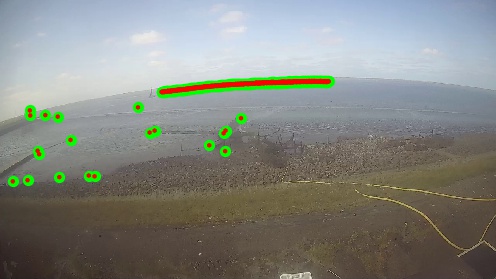} & \includegraphics[width=.16\columnwidth]{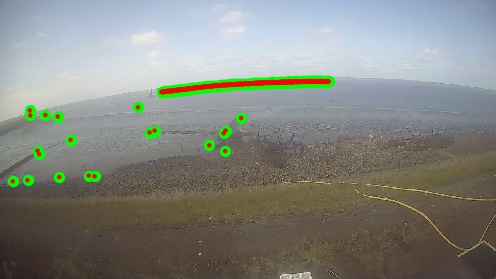} & \includegraphics[width=.16\columnwidth]{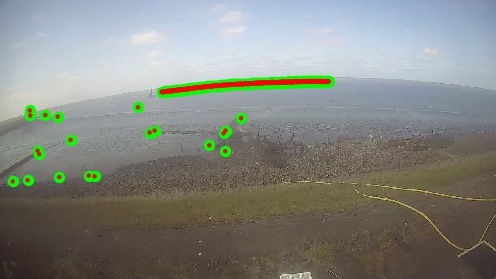}
    \end{tabular}
    \caption{The reconstruction results of the control points and trajectories of LauwHav with initial camera parameters from the two-view calibration.}
    \label{fig:recon_lauw}
\end{figure*}

Figure \ref{fig:recon_res1} and Figure \ref{fig:recon_res2} illustrates the qualitative reconstruction results in the form of reprojection error of the control points and the trajectory. The red vector at each point represents the direction and magnitude of the error. For every camera, the reconstruction results acquired with the static-only, static-dynamic-sync, and static-dynamic-unsync setting are shown from left to right.  For each dataset and estimator method, the top row displays the results with camera parameters estimated from the single-view calibration method, while the bottom row from the two-view calibration approach. Note that, as mentioned in previous section, the single-view approach failed to converge on LauwHav dataset, only the reconstruction results with camera parameters from the two-view calibration approach are tested (Figure \ref{fig:recon_lauw}). 

In Table~\ref{tab:recon_results_full} we show the complete quantitative comparison of reconstruction results initialized by two-view geometry estimation using RANSAC and MAGSAC (denoted as `R` and `M`) for both single- and two-view undistortion initialization (denoted as `Sngl` and `Two`). The trend is that MAGSAC inlier correspondences lead in general to a lower reprojection error on the control points than vanilla RANSAC ones. In case of the TexAir dataset, MAGSAC makes the difference between failure and a success.

\begin{table}[th]
    \centering
    \begin{tabular}{c|c|c|c|c|c|c}
         Dataset              & SIFT & R & M & SG &  R &  M \\
         Croatia              & 434  & 20 & 17      & 505       & 264 & 379     \\
         NewYork             & 279  & 144 & 169      & 549       & 410  & 514    \\
         TimeSq          & 272  & 47  & 52     & 608       & 374   & 468   \\
         AalsHav       & 325  & 32 & 61      & 161       & 96    & 118    \\
         BranHot      & 660  & 530 & 580     & 607       & 521    & 580  \\
         BranPor       & 119  & 12  & 17     & 365       & 199   & 261   \\
         RottPor       & 454  & 23  & 177     & 453       & 159   & 176   \\
         TexAir        & 154  & 68   & 83    & 184       & 117 & 176     \\
         LauwHav  & 13   & 7 & 8       & 323       & 223   & 275   \\
    \end{tabular}
    \caption{Results of matching with SIFT and SuperGlue (SG) and geometric verification using RANSAC (R) and MAGSAC (M). Superglue provides many more matches and the difference is even amplified when using state-of-the-art geometric verification MAGSAC.}
    \label{tab:matching_results_full}
\end{table}

\begin{table*}[]
    \centering
    \renewcommand{\tabcolsep}{4pt}
    \renewcommand{\arraystretch}{1}
    \begin{tabular}{c|c|c|c|c|c|c|c|c|c|c|c|c|c}
         \multicolumn{2}{c|}{ } & \multicolumn{3}{c|}{$e^1_{cp}$}  & \multicolumn{3}{c|}{$e^1_{traj}$}  & \multicolumn{3}{c|}{$e^2_{cp}$}    & \multicolumn{3}{c}{$e^2_{traj}$} \\ \cline{3-14} 
         \multicolumn{2}{c|}{ } & $SO$     & $SD_{sc}$ & $SD_{un}$      & $SO$     & $SD_{sc}$ & $SD_{un}$        & $SO$     & $SD_{sc}$ & $SD_{un}$         & $SO$     & $SD_{sc}$ & $SD_{un}$     \\
         \hline
\rowcolor{green!50}
 & Sngl\_R & 16.6 & 24.7 & 17.7 & 36.7 & 2.8 & 4.9 & 10.6 & 9.0 & 6.6 & 32.8 & 1.1 & 1.8 \\ 
\rowcolor{green!50}
 & Two\_R & 4.2 & 16.5 & 14.3 & 46.9 & 2.9 & 4.4 & 4.9 & 9.9 & 8.4 & 67.2 & 1.9 & 2.7 \\ 
\cline{2-14} 
\rowcolor{green!50}
 & Sngl\_M & 1.8 & 15.2 & 16.3 & 71.3 & 3.6 & 3.6 & 1.3 & 5.1 & 5.7 & 46.6 & 1.2 & 1.3 \\ 
\rowcolor{green!50}
\multirow{-4}{*}{Croatia} & Two\_M & 2.9 & 12.7 & 11.9 & 66.3 & 2.6 & 2.9 & 2.3 & 7.0 & 7.0 & 54.4 & 1.5 & 1.7 \\ 
\hline 
\rowcolor{green!50}
 & Sngl\_R & 0.4 & 0.3 & 0.6 & 6.0 & 0.3 & 0.4 & 0.5 & 0.4 & 0.7 & 6.3 & 0.3 & 0.3 \\ 
\rowcolor{green!50}
 & Two\_R & 0.2 & 0.4 & 0.3 & 2.8 & 0.1 & 0.1 & 0.6 & 1.7 & 1.8 & 6.0 & 0.8 & 1.2 \\ 
\cline{2-14} 
\rowcolor{green!50}
 & Sngl\_M & 0.5 & 0.4 & 0.5 & 3.2 & 0.3 & 0.3 & 0.5 & 0.6 & 0.7 & 3.1 & 0.3 & 0.3 \\ 
\rowcolor{green!50}
\multirow{-4}{*}{NewYork} & Two\_M & 0.2 & 0.2 & 0.2 & 1.8 & 0.1 & 0.1 & 0.6 & 0.6 & 0.7 & 5.0 & 0.4 & 0.4 \\ 
\hline 
\rowcolor{green!50}
 & Sngl\_R & 1.3 & 0.9 & 0.9 & 4.6 & 0.4 & 0.5 & 1.0 & 0.7 & 0.6 & 4.6 & 0.4 & 0.4 \\ 
\rowcolor{green!50}
 & Two\_R & 3.2 & 0.7 & 0.7 & 24.6 & 0.3 & 0.6 & 1.9 & 0.6 & 0.4 & 15.6 & 0.3 & 0.4 \\ 
\cline{2-14}  
\rowcolor{green!50}
 & Sngl\_M & 3.5 & 1.1 & 0.8 & 24.8 & 0.5 & 0.7 & 4.3 & 0.7 & 0.4 & 35.5 & 0.4 & 0.5 \\ 
\rowcolor{green!50}
\multirow{-4}{*}{TimeSq} & Two\_M & 2.7 & 0.9 & 0.8 & 22.5 & 0.4 & 0.5 & 3.8 & 0.7 & 0.5 & 34.5 & 0.4 & 0.5 \\ 
\hline 
\rowcolor{red!50}
 & Sngl\_R & 2.8 & 24.2 & 23.2 & 0.3 & 0.3 & 0.3 & 41.6 & 89.2 & 101.6 & 4.4 & 0.7 & 0.8 \\ 
\rowcolor{red!50}
 & Two\_R & 421.6 & 206.3 & 68.6 & 22.9 & 0.7 & 0.7 & 43.7 & 466.1 & 117.2 & 9.8 & 0.4 & 0.5 \\ 
\cline{2-14}  
\rowcolor{red!50}
 & Sngl\_M & 2.8 & 3.0 & 23.2 & 0.3 & 0.3 & 0.3 & 41.6 & 3.7 & 101.6 & 4.4 & 0.6 & 0.8 \\ 
\rowcolor{red!50}
\multirow{-4}{*}{AalsHav} & Two\_M & 34.7 & 29.0 & 32.7 & 7.2 & 0.3 & 0.6 & 54.8 & 41.5 & 20.2 & 12.7 & 0.6 & 0.8 \\ 
\hline 
\rowcolor{green!50}
 & Sngl\_R & 0.6 & 0.5 & 0.6 & 0.9 & 0.7 & 0.8 & 0.2 & 0.3 & 0.4 & 0.4 & 0.4 & 0.5 \\ 
 \rowcolor{green!50}
 & Two\_R & 0.5 & 0.4 & 0.4 & 0.5 & 0.5 & 0.5 & 0.5 & 0.5 & 0.5 & 0.6 & 0.6 & 0.6 \\ 
\cline{2-14}  
\rowcolor{green!50}
 & Sngl\_M & 0.5 & 0.6 & 0.6 & 1.3 & 0.6 & 0.7 & 0.3 & 0.2 & 0.2 & 0.7 & 0.3 & 0.3 \\ 
 \rowcolor{green!50}
\multirow{-4}{*}{BranHot} & Two\_M & 0.3 & 0.4 & 0.5 & 1.0 & 0.5 & 0.5 & 0.3 & 0.4 & 0.5 & 1.1 & 0.5 & 0.6 \\ 
\hline 
\rowcolor{green!50}
 & Sngl\_R & 1.2 & 1.2 & 1.2 & 1.2 & 0.2 & 0.2 & 1.0 & 1.2 & 1.2 & 1.2 & 0.3 & 0.3 \\ 
 \rowcolor{green!50}
 & Two\_R & 0.9 & 3.6 & 2.4 & 1.5 & 0.3 & 0.2 & 1.5 & 2.2 & 1.7 & 3.3 & 0.2 & 0.2 \\ 
\cline{2-14}  
\rowcolor{green!50}
 & Sngl\_M & 1.4 & 2.3 & 2.1 & 0.9 & 0.2 & 0.2 & 1.1 & 1.8 & 1.7 & 0.8 & 0.2 & 0.2 \\ 
 \rowcolor{green!50}
\multirow{-4}{*}{BranPor} & Two\_M & 1.0 & 2.3 & 2.2 & 0.7 & 0.2 & 0.2 & 1.1 & 2.3 & 2.2 & 0.9 & 0.3 & 0.3 \\ 
\hline 
\rowcolor{green!50}
 & Sngl\_R & 1.6 & 1.4 & 1.1 & 1.6 & 0.2 & 0.2 & 2.5 & 2.1 & 1.7 & 2.7 & 0.3 & 0.3 \\ 
 \rowcolor{green!50}
 & Two\_R & 1.0 & 1.2 & 0.9 & 1.3 & 0.2 & 0.2 & 1.4 & 1.8 & 1.3 & 2.0 & 0.3 & 0.4 \\ 
\cline{2-14}  
\rowcolor{green!50}
 & Sngl\_M & 1.1 & 1.1 & 1.2 & 0.7 & 0.2 & 0.2 & 1.6 & 1.7 & 1.8 & 1.2 & 0.4 & 0.3 \\ 
 \rowcolor{green!50}
\multirow{-4}{*}{RottPor} & Two\_M & 1.3 & 1.2 & 1.2 & 2.1 & 0.2 & 0.2 & 1.8 & 1.7 & 1.6 & 3.2 & 0.3 & 0.3 \\ 
\hline 
\rowcolor{red!50}
 & Sngl\_R & 1.4 & 83.7 & 20.3 & 49.7 & 0.9 & 1.0 & 0.3 & 2.7 & 2.8 & 8.0 & 0.1 & 0.2 \\ 
\rowcolor{red!50}
 & Two\_R & 2.4 & 11.6 & 22.6 & 32.8 & 1.3 & 1.5 & 0.0 & 0.0 & 0.0 & 0.0 & 0.0 & 0.0 \\ 
\cline{2-14} 
\rowcolor{green!50}
 & Sngl\_M & 1.0 & 0.2 & 0.3 & 4.4 & 0.1 & 0.1 & 0.4 & 0.8 & 0.9 & 1.6 & 0.5 & 0.5 \\ 
\rowcolor{green!50}
\multirow{-4}{*}{TexAir} & Two\_M & 2.0 & 2.0 & 2.4 & 5.9 & 1.1 & 1.3 & 0.0 & 0.2 & 0.2 & 0.1 & 0.1 & 0.1 \\ 
\hline 
\rowcolor{green!50}
 & Two\_R & 0.6 & 0.9 & 0.9 & 1.1 & 0.3 & 0.3 & 0.6 & 1.1 & 1.2 & 0.9 & 0.3 & 0.3 \\ 
\cline{2-14}  
\rowcolor{green!50}
\multirow{-2}{*}{LauwHav} & Two\_M & 0.5 & 0.5 & 0.5 & 0.7 & 0.3 & 0.3 & 0.5 & 0.6 & 0.6 & 0.6 & 0.3 & 0.3 \\ 
 \hline

    \end{tabular}
    \caption{Reprojection error of static control points and trajectories in both cameras under the static-only, static-dynamic-sync, and static-dynamic-unsync setting with two different initialization of camera parameters from the single- and two-view calibration. Both vanilla RANSAC and MAGSAC robust geometry estimators are compared as `R` and `M` respectively. The success cases are marked in green, and the failure cases in red. For TexAir dataset, using MAGSAC is the difference between a failure and a success.}
    \label{tab:recon_results_full}
\end{table*}

\begin{table*}[th]
    \centering
    \begin{tabular}{c|c|c|c|c|c|c|c|c}
         Dataset              & $f^1_{s}$  & $f^2_{s}$  & $f^1_{t}$  & $f^2_{t}$ & $d^1_{s}$  & $d^2_{s}$  & $d^1_{t}$  & $d^2_{t}$ \\
         Croatia              & 1068.3    & 991.2     & 842.9     & 616.6    & -0.53      & -0.44      & -0.38      & -0.14 \\
         NewYork             & 1408.3    & 1075.1    & 1296.5    &  396.9    & -0.04      & -0.06      &  0.23       &  -0.02 \\
         TimeSq          & 767.5     & 878.9     & 938.9     & 878.5    & -0.08      & -0.06      & -0.05      & -0.05 \\
         AalsHav       &  2688.9    &  1031.9    & 176.1     &  357.8    &  -1.98      &  -0.47      &  -0.01      &  -0.02 \\
         BranHot      &  1001.6    &  1705.3    &  1087.9    &  897.8    &  -0.43      &  -1.56      &  -0.52      &  -0.27 \\
         BranPor       & 1198.3    & 1286.9    & 1825.7    & 1762.5   & -0.93      & -0.77      & -3.72      & -1.33 \\
         RottPor       &  1390.0    &  896.6     & 1112.7    &  885.9    &  -0.61      &  -0.51      &  -0.41      &  -0.56 \\
         TexAir        &  1394.7    &  1696.9    &  828.8     &  12690.8  &  -0.00      &  -1.71      &  0.04       &  -132.10 \\
         LauwHav &  -          &  -          &  951.0     &  1039.6   &  -          & -          &  -0.56      &  -0.65 \\
    \end{tabular}
    \caption{Resulting focal lengths and distortion parameters for the tested datasets. Subscripts $s$ and $t$ denotes single-view method and two-view method respectively. }
    \label{tab:calib_results_full}
\end{table*}
\end{document}